\documentclass[a4paper,10pt]{article}

\usepackage[utf8]{inputenc} 
\usepackage[T1]{fontenc}    
\usepackage{hyperref}       
\usepackage{url}            
\usepackage{booktabs}       
\usepackage{amsfonts}       
\usepackage{nicefrac}       
\usepackage{microtype}      
\usepackage{lipsum}
\usepackage{fancyhdr}       
\usepackage{graphicx}       
\usepackage{orcidlink}
\usepackage{amscd}
\usepackage{mathtools}
\usepackage{mathrsfs}
\usepackage{enumerate}
\usepackage{bm}
\usepackage{siunitx}
\usepackage{algorithm}
\usepackage{algpseudocode}
\usepackage{caption}
\usepackage{epstopdf, epsfig}
\usepackage{amsmath}
\usepackage{amssymb}
\usepackage{amsopn}
\usepackage{amsthm}
\usepackage{tikz}
\usepackage{onimage}
\usepackage[caption=false,position=top]{subfig}
\usepackage{array}
\usepackage{cleveref}

\usetikzlibrary{cd}
\graphicspath{{media/}}     
\ifpdf
\fi

\ifpdf
\DeclareGraphicsExtensions{.eps,.pdf,.png,.jpg}
\else
\DeclareGraphicsExtensions{.eps}
\fi

\newtheorem{lemma}{Lemma}
\newtheorem{proposition}{Proposition}
\newtheorem{theorem}{Theorem}

\newtheorem{remark}{Remark}

\DeclareMathOperator*{\argmin}{\arg\!\min\enskip}
\DeclareMathOperator*{\argmax}{\arg\!\max\enskip}

\newcommand{\aligntridown}{\raise.4ex\hbox{$\bigtriangledown$}}

\numberwithin{equation}{section}

\DeclareMathOperator*{\supp}{supp}
\DeclareMathOperator{\spann}{span}

\DeclareMathOperator*{\score}{Score}
\DeclareMathOperator*{\SSR}{SSR}
\DeclareMathOperator*{\Pareto}{Pareto}

\allowdisplaybreaks
\ifpdf
  \DeclareGraphicsExtensions{.eps,.pdf,.png,.jpg}
\else
  \DeclareGraphicsExtensions{.eps}
\fi

\newcommand{\email}[1]{\protect\href{mailto:#1}{#1}}
\newcommand\funding[1]{\protect\\ \hspace*{15.37pt}{\bfseries Funding:} #1}

\begin{document}

\title{From STLS to Projection-based Dictionary Selection in Sparse Regression for System Identification
\thanks{Submitted to the editors DATE.
\funding{
The work by H. Cho was supported by the National Research Foundation of Korea (NRF) grant funded by the Korea government (MSIT) (RS-2023-00253171),
C. Oishi was supported by National Council for Scientific and Technological Development (CNPq), grants 305383/2019-1 and 307228/2023-1,
F. Amaral and C. Oishi were supported by the São Paulo Research Foundation (FAPESP) process numbers  2013/07375-0, 2021/13833-7, 2023/06035-2, 2021/07034-4. 
The authors acknowledge support from the National Science Foundation AI Institute in Dynamic Systems (grant number 2112085).}}}

\author{
Hangjun Cho \orcidlink{0000-0001-9920-094X} \thanks{AI Institute in Dynamic Systems, Department of Mechanical Engineering, University of Washington, Seattle, WA 98195, United States (\email{cho.1363@osu.edu}, \email{sbrunton@uw.edu})}
\and Fabio V.G. Amaral \orcidlink{0000-0001-6945-8376}\thanks{Departamento de Matemática e Computação, Faculdade de Ciências e Tecnologia, Universidade Estadual Paulista “Júlio de Mesquita Filho”, Presidente Prudente, Brazil (\email{fabio.amaral@unesp.br}, \email{cassio.oishi@unesp.br})}
\and Andrei A. Klishin \orcidlink{0000-0002-5740-8520}\thanks {Department of Mechanical Engineering, University of Hawai`i at M\=anoa, Honolulu 96822, United States (\email{aklishin@hawaii.edu})} 
\and Cassio M. Oishi \orcidlink{0000-0002-0904-6561}\footnotemark[3]
\and Steven L. Brunton \orcidlink{0000-0002-6565-5118}\footnotemark[2]
}
\maketitle

\begin{abstract}
In this work, we revisit dictionary-based sparse regression, in particular, Sequential Threshold Least Squares (STLS), and propose a score-guided library selection to provide practical guidance for data-driven modeling, with emphasis on SINDy-type algorithms. 
STLS is an algorithm to solve the $\ell_0$ sparse least-squares problem, which relies on splitting to efficiently solve the least-squares portion while handling the sparse term via proximal methods. It produces coefficient vectors whose components depend on both the projected reconstruction errors, here referred to as the scores, and the mutual coherence of dictionary terms. 
The first contribution of this work is a theoretical analysis of the score and dictionary-selection strategy. This could be understood in both the original and weak SINDy regime.
Second, numerical experiments on ordinary and partial differential equations highlight the effectiveness of score-based screening, improving both accuracy and interpretability in dynamical system identification.
These results suggest that integrating score-guided methods to refine the dictionary more accurately may help SINDy users in some cases to enhance their robustness for data-driven discovery of governing equations.
\end{abstract}

\textbf{MSCcodes}
37M10, 62J99, 65L09, 93B30

\textbf{keywords}
SINDy, STLS, Hard thresholding, Sparse Regression, System Identification, Equation Discovery, Backward Variable Selection, Sum of Squares

%
%
%
%
\section{Introduction}
\label{sec:1}
System identification from time-series data remains a longstanding and important challenge. Least squares problems have appeared in linear modeling \cite{Bjorck96}, especially in signal processing \cite{Mallat99} and statistics \cite{wainwright2019}. In this context, several seminal works have leveraged sparse regression for reconstruction, leading to the development of the method of compressed sensing \cite{Donoho06, Candes06, tropp2006} and sparse optimization \cite{Tibshirani96}.
Recently, data-driven approaches -- particularly those based on symbolic regression -- have received significant attention \cite{Bongard07, Schmidt09}. Within the framework of compressed sensing which typically consider underdetermined systems \cite{Donoho06, Candes06}, several system identification methods have been proposed \cite{Wang11, Naik12}.
In the context of sparse optimization for overdetermined systems, Sparse Identification of Nonlinear Dynamics (SINDy), a dictionary-based sparse regression method, has demonstrated numerical performance, particularly in applications to fluid flows \cite{Brunton16}, and is supported by convergence theory \cite{Zhang19}.
While Least Absolute Shrinkage and Selection Operator (LASSO) offers an approach to system identification, it results that errors occur early on the LASSO path \cite{Su17}, leading to false discoveries. 
Sequential Threshold Least Squares (STLS), the algorithm used in the papers \cite{Brunton16,Zhang19}, serves as an effective alternative to LASSO with improved performance \cite{Zheng18} and has become a common choice for SINDy implementations.

One challenge in dictionary-based learning is building or pruning the library, particularly for terms with small coefficients. Indeed, STLS may eliminate functionally important terms because their coefficients are small. For example, in a system undergoing a Hopf bifurcation, even a small coefficient may have a critical impact on the stability. Also, STLS focuses only on the $\argmin$, that is, on finding the coefficients that minimize the reconstruction error after thresholding. It does not address how these minimal errors relate to observable patterns in the sequence of scores across sparsity levels. This implies that selection of the threshold is crucial to the result, and users typically select it through trial and error.

\begin{figure}
    \centering
    \includegraphics[width=\linewidth]{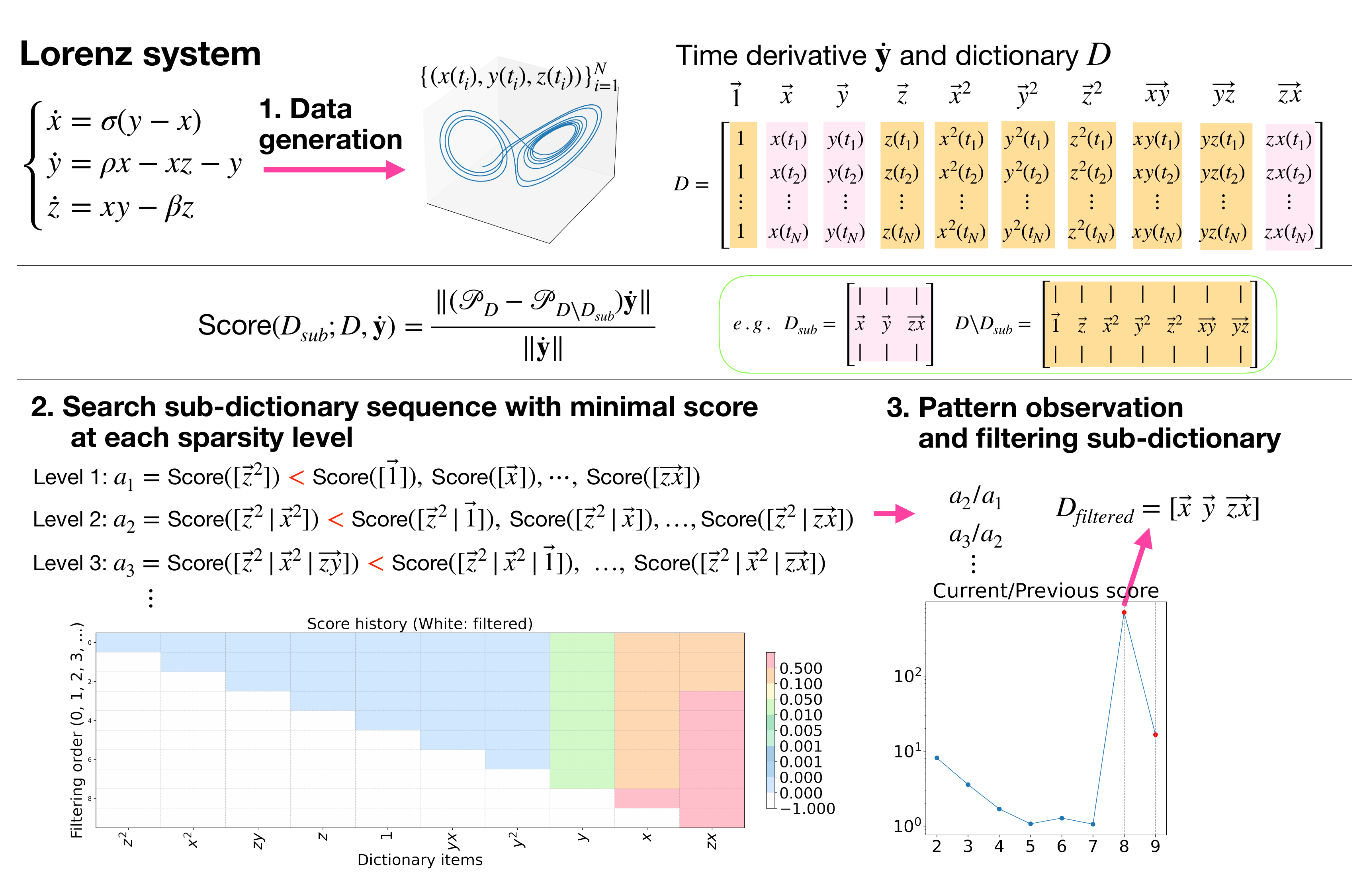}
    \caption{
    A schematic of the scoring procedure and its data pattern over a dataset generated by the Lorenz system, which includes the time history of the states $(x,y,z)$. At each stage, we compute the score of sub-dictionaries and select the one with the smallest score, starting with single items, then this procedure is repeated recursively with the inclusion of the previously selected items. Here, $\mathcal{P}_A$ is the projection map to the column space of a given matrix $A$. A sharp increase in minimal scores at $8$th step indicates that the set of the removed items may contain an important item. We filter out the first seven terms from the dictionary. A regular regression with the reduced dictionary yields the final model for $\dot{y}$. The parameters are $\sigma = 10$, $\rho = 26$, $\beta = 2.66667$, and the initial state is $(x(0),y(0),z(0))= (-8, 8, 27)$.
    }
    \label{fig:main}
\end{figure}

In this paper, we present a useful scoring-based tool for helping SINDy users. Through this tool, users may observe how low projected model reconstruction error can get for a given dictionary, indicating how well the dictionary explains the signal. Thus, this tool may be a pruning method to refine the dictionary more accurately. Moreover, our tool may prevent STLS from thresholding terms with small coefficients.
Indeed, we demonstrate the power and limitations of our tool theoretically and numerically. A schematic overview is in Figure~\ref{fig:main}.
One may observe clear empirical data patterns that help to determine an optimal number -- an optimal sparsity level -- of dictionary items. 

\subsection{Related work}
In statistics, the square of the score we will use is called a sum of squares \cite{Montgomery21}. Those quantities are generalized F-values and are used for model selection (See Section \ref{sec:3.1}).
Including those, the pruning variables problem has been extensively studied in the statistics literature, particularly in the context of overcomplete dictionaries. Our use case, however, is distinct. We consider a regime that is only mildly overcomplete and arises in a different context, where nonlinear correlations exist among the dictionary columns. These features make the problem different from the classical setting.

Our regressor resembles the Stepwise Sparse Regressor (SSR) \cite{Boninsegna18} with different criteria for eliminating one item at each iteration. In their paper, the cross-validation score, which is the average reconstruction error obtained through $k$-fold cross-validation, was used to select the optimal sparsity level automatically (See Section \ref{sec:stepwise_Sparse_Regressor}). Likewise, we observe the data pattern of scores, which represent errors in fully projected signals, and may set a sparsity level.

Beyond these, we also highlight recent developments in SINDy and related frameworks.
On recent progress in general, we refer to \cite{Ghadami22,Brunton22,Bramburger24}. Here we only review several related papers among lots of great developments in literature.

In Ref. \cite{KBKM24}, the authors replaced thresholding based on on the coefficient magnitude with $l_0$ sparsity penalized Bayesian inference it order to investigate how the identified models depend on sparsity, amount of data, and noise. A key observation was that the results produced by SINDy depend significantly on the choice of sparsity penalty and the identified models change discontinuously by adding or removing terms. To explain this phenomenon, the authors employed tools from statistical mechanics, introducing a notion of free energy associated with dictionary items. Based on this perspective, they proposed an alternative method called Z-SINDy.

The SINDy variants incorporate statistical tools for uncertainty quantification and model selection. Ensemble-SINDy (E-SINDy) leverages bagging and bootstrap techniques to improve robustness to noise \cite{FKBB22} with theoretical foundations \cite{GFBK23}. $k$-fold cross-validation has also been used to do automatic dictionary selection \cite{Boninsegna18}, and Bayesian-SINDy has been developed within a Bayesian framework \cite{fung25}.  Model selection in SINDy can be formalized with criteria based on information theory \cite{Mangan17}. Additional robustness on noisy data can be achieved by Simultaneous Identification and Denoising of Dynamical Systems (SIDDS) that introduces an auxiliary variable in the Least Squares Operator Inference (LSOI) problem \cite{Hokanson23}, and Derivative-based SINDy (D-SINDy) \cite{Wentz23} that uses a projection-based denoising approach.

Since we solve linear regression, some optimization strategies were used in Sparse Relaxed Regularized Regression (SR3) \cite{Zheng18, Champion20} and the Conditional gradient-based approach CINDy, which adopts an integral formulation of LASSO \cite{Carderera21}. Adam-SINDy employs Adam optimization for dictionary refinement \cite{Viknesh24}. Efforts to refine dictionary selection include SINDy with sensitivity analysis (SINDy-SA), which compares reconstructed coefficient vectors from least squares and ridge regression to select dictionary items \cite{Naozuka22}.
  
Extensions to parametric systems have also been developed \cite{Nicolaou23}, along with frameworks for systems with control inputs \cite{Fasel21}. Further methodological variations include SINDy-PI, which identifies implicit differential equations \cite{Kaheman20}, and SINDy-BVP, designed to handle boundary value problems \cite{Shea21}. Stochastic diffusions have also been considered on stochastic differential equations \cite{Wanner24, Jacobs24}. SINDy has been extended to discrete-time mappings such as the Poincaré map \cite{Bramburger20}, and to Reinforcement Learning contexts through SINDy-RL \cite{Zolman24}. Beyond regression-based improvements, SINDy has also been integrated with neural architectures, which detect intrinsic coordinates within an autoencoder framework \cite{Champion19}.

\subsection{Contributions and Paper Structure}

Our main contributions are the following:

\begin{enumerate}[(i)]
\item Based on the score, we propose dictionary selection algorithms: exhaustive version \eqref{eq:iteration} and greedy one \eqref{SSR_score}. We also provide a theoretical comparison between STLS and the score-based dictionary selection method.

\item To improve the results and performance, we apply the weak formulation, as described in \Cref{sec:pratical_implementation}. 

\item We implement noise robustness tests, a PDE example, and an unsupervised dataset to illustrate the advantages and limitations of our method, especially in cases where STLS fails to detect small coefficients. 
\end{enumerate}

{\bf Paper Structure.}
This paper is organized as follows.
Section \ref{sec:2} introduces notation and briefly reviews related methodologies.
Section \ref{sec:3} presents our proposed score-based method, derived from the coefficient formula of STLS.
Section \ref{app:theoretical-result} provides a theoretical result, which gives a detailed relationship between scoring and STLS.
In Section \ref{sec:4}, several numerical simulations show both the strengths and limitations of the proposed scoring methods.
Finally, Section \ref{sec:conclustion} summarizes our work and outlines future research directions.
For readability, all proofs are collected in Appendix \ref{app:proofs}.

Code for all of these examples can be found at \url{https://github.com/fabioamaral08/libselection}.

\section{Background}\label{sec:2}
In this section, we first summarize several notation to be used throughout the paper, and then review STLS, Pareto curves, Stepwise Sparse Regressor, Weak-SINDy and PDE-FIND.

\subsection{Notation}
We summarize the notation used throughout the paper, beginning with notation related to the dictionary.

Let $d_1,d_2,\ldots,d_n$ be given dictionary functions (often called {\it variables} in statistics or {\it dictionary elements} in signal processing). We denote the unordered set of these items by $\mathcal{D}=\{d_1,d_2,\ldots,d_n\}$, and the ordered tuple by $\vec{\mathcal{D}}=(d_1,d_2,\ldots,d_n)$. Given a time sequence $(t_j)_{j\in[m]}$, we define the evaluation vector corresponding to the function $d_i$ as $\mathbf{d}_i = \begin{bmatrix} d_i(t_1) & \cdots & d_i(t_m) \end{bmatrix}^{{\rm T}}$. We then construct the data matrix $D\in\mathbb{R}^{m\times n}$ whose $i$th column is $\mathbf{d}_i$, i.e., $D = \begin{bmatrix}\mathbf{d}_1 & \cdots & \mathbf{d}_n\end{bmatrix}$.

We now introduce general notation used throughout the paper.
For $N\in\mathbb{N}$, define $[N]\coloneqq \{1,\ldots, N\}$ and $[N]_0\coloneqq \{0\}\cup[N]$.
For a finite set $S$, $|S|$ denotes its cardinality.
For a vector $\mathbf{x}=(\mathbf{x}_1,\ldots,\mathbf{x}_d)\in\mathbb{R}^d$, $\ell^p$ ($p\geq0$) is defined as:
\[
\|\mathbf{x}\|_0 \coloneqq |\{i \in [d]: \mathbf{x}_i\neq 0\}|,\quad \|\mathbf{x}\|_p \coloneqq \sum_{i=1}^d |\mathbf{x}_i|^p,\quad p>0.
\]
The support of a vector $\mathbf{x}\in\mathbb{R}^d$ is defined as $\supp(\mathbf{x}) = \{i\in[d]: \mathbf{x}_i \neq 0\}$. A vector is called $k$-sparse when $|\supp(x)|\leq k$.

For a matrix $A = [\mathbf{a_1}|\cdots|\mathbf{a}_m]\in\mathbb{R}^{n\times m}$, let $A^\dagger$ denote its Moore-Penrose pseudoinverse. The notation $[\mathbf{a}_1|\cdots| \hat{\mathbf{a}}_i|\cdots|\mathbf{a}_m]$ denotes the matrix obtained by omitting the $i$th column.

For an index subset $S\subset [m]$ with $|S| = s$, a matrix $A\in\mathbb{R}^{n\times m}$, $A_S$ denotes the submatrix consisting of columns of $A$ indexed by $S$. Similarly, for $x\in\mathbb{R}^m$, $x_S\in\mathbb{R}^s$ denotes the subvector containing the components indexed by $S$.\footnote{For these notations, we refer to \cite{Zhang19}.} 

In $\mathbb{R}^d$, we use $\langle\cdot,\cdot\rangle$ and $\|\cdot\|$ to denote the standard inner product and Euclidean norm, respectively. That is, for $\mathbf{x}=(x_i)$, $\mathbf{y}=(y_i)\in\mathbb{R}^d$, $\langle \mathbf{x},\mathbf{y}\rangle = \sum_{i=1}^d x_iy_i$ and $\|\mathbf{x}\| =\sqrt{\langle \mathbf{x},\mathbf{x}\rangle}=\|\mathbf{x}\|_2$. For a matrix $A\in\mathbb{R}^{n\times m}$, its operator norm $\|A\|_{op}$ is defined as $\|A\|_{op}=\sup\{\|Ax\|/\|x\|:x\in\mathbb{R}^m,\|x\|=1\}$.

For a set $S$ of functions, $\spann(S)$ denotes the linear span of the elements of $S$. For a matrix $A$, $\spann(A)$ denotes its column space.
The orthogonal projection onto a function space $S$ is denoted by $\mathcal{P}_S$. For a matrix $A$, $\mathcal{P}_A$ denotes the orthogonal projection onto its column space, and for a nonzero vector $\mathbf{a}$, $\mathcal{P}_\mathbf{a}$ denotes the projection onto $\mathbf{a}$.

\subsection{Sequential Threshold Least Square}
For a given time series dataset $\mathbf{x}=[x(t_i)]_{1\leq i\leq m}\in\mathbb{R}^m$, we approximate an empirical time derivative as $y\approx [\dot{x}(t_i)]_{1\leq i\leq m}\in\mathbb{R}^m$, typically generated by a finite time difference methods. We prepare a set of dictionary functions $\{d_1,\ldots,d_n\}$, and consider the associated linear problem $Dc=y$ where $y\in\mathbb{R}^m$ and $D\in\mathbb{R}^{m\times n}$ with $m>n$.
We also assume the dictionary matrix, $D=[\mathbf{d}_1\vert\cdots\vert\mathbf{d}_n]$, has full-rank.
The algorithm STLS compares a given threshold $\lambda>0$ and coefficients of the projection vector of $y$ to the column space of $D$. 

$\bullet$ (STLS) The algorithm proceeds through the following iterative scheme with a thresholding parameter $\lambda>0$ \cite{Zhang19}:
\begin{align}
\begin{aligned}\label{SINDy}
    &\xi^0 = D^\dagger y,\\
    &S^i = \{j\in[N]: |\xi_j^i|\geq \lambda \},\quad i=0,1,\ldots,\\
    &\xi^{i+1} = \argmin_{\supp(\xi)\subset S^i} \| D\xi- y\|_2.
\end{aligned}
\end{align}
STLS generates a sequence of coefficients $(\xi^k)$ such that the cost function $\xi\mapsto \|y- D \xi^k\|_2^2 + \lambda^2 \|\xi^k\|_0$ decreases with respect to $k$. Moreover, the sequence converges to a local minimizer of the cost function \cite{Zhang19}. We now recall one basic lemma:
\begin{lemma}\label{L1.1}
    For each $i=1,\ldots, n$,
    \[
    |[D^\dagger y]_i| = \| (\mathcal{P}_D -\mathcal{P}_{D_{[n]\setminus\{i\}}})y \|_2/\|\mathbf{d}_i - \mathcal{P}_{D_{[n]\setminus\{i\}}}\mathbf{d}_i \|_2.
    \]
\end{lemma}
\begin{remark}\label{R1.1}
In the first iteration of STLS with the threshold $\lambda$, indices $i$ satisfying the following will vanish:
\begin{align}\label{A-3}
    \| (\mathcal{P}_D -\mathcal{P}_{D_{[n]\setminus\{i\}}})y \|_2 < \lambda\|\mathbf{d}_i - \mathcal{P}_{D_{[n]\setminus\{i\}}}\mathbf{d}_i \|_2.
\end{align}
\end{remark}
\noindent Remark \ref{R1.1} gives us an intuitive explanation of STLS. Let an one-dimensional finite time sequence $\mathbf{x}$ be given and let $y = f(\mathbf{x}) + g(\mathbf{x})$ for some linearly independent function $f$ and $g$. If we have the dictionary matrix $D=D(\mathbf{x}) = [f(\mathbf{x})|g(\mathbf{x})|h(\mathbf{x})]$ for some $h$, which linearly independent to $f$ and $g$, and this matrix has full-rank, it is natural to expect the following result
\begin{align*}
    \min_{\Xi\in \mathbb{R}^3} \| y - D\cdot\Xi\| = \begin{bmatrix} 1\\1\\0\end{bmatrix}.
\end{align*}
And also, we expect STLS with an appropriate threshold yields the same answer. Indeed, it does since $\mathcal{P}_{[f(\mathbf{x})|g(\mathbf{x})|h(\mathbf{x})]}y=\mathcal{P}_{[f(\mathbf{x})|g(\mathbf{x})]}y=f(\mathbf{x}) + g(\mathbf{x})$ and \eqref{A-3}. In general cases with a dictionary matrix $D = [f_1(\mathbf{x})|\cdots| f_n(\mathbf{x})]$, if $y\in\mbox{span}(D_{[n]\setminus\{j_0\}})$, $[D^\dagger y]_{j_0} =0<\lambda$ for any $\lambda>0$ and the index $j_0$ will not survive in the first iteration of STLS.

In these examples, the quantities of the form $\|(\mathcal{P}_{D} -\mathcal{P}_{D_{[n]\setminus\{i\}}})y\|$ in \eqref{A-3} play important role. We refer to these as scores and provide their definition in Section \ref{sec:3.1}. One may wonder the relationship between the model reconstruction error to these quantities, which are distinguishable when the dictionary doesn't fully explain the signal:
\begin{align}\label{eq:comparison-error}
\| y - \mathcal{P}_{D_{[n]\setminus\{i\}}} y\|_2^2 - \|  \mathcal{P}_{D_{[n]\setminus\{i\}}} y - \mathcal{P}_D y\|_2^2 = \| y - \mathcal{P}_D y\|_2^2>0.
\end{align}

When analyzing the STLS algorithm, we focus only on the first step because the same pattern emerges regardless of which dictionary terms are eliminated - whether one or multiple. This implies that at each iteration, the algorithm repeatedly faces the same problem of solving a least squares system using the pseudoinverse, as described in the following lemma:
\begin{lemma}[\cite{Foucart13} Lemma 3.4]
For an index set $S\subset [n]$, if 
$$\xi = \argmin_{\substack{ z\in\mathbb{C}\\ \supp(z)\subset S}}\|y-Dz\|_2,$$
then
\[
(D^*(y-D\xi))_S=\mathbf{0}.
\]
Thus, $\xi = D_S^\dagger y$.
\end{lemma}

\subsection{Pareto curve}
A Pareto curve is a graph of one-norm $\tau$ versus corresponding residual for solution to LASSO \cite{Tibshirani96} with parameter $\tau$ \cite{Foucart13}. Precisely, this is represented by the function $\phi:\mathbb{R}_{\geq0}\to\mathbb{R}_{\geq0}$ defined as:
\begin{align*}
\phi(\tau) = \| y - Dc_\tau \|_2,\quad
c_\tau = \argmin_{\substack{x \\ \|x\|_1 \leq \tau }} \|y - Dx\|_2,\quad \tau\geq0.
\end{align*}
This curve is convex and decreasing, and so connects a basis pursuit denoise problem to a LASSO problem \cite{Van09}.

In our setting, we consider an $\ell^0$-analogue, which is discrete rather than continuous, obtained by recording the residual relative to the projected target vector at each sparsity level.

\subsection{Stepwise Sparse Regressor}\label{sec:stepwise_Sparse_Regressor}
The Stepwise Sparse Regressor (SSR) iteratively removes one dictionary element at each iteration and employs cross validation to determine the optimal sparsity level \cite{Boninsegna18}. Without the cross validation part, the algorithm is as follows:
\begin{align}
\begin{aligned}\label{SSR}
    &\xi^0 = D^\dagger y,\quad S^0 = \emptyset,\\
    &S^{i+1} = S^{i}\cup\big\{\argmin_{j\in[n]\setminus S^{i}}|\xi^{i}_j| \big\},\quad i=0,1,\ldots,\\
    &\xi^{i+1} = \argmin_{\supp(\xi)\subset S^{i+1}} \| D\xi- y\|_2.
\end{aligned}
\end{align}
One typically observes a data pattern on the cross-validation scores, with a drastic increase at the optimal sparsity level.
SSR has been shown to be effective in pruning unimportant dictionary elements from a large dictionary when applied to (stochastic) ordinary differential equations.

As in \cite{Boninsegna18}, we denote the coefficient vector by $\xi^i \eqqcolon \SSR(D,\xi)_i$.
The definition of the cross-validation score is provided in Section~\ref{sec:4}, where it is compared with our proposed score.

\subsection{Weak-SINDy}\label{app2-W}
Weak formulations of dynamical systems makes a Galerkin-based model selection method \cite{Messenger21,Russo24,Russo25,Messenger24}.
Let a time-series dataset $\{U_i\}_{0\leq i\leq n}$ with a time sequence $\{t_i\}_{0\leq i\leq n}$ be given. We assume that the dataset satisfies $U_i=x(t_i)$ for a solution $u$ of the ODE $\dot x = f(x(t))$ where $f$ is a continuously smooth vector field.
If $\phi$ is a real-valued test function which has a (connected) compact support, denoting a (time) interval $(a,b)$, in the real-line. By the integration by parts, we have an equality:
\[
\int_a^b \big( \phi'(t) x(t) + \phi(t) f(x(t))\big) dt =0.
\]
This equality gives us a different discretization to identify a system.
For a time sequence $(t_i)$ such that $t_0\leq a<b\leq t_n$, the integral equity implies
\[
\sum_{i=0}^n \phi'(t_i) x(t_i) + \phi(t_i) f(x(t_i)) \approx 0.
\]
Now if we have multiple test functions $\phi_1,\ldots,\phi_k$ and $f$ is of the form $Dx$ with a dictionary matrix $D$ and a coefficient vector $x$, then we can make a least square problem to minimize $\|Gx - b\|_2$ where $G=\Phi D$ and $b=-\Phi x$.

Designing test function is essential, so we follow the selection in Ref. \cite{Messenger21}: $\phi_i(t)$ has of the form $C(t-a)^p(b-t)^q$.
Therefore, hyperparameters are the degree of the polynomial, $p$ and $q$, the number of test functions, $k$, and the length of compact support for each test function $b-a$.

For a weighted least square formulation with a regularizer, we refer to the paper \cite{Messenger21}.
In this paper, we only adopt the simplified form.

\subsection{PDE-FIND}
Sparse identification in symbolic regression can be applied to datasets generated by partial differential equations \cite{Rudy17}. Dictionaries typically are composed with time and spatial derivative. PDE version of weak SINDy also was developed \cite{Messenger21-pde}. Here we describe a simplified version of them, following notation in the paper \cite{Messenger21-pde}.

Our spatiotemporal ambient space is $\mathbb{R}^1\times\mathbb{R}^d$. For an open bounded subset $\Omega$ in $\mathbb{R}^d$, let a spatiotemporal dataset $\{U_i\}_{0\leq i\leq n}$ be given on the spatial grid $X\subset \bar{\Omega}$ over a time sequence $\{t_i\}_{0\leq i\leq n}\subset [0,T]$ . We assume that the dataset satisfies $U_i=u(X,t_i)$ for a solution $u$ of the PDE
\[
\frac{\partial}{\partial t} u(x,t) = \mathfrak{D}^{\alpha^1}g_1(u(x,t)) + \mathfrak{D}^{\alpha^2}g_2(u(x,t))+\cdots + \mathfrak{D}^{\alpha^S}g_S(u(x,t)),\quad x\in\Omega,~t\in(0,T).
\]
for some set of unknown true functions $\{g_i\}_{i \in [S]}$.
Here, we used the multi-index notation to write the partial differentiation:
\[
\alpha^i = (\alpha^i_1,\ldots, \alpha^i_{d+1})\in\mathbb{N}^{d+1},\quad
\mathfrak{D}^{\alpha^s}u(x,t) =\frac{\partial^{\alpha^s_1+\cdots + \alpha^s_{d+1}}}{\partial x_1^{\alpha^s_1}\cdots \partial x_d^{\alpha^s_d} \partial t^{\alpha^s_{d+1}}}u(x,t).
\]
Our goal is to identify this differential equation using a dictionary whose elements are spatiotemporal derivatives of a family of functions $(f_j)_{j\in[m]}$ (called as the trial functions). Again we assume that $\{g_i\} \subset \spann_{j\in[m]} (f_j)$ so that the PDE can be rewritten by
\begin{align}\label{eq:PDE-SINDy}
\frac{\partial}{\partial t} u(x,t) = \sum_{s=1}^S \sum_{j=1}^m w_{(s-1)m+j}\mathfrak{D}^{\alpha^s} f_j(u),
\end{align}
where the vector $w\in\mathbb{R}^{mS}$ is the coefficients, which is assume to be sparse. Flattening the snapshots $U_i$ at each time $t_i$ gives us a setting for a sparse linear regression, namely PDE-FIND \cite{Rudy17}.

On the other hand, with a smooth test function $\psi(x,t)\in {\rm L}^2(\Omega)$ which compactly supported in $\Omega\times (0,T)$, one may get the weak formulation of the dynamics \eqref{eq:PDE-SINDy} as follows:
\[
-\left\langle \frac{\partial}{\partial t}\psi, u\right\rangle_{ {\rm L}^2(\Omega)} = \sum_{s=1}^S\sum_{j=1}^m \left\langle (-1)^{|\alpha^s|}\mathfrak{D}^{\alpha^s}\psi, f_j(u)\right\rangle_{{\rm L}^2(\Omega)}.
\]
Likewise the ODE version, we use multiple test functions $\psi_1,\ldots,\psi_k$ to deduce a linear system of the form $b = Gw$ where $b=(b_1,\ldots,b_k)\in\mathbb{R}^k$, $G=(G_{i,j})\in\mathbb{R}^{k\times mS}$, $b_\ell = -\left\langle \frac{\partial}{\partial t}\psi_\ell, u\right\rangle_{ {\rm L}^2(\Omega)}$, $G_{\ell,j} = \langle (-1)^{|\alpha^s|}\mathfrak{D}^{\alpha^s}\psi_\ell,f_j(U)\rangle_{{\rm L}^2(\Omega)}$. Weak SINDy for PDE works over this setting \cite{Messenger21-pde}. In particular, the authors took $k$ translated functions over one reference test function $\psi$; $\psi(x,t)= \psi(x-x_\ell,t-t_\ell)$, so that the inner products could be rewritten in the form of convolutions. Leveraging the Fourier transform and its property, one may reach out a reduced linear system again. Solving the least square problem over the given dataset to find the coefficient vector effectively works for several examples \cite{Messenger21-pde}.

\section{Scoring-Based Dictionary Pruning Method}\label{sec:3}
In this section, we introduce scores and our dictionary-selection method.

Let $y\in\mathbb{R}^m$ be a given target vector, and we seek to construct a model for $y$ using a dictionary matrix $D=D(\mathbf{x}) =[\mathbf{d}_1(\mathbf{x})|\cdots|\mathbf{d}_n(\mathbf{x})]\in\mathbb{R}^{m\times n}$ whose columns are generated by the dictionary items $d_i\in{\rm L}^2(\mathbb{R}^d;\mathbb{R})={\rm L}^2$. Here, the space ${\rm L}^2$ is equipped with the empirical measure $\sum_{i=1}^n\delta_{\mathbf{x}_i}$ and $\delta_{\mathbf{x}_i}$ is the Dirac measure.
Our prior assumptions for learning dynamics are dealt with in Appendix \ref{sec:App1}.

\subsection{Score}\label{sec:3.1}
We define the {\it (projected) score} of each dictionary item $d_i$ as follows:
\begin{align*}
    \mbox{Score}(\mathbf{d}_i;D,y) 
    &= \frac{\left\|(\mathcal{P}_{D} - \mathcal{P}_{D_{[n]\setminus\{ i\}}}) y\right\|_{{\rm L}^2}}{\|y\|_{{\rm L}^2}}.
\end{align*}
Since we will compare such scores, we impose normalization by their denominator, which similar to the concept of the fraction of variance unexplained.
If $y$ were in $\mbox{span}(D_{[n]\setminus\{i\}})$, then the score of $d_i$ would be zero. 
This score quantifies how informative a given dictionary item is relative to the entire dictionary $\mathcal{D}$. As we have seen in Remark \ref{R1.1}, the STLS algorithm eliminates indices corresponding to low-score items. Thus, the smallness of the score assigned to a dictionary function can be interpreted as a sign for its potential removal in STLS. 

We regard a dictionary item as non-informative to explain the signal when it has a small score relative to others. On the other hand, it is well-known that significant variables appear uncorrelated until conditioned on other variables. Thus, pruning dictionary items based on the numerator in the score formula may not help explain the signal. One natural extension of the formula from the proximity of the partially projected signal to the fully projected signal onto the dictionary matrix may tackle this matter since it considers all possible combinations of sub-dictionaries. Again, we consider a sub-dictionary non-informative for explaining the signal when it has a small (generalized) score relative to others.

Let $D_{sub}=D_{sub}(\mathbf{x})$ be a submatrix of $D$ whose columns are selected from $D$. Denote the remaining columns, i.e., the (column) complement of $D_{sub}$, by $D\setminus D_{sub}$. The score of $D_{sub}$ given $D$ is defined as:
\begin{align*}
    \mbox{Score}(D_{sub};D,y)
    &= \frac{\left\|(\mathcal{P}_{D} - \mathcal{P}_{D\setminus D_{sub}}) y\right\|_{{\rm L}^2}}{\|y\|_{{\rm L}^2}}.
\end{align*}
This score quantifies how informative the sparse sub-dictionary $\mathcal{D}_{sub}$ is relative to the entire dictionary $\mathcal{D}$. We now list several basic properties of the scores.
\begin{lemma}\label{L1.3}
The scores lie in the interval $[0,1]$:
\[
0\leq \score (D_{sub};D,y)
\leq 1
\]
\end{lemma}

One relationship between the score sequence and the sub-dictionary score is as follows:
we fix a permutation $\sigma:[n]\to[n]$ and a sub-dictionary $\mathcal{D}_{sub}=\{d_{\sigma(1)},\ldots,d_{\sigma(\ell)}\}$ $\subset\mathcal{D}$ satisfying $|\mathcal{D}_{sub}|=\ell\leq n$. Let $\mathbf{h}_1,\ldots,\mathbf{h}_n$ be the orthonormalization of $\mathbf{d}_{\sigma(n)},\ldots$, $\mathbf{d}_{\sigma(2)},\mathbf{d}_{\sigma(1)}$. Then, we have
\begin{align}\label{S-3}
\begin{aligned}
&\| (\mathcal{P}_{D} - \mathcal{P}_{D_{sub}})y\|_2^2
= |\langle \mathbf{h}_{n-\ell+1},y\rangle|^2+\cdots+|\langle \mathbf{h}_n,y\rangle|^2\\
&= \|y\|_2^2\Big((\score(\mathbf{d}_{\sigma(\ell)};D_{[n]\setminus\{\sigma(1),\ldots,\sigma(\ell-1)\}},y  ))^2+\cdots+(\score(\mathbf{d}_{\sigma(1)};D,y))^2 \Big).
\end{aligned}
\end{align}
Thus, we have
\begin{align}\label{S-4}
\begin{aligned}
&\mbox{Score}(D_{sub};D,y)^2\\
&\overset{\eqref{S-3}}{=} (\score(\mathbf{d}_{\sigma(\ell)};D_{[n]\setminus\{\sigma(1),\ldots,\sigma(\ell-1)\}},y ))^2+\cdots+(\score(\mathbf{d}_{\sigma(1)};D,y))^2.
\end{aligned}
\end{align}
One trivial inequality is as follows:
\begin{align}\label{S-5}
    \mbox{Score}(\mathbf{d}_i;D,y) < \mbox{Score}(D_{sub};D,y),\quad \forall~d_i\in\mathcal{D}_{sub}.
\end{align}

This score is related to the extra-sum-of-squares method in statistic.
Here, we refer to \cite{Montgomery21}. For a model of the form $y = D\beta +\varepsilon$, $D\in\mathbb{R}^{m \times n}$, the regression sum of squares of the this (full) model is given by
\[
\mbox{SS}_R(\beta) \coloneqq (D^\dagger y)^TD^Ty = y^T \mathcal{P}_D y = \|\mathcal{P}_D y\|_2^2.
\]
For a partition $\beta = [\beta_1^T ~|~ \beta_2^T]^T$ with $\beta_1\in\mathbb{R}^{n-r},\beta_2\in\mathbb{R}^r$, we have the reduced model $y=D_{sub}\beta_1 + \varepsilon$. In this case, the regression sum of squares for this (reduced) model is given by
\[
\mbox{SS}_R(\beta_1) = (D_{sub}^\dagger y)^TD_{sub}^Ty = \|\mathcal{P}_{D_{sub}}y\|^2_2
\]
so that\footnote{These quantities lead the statistic $F_0$ given by
\[
F_0 \coloneqq \frac{\mbox{SS}_R(\beta_2 |\beta_1)/r}{\mbox{MS}_{Res}},\quad \mbox{MS}_{Res}\coloneqq \frac{y^Ty - \mbox{SS}_R(\beta)}{m-n}=\frac{\|y-\mathcal{P}_D y\|_2^2}{m-n}.
\]
This quantity effectively used for variable selections in statistics. Also, it is known that when $y$ follows a Gaussian distribution, each $\mbox{SS}_R$ follows a $\chi^2$ distribution (e.g. \cite{Montgomery21} Appendix C.3.4).  However, this statistical context is beyond our scope.}
\[
\mbox{SS}_R(\beta_2|\beta_1) \coloneqq \mbox{SS}_R(\beta) - \mbox{SS}_R(\beta_1) = \|\mathcal{P}_Dy - \mathcal{P}_{D_{sub}}y\|_2^2.
\]

\subsection{Dictionary selection}\label{sec:3.2.3}
In this section, we introduce two stepwise regressors, which filter non-important terms based on scores: one exhaustive and the other reductive. The exhaustive one is analogous to drawing a Pareto curve, which plots the score (serving a role similar to the reconstruction error) against the $\ell^0$ norm of the coefficient vector. In particular, it can be interpreted as the $\ell^0$ counterpart of the basis pursuit denoise problem:
\begin{align}\label{eq:minmax}
\begin{aligned}
\max_{\substack{\ell\in[m] \\ |a_\ell|<\varepsilon}} ~\ell\qquad\mbox{where}\quad a_\ell = \min_{\substack{\mathcal{D}_{sub}\subset \mathcal{D} \\ |\mathcal{D}_{sub}|=\ell}} \score(D_{sub};D,y).
\end{aligned}
\end{align}
Here, the admissible tolerance parameter $\varepsilon>0$ determines the sparsity level for model construction, but it will not be considered in the following iterative schemes of this section; instead, its value will be determined empirically in the numerical section.

\subsubsection{Exhaustive Stepwise Regressor}
We propose a sequential procedure to identify an optimal sub-dictionary from a given dictionary: Under the same setting in Subsection \ref{sec:3.1}, we seek a sequence of sub-dictionaries $\{\mathcal{D}_i\}$ satisfying, for $i=1,\ldots,m$,
\begin{align}\label{eq:iteration}\tag{ESR}
\begin{aligned}
&\mathcal{D}_0 = \mathcal{D},\quad
\mathcal{D}_{i} = \mathcal{D}\setminus \mathcal{A}_i, \quad 
\mathcal{A}_i := \argmin_{\substack{\mathcal{D}_{sub}\subset\mathcal{D} \\ |\mathcal{D}_{sub}|=i}}\score(D_{sub};D,y),\quad
a_i := \score(A_i;D,y).
\end{aligned}
\end{align}
Here, $\mathcal{A}_i$ might not be uniquely determined, especially when a score value is zero.
We select a dictionary $\mathcal{D}_i$ among the sequence for SINDy where the associated score $a_i$ is small.
In this paper, we refer to this approach as an exhaustive stepwise regressor.
We note that, unlike this search, which reduces the sparsity level incrementally, STLS eliminates multiple terms at once.

Discarding low-scoring items gives an error bound for the trajectory reconstruction (See Appendix \ref{app:R}).
This scoring could be energy-like formulated. We refer to Z-SINDy in Appendix \ref{sec:tradeoff} and \ref{app2-Z}.
On the other hand, rather than discarding one by one, one may weigh each coefficient along its projection-based reconstruction weight. We refer to D-SINDy in \Cref{app2-D}.

\subsubsection{Greedy Backward Stepwise Regressor}
Computing all possible sub-dictionary scores is computationally expensive. Thus, we propose an inclusive stepwise regressor, which is computationally cheaper. For a given dictionary matrix $D=[\mathbf{d}_1|\cdots|\mathbf{d}_m]$ and target vector $y$, we define an algorithm that iteratively computes a score for each dictionary element, removes the one with the lowest score, and updates the dictionary.
This procedure is formally defined as follows:
\begin{align}\label{SSR_score}\tag{GBSR}
\begin{aligned}
&{j^0}=\argmin_{i\in[m]} \mbox{Score}(\mathbf{d}_i;D,y),\quad J^0=[m]\setminus\{j^0\},\\ 
&{j^i}=\argmin_{\ell\not\in J^{i-1}} \mbox{Score}(D_{\{ j^1,\ldots, j^{i-1}\}\cup\{ \ell\}};D,y),\quad J^i = J^{i-1}\setminus\{j^i\},\quad i=1,\ldots, n. 
\end{aligned}
\end{align}
We denote the counterpart of the sequence $(a_i)$ defined in the previous section by $(b_i)$:
\begin{equation}\label{eq:GBSR_score}
b_i \coloneqq \score(J^i;D,y),\quad i=0,1,\ldots, n.
\end{equation}
One may see this way a backward variable selection method based on the sum of squares regression in statistic. Also, our way reminds us of the SSR introduced in Ref. \cite{Boninsegna18}.
On the other hand, we visit the Orthogonal Matching Pursuit (OMP) using the score in Appendix \cref{app2-OMP}, which is a well-known greedy search for signals.

Note that this algorithm \eqref{SSR_score} may not yield the minimal sub-dictionary score, i.e. possibly
\[
\score(\{\mathbf{d}_{j^0},\ldots,\mathbf{d}_{j^{\ell-1}}\};D,y)\neq \min_{\substack{J\subset [n] \\ |J|=\ell }}\score(D_J;D,y).
\]
However, when $y\in \spann(D_{sub})\subset\spann(D)$, the procedure \ref{SSR_score} arrives at a sub-dictionary that achieves reasonably low score while maintaining maximal sparsity. Assume the case and let $|\mathcal{D}_{sub}|=k<n=|\mathcal{D}|$, then
\[
\score(D_\mathcal{S} ; D,y)=0,\quad \forall~\mathcal{S}\subset \mathcal{D}\setminus\mathcal{D}_{sub}.
\]
And $\mathcal{A}_{n-k}=\mathcal{D}\setminus\mathcal{D}_{sub}$. Indeed, the discussion in this paragraph can be formalized as Theorem \ref{P1.3}.

\begin{remark}\label{rmk:GFSR}
One might wonder the forward one; adding dictionary items one-by-one. We call this Greedy Forward Stepwise Regressor (GFSR). In our numerical simulation, the performance of GFSR is worse than GBSR. See Appendix \ref{app:GFSR}.
\end{remark}

\subsection{Practical Implementation}\label{sec:pratical_implementation}
Since our discussion is just on linear regression, our methods could be combined with different methods, e.g. E-SINDy or weak-SINDy.
Especially, in this paper, we focus on the weak formulation.
The weak-SINDy \cite{Messenger21} provides a different formulation of regression for system identification.
Although this demands additional hyperparameters, there are advantages on computational cost and noise robustness, so we adopt this as our default method in numerical tests section.

Empirically, combining all coordinates in search rather than one coordinate search is more useful.
We summarize this combination in the pseudo-code (Procedure \ref{algorithm}).
\begin{algorithm}
\caption{Library Selection Algorithm}\label{algorithm}
\begin{algorithmic}[1]
\State \textbf{Input:} Time sequence $\mathbf{t}=(t_i)$, Data $\mathbf{x}=(\mathbf{x}(t_i))_i$, dictionary $D=D(\mathbf{x})$, test function matrix $\Phi=\Phi(\mathbf{t})$, time-derivative of test function $\Phi'=\Phi'(\mathbf{t})$
\State \textbf{Set:} $\dot{\mathbf{x}} \gets $time derivative approximation
\If{Use Weak Form}
\State Compute transformed response: $\mathbf{y} \gets \Phi' \mathbf{y}$, $D \gets D\Phi$ 
\Comment{Weak formulation}
\EndIf
\State \textbf{Score type:} Choose type A (One coordinate) or B (All coordinate)
\State \textbf{Set coordinate:} $j$, $y \gets \dot{\mathbf{x}}[:,\,j]$
\If{Exhaustive Stepwise Regressor}
    \For{each level $i=1,2,\ldots$,}
        \State $\mathcal{A}_i \gets \argmin_{\substack{|{\mathcal{D}}_{sub}|=i \\ \mathcal{D}_{sub}\subset \mathcal{D} }} \score(D_{sub}; D,y) $ \Comment{Refer to \eqref{eq:iteration}}
        \State $a_i \gets \score(\mathcal{A}_i ; D,y)$
        \State Apply STLS with $D\setminus A$
    \EndFor
\ElsIf{Greedy Backward Stepwise Regressor}
    \State \textbf{Set:} $\mathcal{A}_0=\emptyset$
    \For{each level $i=1,2,\ldots$,}
        \State $ d_i \gets \argmin_{d_i \in \mathcal{D} \setminus \mathcal{A}_{i-1}} \score(\mathcal{A}_{i-1} \cup \{d_i\}; D,y) $ 
        \Comment{Refer to \eqref{SSR_score}}
        \State $\mathcal{A}_i \gets \mathcal{A}_{i-1} \cup \{d_i\}$
        \State $a_i \gets \score(\mathcal{A}_i ; D,y)$
    \EndFor
\EndIf
\State \textbf{Output:} $\{a_i\}$, $\{\mathcal{A}_i\}$
\end{algorithmic}
\end{algorithm}

\section{Theoretical result: Score and STLS}\label{app:theoretical-result}
In this section, we study mathematical properties of our iterative schemes. Especially, we reveal relationships to STLS.

\subsection{\texorpdfstring{$\ell^0$} --minimization}\label{sec:l0-minimization}
It is well-known that the SINDy algorithm makes a sequence which decreases the objective function $x\mapsto \|Dx-y\|_2^2+\alpha^2\|x\|_0$ \cite{Zhang19}. We could make a similar statement. For a given target vector $y$ and a dictionary matrix $D$, let $\{\xi^i\}_{i=0}$ be a sequence generated by the greedy searching. If one score of an item has small enough, say $\mbox{Score}(d_i;D,y)< \frac{\varepsilon}{\|y\|_2}$ and $d\alpha > \varepsilon$ for some threshold $\alpha$, then
\begin{align*}
\| D \xi^{i} - y\|_2 + \alpha \|\xi\|_0
&> \| D_{[n]\setminus\{i\}} \xi^{i+1} - y\|_2 - \varepsilon + \alpha \|\xi\|_0\\
&= \| D_{[n]\setminus\{i\}} \xi^{i+1} - y\|_2 + \alpha \|\xi^{i+1}\|_0 + (d\alpha -\varepsilon)\\
&> \| D_{[n]\setminus\{i\}} \xi^{i+1} - y\|_2 + \alpha \|\xi^{i+1}\|_0.
\end{align*}
Generally, if one takes the maximum absolute value of coefficients, whose items are discarded, as the parameter $\alpha$, then the objective function decreases. This is summarized as the following statement without a proof, which is a direct result from \cite{Zhang19}:
\begin{proposition}\label{thm_l0min}
Suppose a long-thin matrix $D=[\mathbf{d}_1|\cdots |\mathbf{d}_n]\in\mathbb{R}^{m\times n}$, which has full-rank and satisfies $\|D\|_{op}=1$, is given. Let $y\in\mathbb{R}^m$ be a target vector.
Let the sequence $\{s^i\}_{i=0}^{i=k}$ and $\{S^i\}_{i=0}^{i=k}$ be generated by $k$th iteration of the algorithm \eqref{SSR_score}.
Then, the sequence $\{x^i\}_{i=0}^{i=k+1}$ generated by the iteration \eqref{T-5} decreases the map $F(x) = \|Dx-y\|_2^2 +\alpha^2\|x\|_0$ where $\alpha = \max_{i\in[k]_0} |x^i_{s^i}|$.
\begin{align}\label{T-5}
x^0=D^\dagger y,\quad x^{i+1} = \argmin_{\supp(x)\subset S^i} \|Dx - y\|_2, \quad i=0,\ldots,k.
\end{align}
\end{proposition}

\subsection{Data pattern on a supervised dataset: ideal case}\label{sec:data-pattern}
We will see a pattern in the sequences $(a_i)$ and $(b_i)$ defined in Section \ref{sec:3.2.3}; once a sub-dictionary loses an important item then the corresponding score drastically increases. 
Based on this observation, we may choose an empirical sparsity level and perform linear regression for model selection.
In this case, in terms of the problem \eqref{eq:minmax}, the admissible error bound $\varepsilon$ may be automatically selected.
However, if there is no combination composed with a given dictionary to explain the target vector, then it would be hard to make a stop criterion. In this case, we design a preprocessing step for SINDy that closely aligns with STLS (See Proposition \ref{P1.2}), thereby allowing a controlled adjustment of the solution path.

In an ideal setting, we would observe a clear data pattern that separates important from non-important items when the iteration scheme~\eqref{eq:iteration} is applied.
This observation provides an intuitive justification for why our preprocessing method is effective in identifying the underlying system.

For a system $\dot{x}(t) = F(x(t)) = \sum_{i=1}^kc_id_i(x(t))$ and $\Pi_ic_i\neq0$, set $\mathcal{D}_{sub}=\{d_1,\ldots,d_k\}\subset \mathcal{D}=\{d_1,\ldots,d_n\}$ with $n > k$. For a time sequence $(t_i)$, let $\mathbf{x}_i = x(t_i)$ and $\mathbf{d}_i = [d_i(t_j)]_j$. Let $y_i=F(\mathbf{x}_i)$. In this ideal situation, a sequence $a_i$ generated by \eqref{eq:iteration} will be of the form $a_i=0$ for $i=1,\ldots,n-k$ but $a_i\neq 0$ for $i>n-k$. So the best way to pick up a dictionary is to select $\mathcal{D}_{i_0}$ where $a_{k}=0$ for all $k\leq i_0$. 

\vspace{0.1cm}
\noindent $\bullet$
When $i< n-k$, any subset $\mathcal{A}\subset\mathcal{D}\setminus\mathcal{D}_{sub}$ satisfying $|\mathcal{A}|=i$ implies $\score(A_i;D)=0$. Obviously, there are multiple candidates for $\mathcal{A}_i$ for each $i<n-k$. However, any sub-dictionary $\mathcal{D}\setminus\mathcal{A}$ may generate the same result as STLS.

\vspace{0.1cm}
\noindent $\bullet$
When $i=n-k$, $\mathcal{A}_{n-k}=\mathcal{D}\setminus\mathcal{D}_{sub} = \cup_{i=1}^{n-k}\mathcal{A}_i$. It is because $\score(A_{n-k};D)=0$.

\vspace{0.1cm}
\noindent $\bullet$ $i=n-k+1$ case is a little different since any candidate of $\mathcal{A}_{n-k+1}$ must contain, at least, one important dictionary item due to the Pigeonhole principle. With one important item $d_{i_0}$ $(i_0\leq k)$,
\[
\|\mathcal{P}_{D}y - \mathcal{P}_{D_{sub}\setminus d_{i_0}}y \|_2 
= |c_{i_0}|\| (\mathcal{I} - \mathcal{P}_{D_{sub}}){\mathbf d}_{i_0}\|_2.
\]
A lower bound of the score of $D_{sub}$ could be derived from the mutual incoherence between dictionary items: there exists a positive number $m$ such that for any index $j$,
\[
\| \mathcal{P}_{S} \mathbf{d}_j \|_2 <m < \|\mathbf{d}_j\|_2,\quad \forall\mathcal{S}=\{d_1,\ldots,\hat{d_j},\ldots,d_n\}.
\]

\vspace{0.1cm}

\subsection{Alignment to STLS}
Since the ground-truth dynamics is generally unknown for a given time series dataset, we use STLS as a reference criterion and examine when the results produced by the iterative schemes~\eqref{eq:iteration} and~\eqref{SSR_score} align with those of STLS.
While the alignment holds under strong assumptions, numerical experiments suggest that the behavior persists beyond this regime.

The following proposition demonstrates that if a dictionary item survives the first iteration of STLS, then any sub-dictionary containing this item will not yield an exceptionally low score.
Therefore, a score-based preprocessing step for STLS, which selects sub-dictionaries with low scores, ideally preserves the outcome of the original STLS procedure.
\begin{proposition}\label{P1.1}
Suppose for all $i=1,\ldots,n$, there exists a constant $\omega>0$ such that 
\begin{align}\label{S-6}
    \omega < \|\mathbf{d}_i - \mathcal{P}_{D_{[n]\setminus\{i\}}}\mathbf{d}_i \|_2,\quad i=1,\ldots,n.
\end{align} 
If the score of a sub-dictionary matrix $D_{sub}$ with $\mathcal{D}_{sub}\subset\mathcal{D}$ is sufficiently small with respect to a threshold $\lambda>0$, then the first iteration of STLS with the threshold deletes the sub-dictionary items.
\end{proposition}

When the coefficients from the first iteration of STLS are clearly separated into two groups by the threshold with a large difference between them, the sub-dictionary sequence generated by \eqref{eq:iteration} converges to the same result as STLS.
\begin{proposition}\label{P1.2}
Suppose that all dictionary items are normalized; $\|\mathbf{d}_i\|_2=1$ for all $i=1,\ldots,n$.
Assume a dictionary $\mathcal{D}=\{d_1,\ldots, d_n\}$ satisfies the followings:
\begin{enumerate}
    \item $\mathcal{P}_D y = \sum_{i=1}^n c_i\mathbf{d}_i = D \vec{c}$, $y = \mathcal{P}_D y + \vec{e}$, $\|\vec{e}\|\leq \varepsilon$,
    \item $S^0 = \{ i\in[n]:|c_i|\geq \lambda_1\}$, $|S^0| = k<n$, $[n]\setminus S^0 = \{ i\in[n]:|c_i|\leq \lambda_2\}$, $|S^0| = k<n$,
    \item For any $T\subset[n]$ satisfying $|T|=n-k,~|T\cap S^0|=\ell \leq k$,
    \begin{align}\label{S-11}
        (2n-2k-\ell)\lambda_2 + 2\varepsilon \leq \ell \sqrt{R(G,\vec{c}_{T\cap S^0})}\lambda_1
    \end{align}
    where ${\rm R}(G,\vec{c}_{T\cap S^0})$ is the Rayleigh quotient for the matrix $G=D_{T\cap S^0}^T(I-P_{D_{[n]\setminus T}})D_{T\cap S^0}$ and the vector $\vec{c}_{T\cap S^0}$.
\end{enumerate}
Then, $\mathcal{D}_{n-k}=S^0$. Here, $\mathcal{D}_{n-k}$ is defined in \eqref{eq:iteration}.
\end{proposition}
\begin{remark}
 One example of a dictionary satisfying this assumption in this proposition is when the dictionary is nearly orthogonal.
\end{remark}

\noindent Now we are ready to see the result on the procedure \eqref{SSR_score}.
\begin{theorem}\label{P1.3}
Suppose the same settings in Proposition \ref{P1.2}. Furthermore, assume that for all $i=1,\ldots,n$, there exists a constant $0<\omega$ such that 
\begin{align}\label{S-12}
    \omega< \|\mathbf{d}_i - \mathcal{P}_{D_{[n]\setminus\{i\}}}\mathbf{d}_i \|_2 \eqqcolon \omega_i,\quad 
    (2k-1)\lambda_2+2\varepsilon\leq \lambda_1\omega.
\end{align} 
Then, $J^{k-1}=\mathcal{A}_{n-k}$. Here, $J^{k-1}$ is defined in \eqref{SSR_score}.
\end{theorem}
\begin{remark}
In the proof, one may get the equality $\{j^0,\ldots, j^{k-1}\}=[n]\setminus S^0$ without the condition \eqref{S-11}.
\end{remark}

\subsection{Trade-off between score and sparsity}\label{sec:tradeoff}
One may consider a score-based minimization problem with a sparsity penalty over sub-dictionaries: for some $\alpha>0$,
\begin{align*}
    \min_{\mathcal{D}_{sub}\subset \mathcal{D}}\mbox{Score}(D_{sub};D) +\alpha |\mathcal{D}\setminus\mathcal{D}_{sub}|.
\end{align*}
Interestingly, this formulation resembles the structure of Z-SINDy, particularly in the regime of large sample sizes, a recently developed approach within the framework of statistical mechanics.
While we do not pursue this minimization further in the present paper, we note its connection to Z-SINDy, which addresses a similar problem. A brief discussion is provided in Appendix \cref{app2-Z}.

\section{Numerical simulation}\label{sec:4}
In this section, we numerically test the score in different scenarios, both for ODEs and PDEs. As noted earlier, all simulations are conducted under the weak formulation since we obtained better results than the standard form.
We begin with the ODEs systems (Table \ref{tab:ode_list}) as a show case of how score works in classical benchmark (Lorenz System) and how it can improve the system identification with SINDy in the case with small coefficients (Hopf bifurcation) and a case where SINDy fails to identify the correct system (Pitchfork bifurcation). Then we test how the score behaves with noise, both for ODEs and PDEs. The PDEs equations appeared in Ref. \cite{Messenger21-pde}. Finally, we present an unsupervised dataset from a viscoelastic flow in Section \ref{sec:numerical_viscoelastic}.

\subsection{ODEs Systems}\label{sec:numerical_ode}

\begin{figure}[t!]
    \centering
    \subfloat[Lorenz System]{\includegraphics[width=\textwidth]{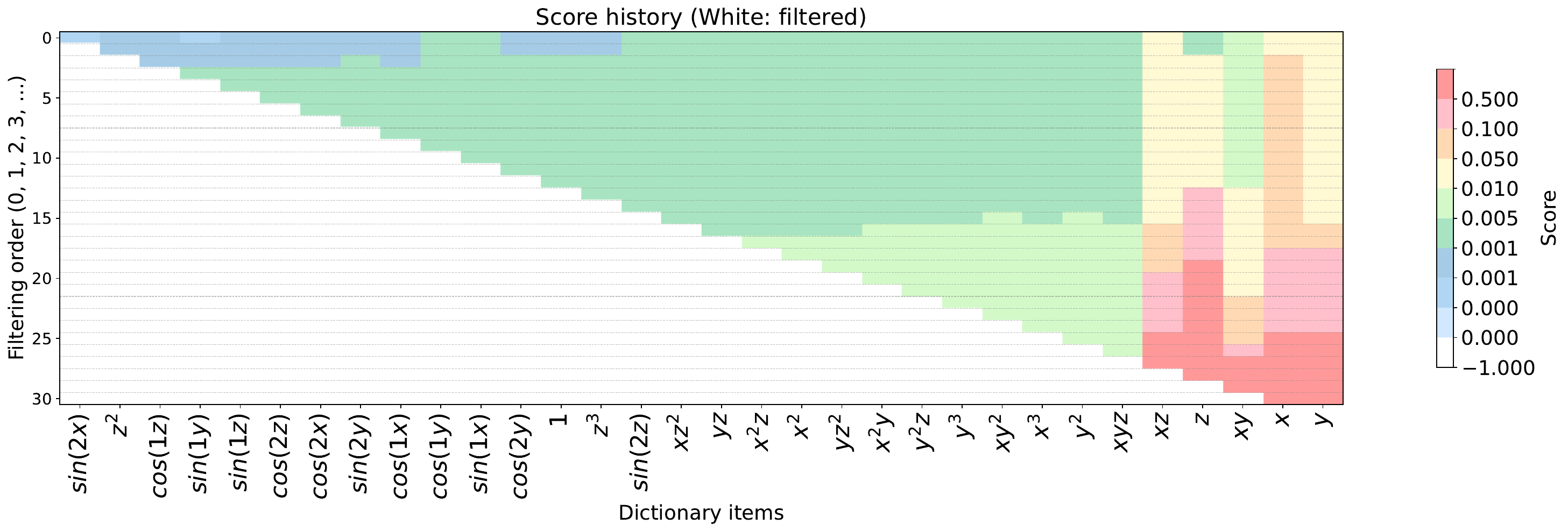}}\\
    \subfloat[Hopf System]{\includegraphics[width=\textwidth]{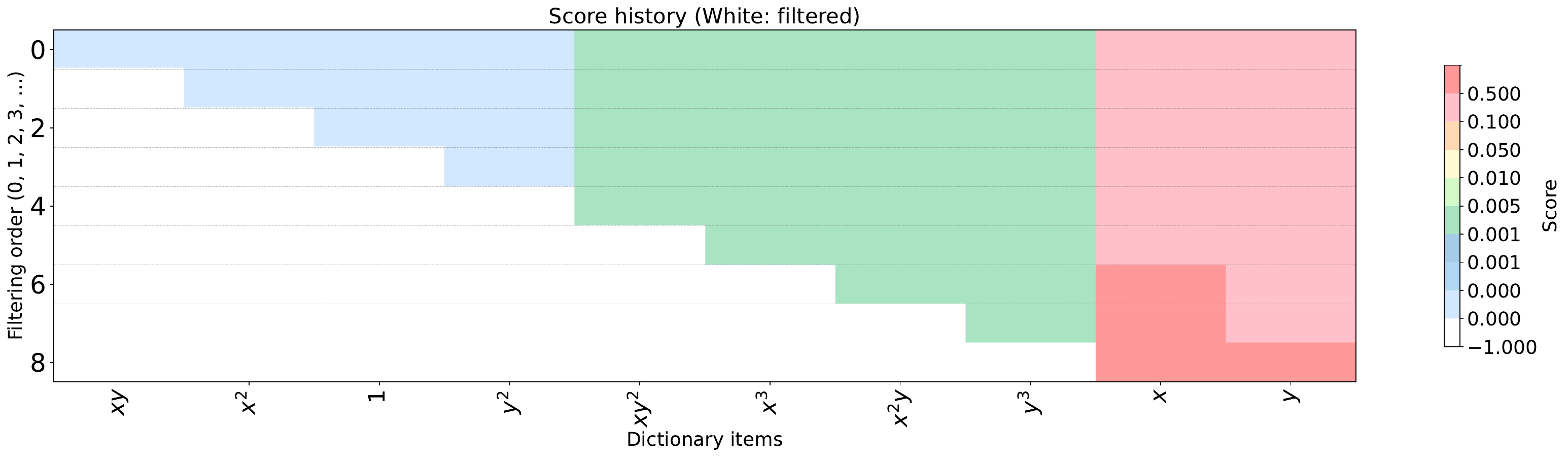}}\\
    \subfloat[Pitchfork Bifurcation]{\includegraphics[width=\textwidth]{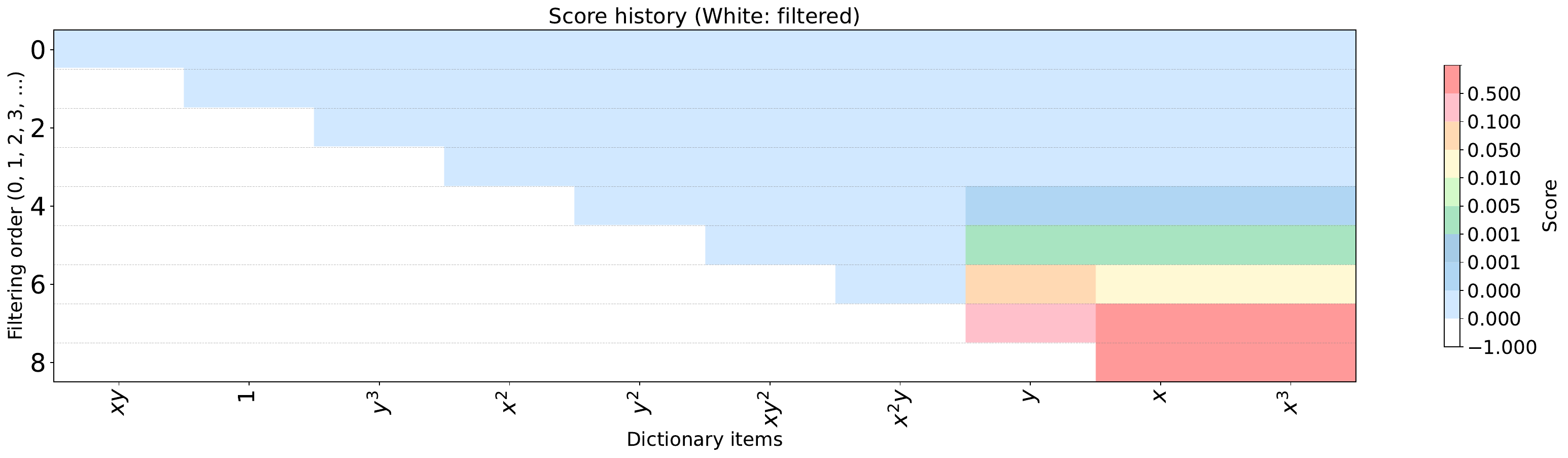}}
    \caption{
    Each row represents a comparison between items, color-coded in the iteration step of \ref{SSR_score}.
    To generate the plots, we iteratively evaluate sub-dictionaries of increasing sparsity. At sparsity level 1 (top row), we compute the total score for each sub-dictionary by summing the scores over all state variables. The sub-dictionary with the lowest total score, $b_0'$, is placed at the leftmost position. At each successive sparsity level (e.g., level 2 for the second row), we evaluate only those sub-dictionaries that contain the previous selection with $b_{i-1}'$, identify the one with the smallest score $a_i$, and place it accordingly. This continues until the desired sparsity level is reached.
    For each system, we analyze the score sequence $(b_i')$  to detect a sudden increase and determine the optimal sparsity level. The last few terms retained are the most important: 5 for the Lorenz system, 6 for Hopf, and 3 for Pitchfork.
    }
    \label{fig:ode_filtered}
\end{figure}

In this section, we apply the score for the equations on Table \ref{tab:ode_list}. We test our stepwise algorithm, which seek to iteratively exclude irrelevant terms in the library until we get only relevant terms. We can see in Figure \ref{fig:ode_filtered} that the correct terms of each equation were kept until the end of filtering, discarding all irrelevant terms first.

\begin{table}[htb!]
\caption{Examples of ODEs Tested}
\label{tab:ode_list}
\resizebox{\linewidth}{!}{
\renewcommand{\arraystretch}{1.7}
    \begin{tabular}{|cc|l|l|}
    \hline
    \multicolumn{2}{|c|}{ODE} & Form & Parameters Used\\
    \hline
     \textbf{\begin{tabular}[c]{@{}c@{}}Lorenz\\ System\end{tabular}}               & 
    \begin{minipage}{.3\textwidth}
        \centering
      \includegraphics[width=.7\linewidth]{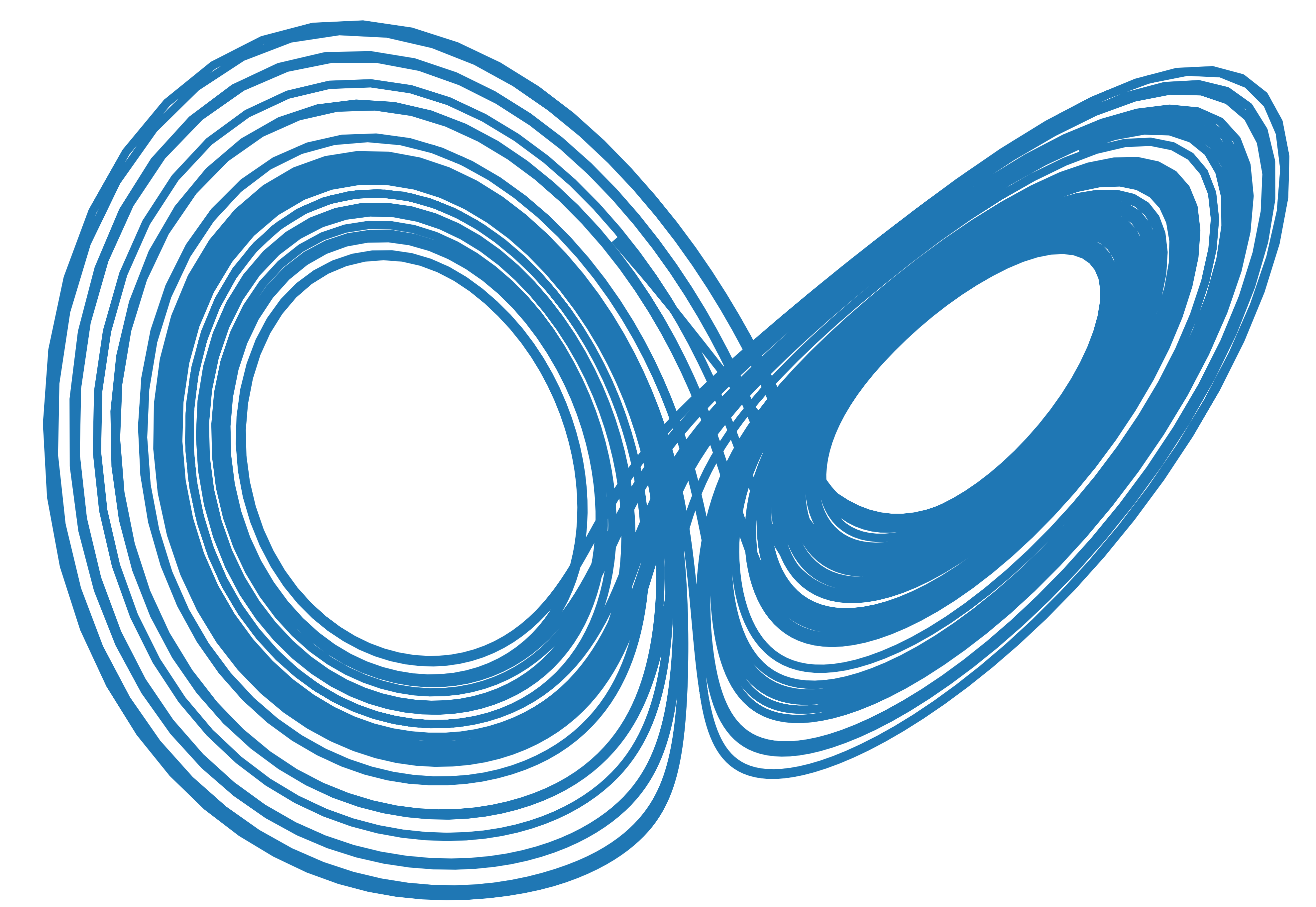}
    \end{minipage}
     &
  \stepcounter{equation}%
    \begin{minipage}{.3\textwidth}\centering
      \stepcounter{equation}%
      \begin{equation*}
        \begin{cases}
          \dot{x} = \sigma (y - x),\\
          \dot{y} = x (\rho - z) - y,\\
          \dot{z} = x y - \beta z
        \end{cases}
        \label{eq:lorenz}
      \end{equation*}
    \end{minipage}
     &
        \begin{tabular}[c]{@{}c@{}}$\sigma = 10,\ \rho = 26,\, \beta = \frac{8}{3};$\\ Initial condition: $(-8, 8, 27)$\\$T \in [0,10],\, \Delta t = 0.01$\end{tabular}\\
    \hline
    \textbf{\begin{tabular}[c]{@{}c@{}}Hopf\\ Bifurcation \end{tabular}} & 
    \begin{minipage}{.3\textwidth}
        \centering
      \includegraphics[width=\linewidth]{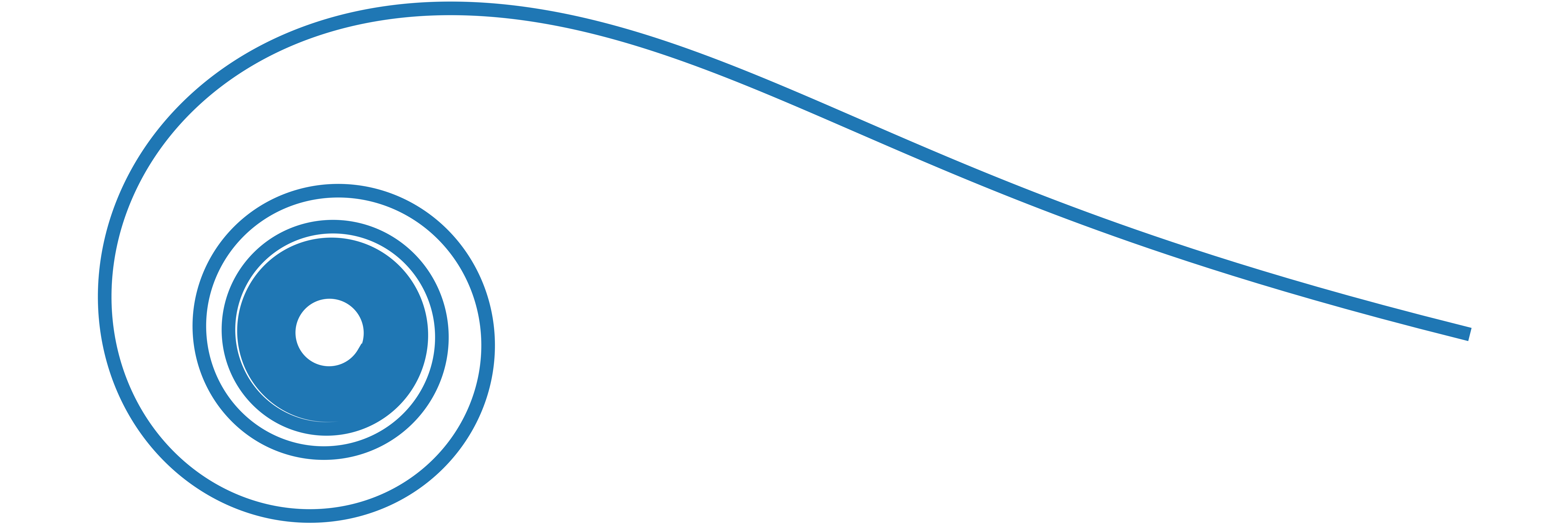}
    \end{minipage}
    & 
    \begin{minipage}{.4\textwidth}\centering
      \stepcounter{equation}%
      \begin{equation*}
        \begin{cases}
          \dot{x} = \mu x - \omega y - x(x^2+y^2), \\
          \dot{y} = \omega x + \mu y - y(x^2+y^2)
        \end{cases}
        \label{eq:hopf}
      \end{equation*}
    \end{minipage}
    &
        \begin{tabular}[c]{@{}c@{}}$\mu = -10^{-5},\ \omega = 1;$\\ Initial condition: $(5, 0)$\\$T \in [0,100],\, \Delta t = 0.01$\end{tabular}\\
    \hline
    \textbf{\begin{tabular}[c]{@{}c@{}}Pitchfork\\ Bifurcation \end{tabular}} & 
    \begin{minipage}{.3\textwidth}
        \centering
      \includegraphics[width=0.7\linewidth]{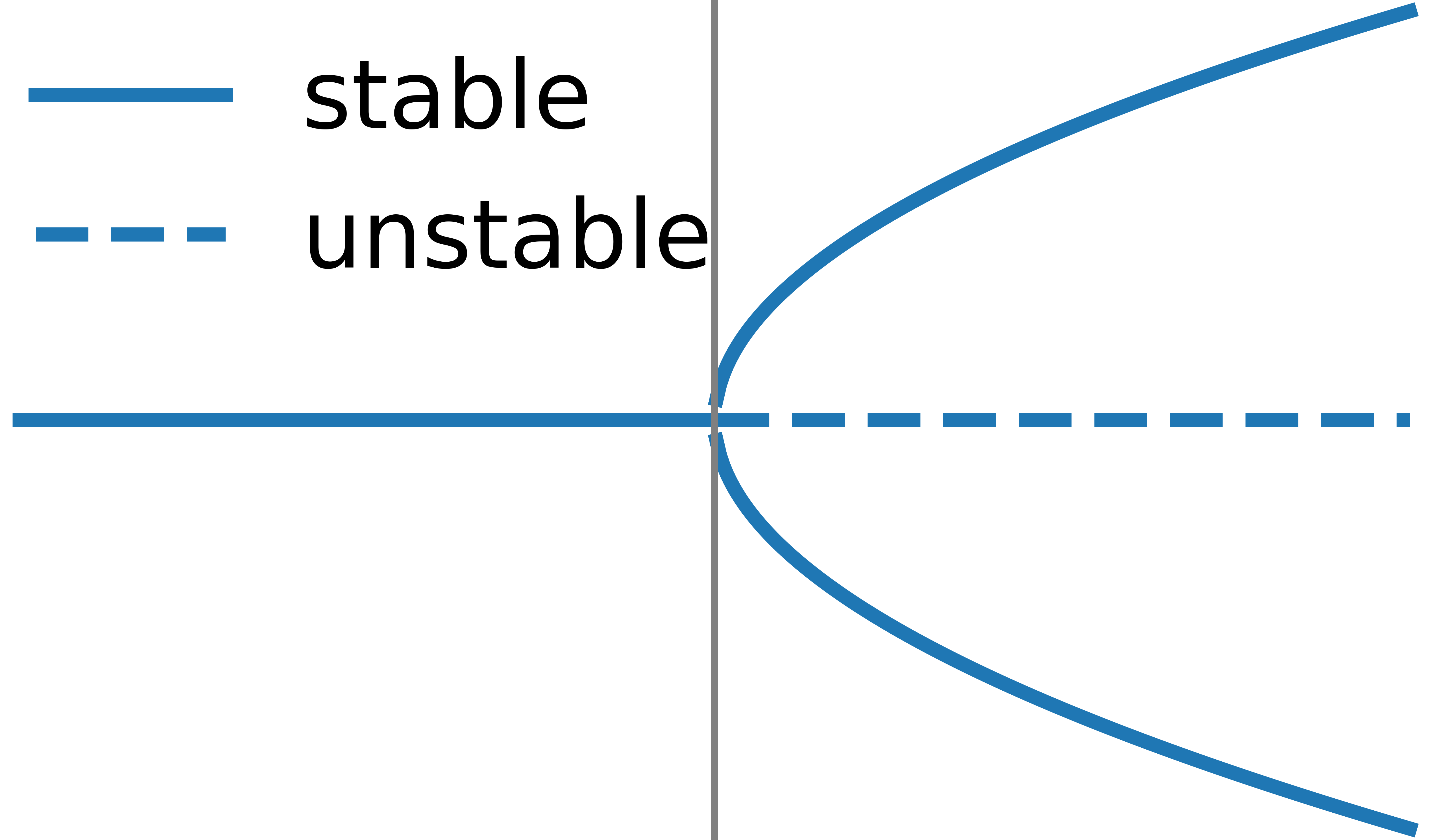}
    \end{minipage}
    &
    \begin{minipage}{.3\textwidth}\centering
      \stepcounter{equation}%
      \begin{equation*}
        \begin{cases}
          \dot{x} = \mu x - x^3, \\
          \dot{y} = -y
        \end{cases}
        \label{eq:pitchfork}
      \end{equation*}

    \end{minipage}
        &
        \begin{tabular}[c]{@{}c@{}}$\mu = 0.5;$\\ Initial condition: $(-1.5, 1)$\\$T \in [0,10],\, \Delta t = 0.01$\end{tabular}\\
    \hline
    \end{tabular}}
\end{table}

\begin{figure}[t!]
    \centering
    \subfloat[Lorenz System]{\includegraphics[width=0.49\linewidth]{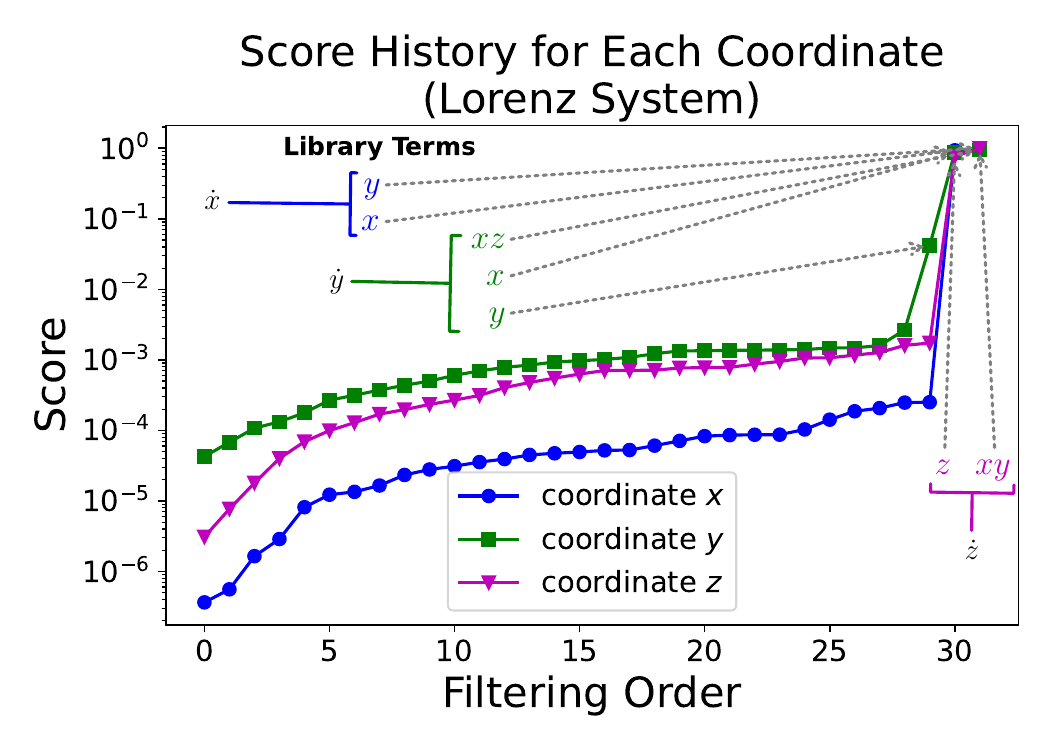}}
    \subfloat[Hopf System]{\includegraphics[width=0.49\linewidth]{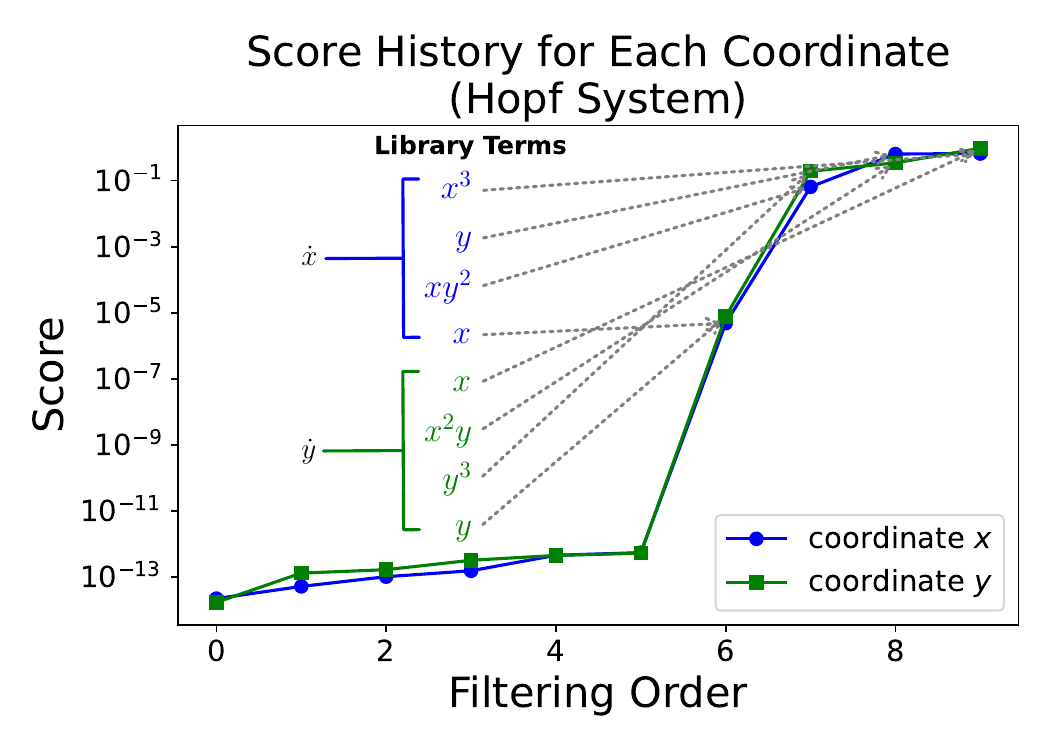}}\\
    \subfloat[Pitchfork Bifurcation]{\includegraphics[width=0.49\linewidth]{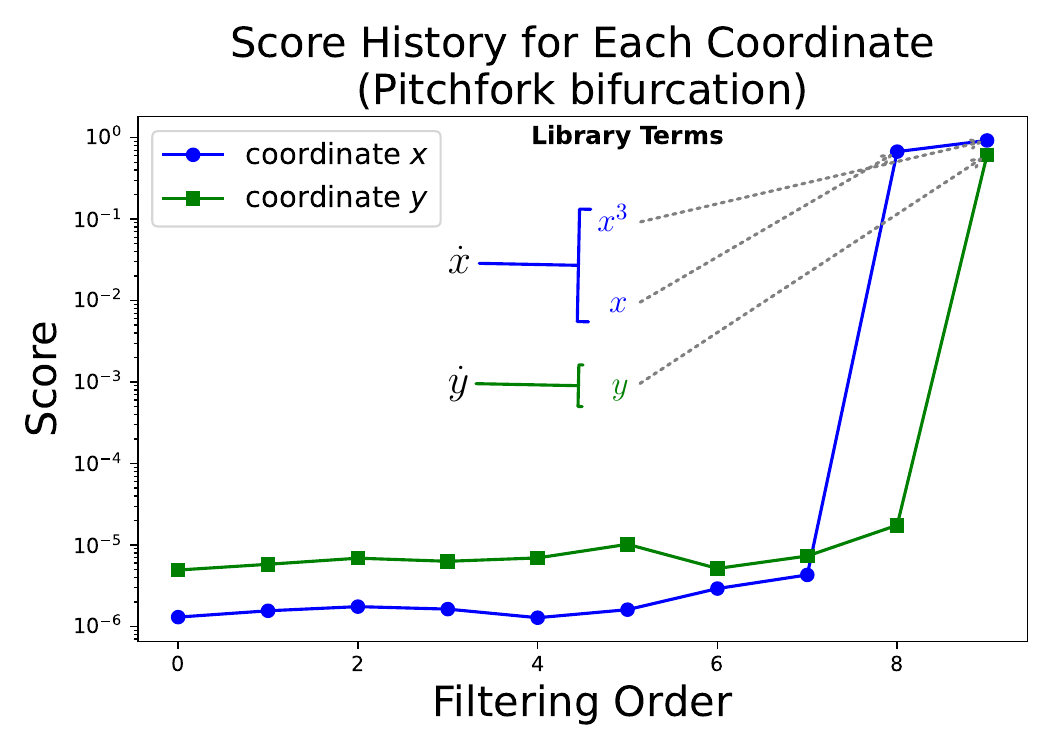}}
    \caption{Minimum score sequence $(b_i)$ defined in \eqref{eq:GBSR_score} of the algorithm \ref{SSR_score} for each coordinate of the ODE systems. For each variable, a noticeable increase in the score occurs when a correct term is removed from the library, indicating the point of optimal sparsity.}
    \label{fig:score_component-wise}
\end{figure}

We also observe a pattern in the scores over the iterations for the trajectory of each coordinate of the system. Figure \ref{fig:score_component-wise} shows the score of each iteration and, when the process filter out a relevant term for the equation, there is a "jump" in the score, which is a indicative of optimal sparsity.

We could not automatically identify this jump in the score, so the selection of optimal sparsity is made empirically by the observation of the relative scores.

\subsubsection{Scores}
For the purpose of comparison, we introduce two other scores: Pareto-score and cross validation scores. 
Recall the definition of our (projective) score and then it is natural to directly use the reconstruction error in the definitions. So we denote it by Pareto-score:
\begin{equation}\label{eq:pareto_score}
\Pareto(D_{sub};D,y) = \frac{\left\|y - \mathcal{P}_{D\setminus D_{sub}} y\right\|_{{\rm L}^2}}{\|y\|_{{\rm L}^2}}
\end{equation}
We also denote by $c_i$ the Pareto-score counterpart of the sequence $(a_i)$ defined in Section~\ref{sec:3.2.3}.

For the $k$-fold cross validation score (See Section \ref{sec:stepwise_Sparse_Regressor}), we use the same notation $\delta$ as in Ref. \cite{Boninsegna18}. For a given dataset $\mathbf{x}=[x(t_i)]_{1\leq i\leq m}$, we split the (sampling) index set $[m]$ into $k$ disjoint equivalent subsets $A_j$ satisfying $\cup A_j = [m]$. Denoting the restriction of a vector (resp. matrix) $y$ onto $A_j$ by $y|_{A_j}$, we define the $i$th cross validation score as
\begin{equation}\label{eq:cross_validation}
\delta^2[i] = \frac{1}{k}\sum_{\ell=1}^k \| \xi|_{A_\ell} - D|_{A_\ell} \cdot \SSR(D|_{B_\ell}, y|_{B_\ell})_{i} \|_2^2,
\end{equation}
where $B_n = \cup_{j\neq n}A_j$.

To observe the data patterns, we will use the relative scores of the form $a_i/a_{i-1}$, $b_i/b_{i-1}$, and $\delta[i]/\delta[i-1]$.

\subsubsection{Lorenz System (without noise)}
The Lorenz system is widely used as a benchmark for Sparse Identification of Nonlinear Dynamical Systems (SINDy) due to its chaotic nature and the challenge it poses in identifying the underlying governing equations from data.

For the library, we used a set of polynomial basis functions up to degree 3, including all possible monomials of the form $x^i y^j z^k$ where $i + j + k \leq 3$ combined with trigonometric functions with 2 periodicities, resulting in a total of 32 functions:
\begin{equation}
\label{eq:lorenz_lib}
    \Theta(\mathbf{X}) = \left[\begin{array}{cccccccccccc}
        1 & x & y & z  & \cdots & z^3 & \sin(x) & \cos(x) & \sin(y)  & \cdots & \sin(2z) & \cos(2z)
    \end{array}
    \right]
\end{equation}

Even with a large library, the scoring method is able to identify the correct terms. An interesting case arises for the $y$-coordinate: although the relevant library term ($y$) is correctly ranked above the incorrect items, its score is noticeably lower than those of the other true items. This may cause one to incorrectly eliminate one true dictionary item for a unsupervised system with only empirical observation of the score curve. A similar result was obtained in ZSINDy where the same $y$ is first to be identified incorrectly under noise \cite{KBKM24}.

\subsubsection{Hopf bifurcation}\label{sec:numerical_hopf}
We revisit the Hopf bifurcation example in Ref. \cite{Brunton16} (\Cref{tab:ode_list}), where $\omega>0$ is a fixed parameter that determines the rotational frequency, and $\mu$ is a bifurcation parameter that controls the growth or decay of amplitude. The asymptotic behavior depends on $\mu$. When $\mu>0$, limit cycles appear. When $\mu=0$, the system undergoes a Hopf bifurcation. The origin is a center: trajectories neither spiral in nor out but form closed orbits (non-asymptotic periodic motion). When $\mu<0$, the origin becomes a globally asymptotically stable spiral (attractor). All trajectories decay toward the origin regardless of initial conditions.

When $\mu<0$ but with small magnitude, short-time trajectories appear similar to those at $\mu=0$. However, their asymptotic behaviors differ. STLS may detect such linear terms if an appropriate threshold is chosen. In this example, in order for SINDy to correctly identify the system, the threshold must be smaller than the $\mu$ parameter, which was chosen to be small ($\mu = -10^{-5}$). In contrast, the score-based strategy may perform robustly regardless of the threshold.

Here, the dictionary is monomials with degrees less than or equal to $3$; $\mathcal{D}=\{x^iy^j:i+j\leq 3\}$. Without the pre-processing, the SINDy with threshold $\lambda = 5\times10^{-6}$ yields the following system:
 \begin{equation}
     \begin{cases}
         \dot{x} =& -0.003086 + 0.038567 x -1.013777 y + 0.390146 x^2 -0.015740 x y \\
         & -0.104667 y^2 -1.159224 x^3 -0.348013 x^2 y -1.443307 x y^2 + 0.277868 y^3 \\
         \dot{y} =& 7.3 \times10^{-6} + 0.999047 x + 0.000333 y -0.009260 x^2 + 0.000258 x y\\
         &+ 0.002534 y^2 + 0.003947 x^3 -0.992273 x^2 y + 0.011203 x y^2 -1.006687 y^3.
     \end{cases}
 \end{equation}
This model includes many incorrect terms with coefficients larger than those in the true system, which causes SINDy to select spurious terms in the final model. In contrast, the proposed pre-processing of the library does not require any thresholding. The user can identify relevant terms by examining the score behavior for each coordinate, as shown in Figure~\ref{fig:score_component-wise}, and then perform regression using only the selected terms. When the weak formulation is applied, STLS correctly identifies the relevant terms. However, in scenarios where the user lacks prior knowledge of the true terms, an inappropriately high threshold may be chosen---potentially larger than some true coefficients---which again results in an incorrect system.

\subsubsection{Pitchfork Bifurcation}
\begin{figure}[t!]
    \centering
    \includegraphics[width=\linewidth]{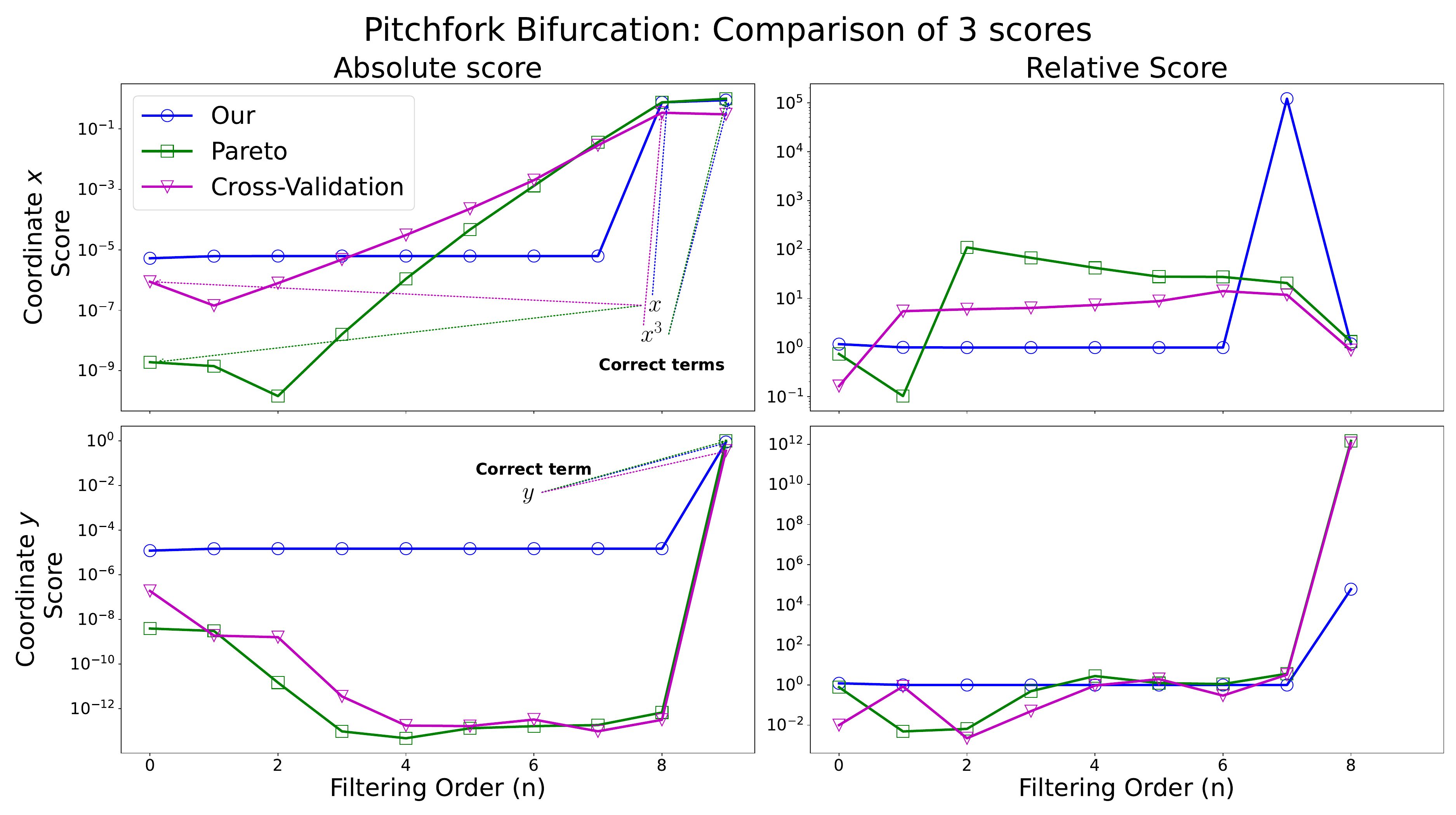}
    \caption{Comparison of our score with the Pareto score and cross-validation score. Only our method successfully identify the correct terms of the system for this setting of weak formulation hyperparameters. The first column shows the absolute score obtained by each method, while the second column presents the score ratio between steps $n$ and $n-1$ (relative score). The top row corresponds to the results for the coordinate $x$, and the bottom row shows results for the coordinate $y$. Our score (blue circle)  exhibit a clear peak in the relative score, which enables identification of the optimal sparsity. In contrast, the Pareto score (green square) and the cross-validation score (magenta triangle) shows this peak behavior only for the $y$-coordinate, offering no consistent criterion for determining optimal sparsity across all coordinates.}
    \label{fig:score_SSR_pareto}
\end{figure}

 While the filtering procedure are sensitive to the weak formulation parameters (e.g., polynomial degree, number of test functions, and support size), making careful hyperparameter tuning essential, we found that an appropriate choice of weak formulation parameters allowed the score-based filtering to recover the correct terms for a single trajectory, whereas SINDy alone could not identify the correct solution without filtering. Furthermore, this example shows that our scores outperfom other scoring strategies, as illustrated in Figure~\ref{fig:score_SSR_pareto}. In our tests, we used a number of test functions equal to twice the number of library terms. The polynomial degree of the test functions was set to 17. The support size for the test functions was determined following the method proposed in \cite{Messenger21}, which leverages Fourier modes to capture large variations in the system’s dynamics within each test function interval. We observed that small changes in any of these hyperparameters significantly affect the results of both the filtering and SINDy procedures, making careful hyperparameter tuning essential.

With this choice of weak formulation hyperparameters, SINDy produces the following system:
\begin{equation*}
    \label{eq:sindy_pitchfork_nofilt}
    \begin{cases}
    \dot{x} &= 1.766 1 + 0.500 x  -3.531 x^2  -1.000 x^3 + 2.747 x^2 y \\
    \dot{y} &= 4.087 1  -1.000 y  -8.173 x^2 + 6.357 x^2 y 
    \end{cases}
\end{equation*}
This model includes the correct terms with accurate coefficients but also contains additional spurious terms. It is important to note that the results are highly sensitive to the choice of hyperparameters. Therefore, it is possible that an optimal selection of these parameters could lead SINDy to recover the exact system without extra terms.

A practical note is that using multiple trajectories generated under different simulation parameters may mitigate this sensitivity. In this setting, both SINDy and the score-based method are able to recover the correct system, and their performance becomes much less dependent on hyperparameter choices.

\subsubsection{Noisy data}
One important aspect is how the method behaves with noisy data.
SINDy variations such as Weak SINDy (WSINDy) \cite{Messenger21} and Ensemble SINDy (ESINDy) \cite{FKBB22} are examples of methods that deal with noise in the data, making them more useful in practial applicaions.

The noisy data $X_{\text{noise}}$ is obtained by adding a gaussian noise to the original data $X$ according to:
\begin{equation*}
    \label{eq:add_noise}
    X_{\text{noise}} = X + \eta \sqrt{\frac{\sum X_i^2}{N}}\mathcal{N}(0, 1)
\end{equation*}
where $\eta$ is the relative noise level. For each noise level, 100 samples were generated.

The stepwise regressor are built under the heuristic assumption that $\mathcal{A}_{i} \subset \mathcal{A}_{i+1}$, where $\mathcal{A}_{i}$ is defined by Eq.~\eqref{SSR_score}. However, the sharp increasing pattern was not observed in the presence of noise, indicating that exhaustive search \eqref{eq:iteration} is the recommended method in such cases.

To test the correct identification of terms after filtering, we evaluated the ability to recover the terms for the $y$-coordinate of the Lorenz system under noisy conditions. This coordinate was chosen because the contribution of its correct term to the score is relatively low, as shown in Figure \ref{fig:score_component-wise} (the $y$ term has a lower score compared to the others).

\begin{figure}
    \centering
    \includegraphics[width=\linewidth]{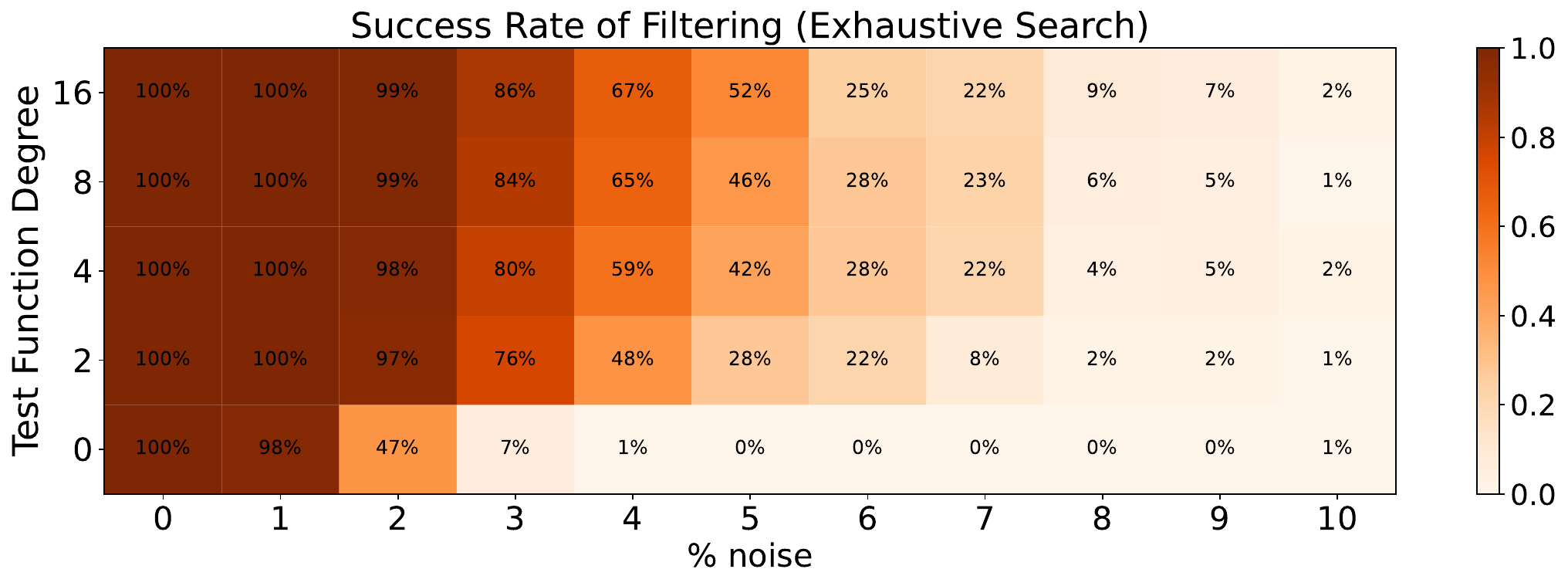}
    \caption{Each row presents the success rates of recovering the correct terms associated with the $y$ coordinate of the weak-formulated Lorenz system using score-based exhaustive search. Each column indicates the level of additive noise.
    }
    \label{fig:noise_robustness}
\end{figure}

We performed an exhaustive search over all possible combinations of three functions from the library and considered the identification successful only when all three correct terms from the original equation were selected. The results are presented in Figure \ref{fig:noise_robustness}. We also tested the identification of the correct library using all coordinates, searching for the best combination of five terms. This resulted in 100\% correct identification, likely because the correct terms for the other variables have significantly higher scores than the incorrect ones.

We also highlight the use of the weak formulation with high-degree polynomials as test functions. In this case, the number of test functions was four times the number of library terms. There are some hyperparameters in the weak formulation that could be optimized for improved performance. Additionally, the computational cost is reduced due to the decrease in matrix size, from the number of time points to the number of test functions.

\subsection{PDE-FIND}\label{sec:pde_find}
In this section, we apply our scoring to some time-evolution-ary PDEs.

\begin{table}[t!]
\caption{PDEs used for testing the selection}
\label{tab:pde_list}
\resizebox{\linewidth}{!}{
\renewcommand{\arraystretch}{1.7}
    \begin{tabular}{|cc|l|}
    \hline
     \multicolumn{2}{|c|}{PDE} & Form\\
    \hline
     \textbf{\begin{tabular}[c]{@{}c@{}}Inviscid\\ Burgers\end{tabular}}               & 
    \begin{minipage}{.3\textwidth}
        \centering
      \includegraphics[width=.7\linewidth]{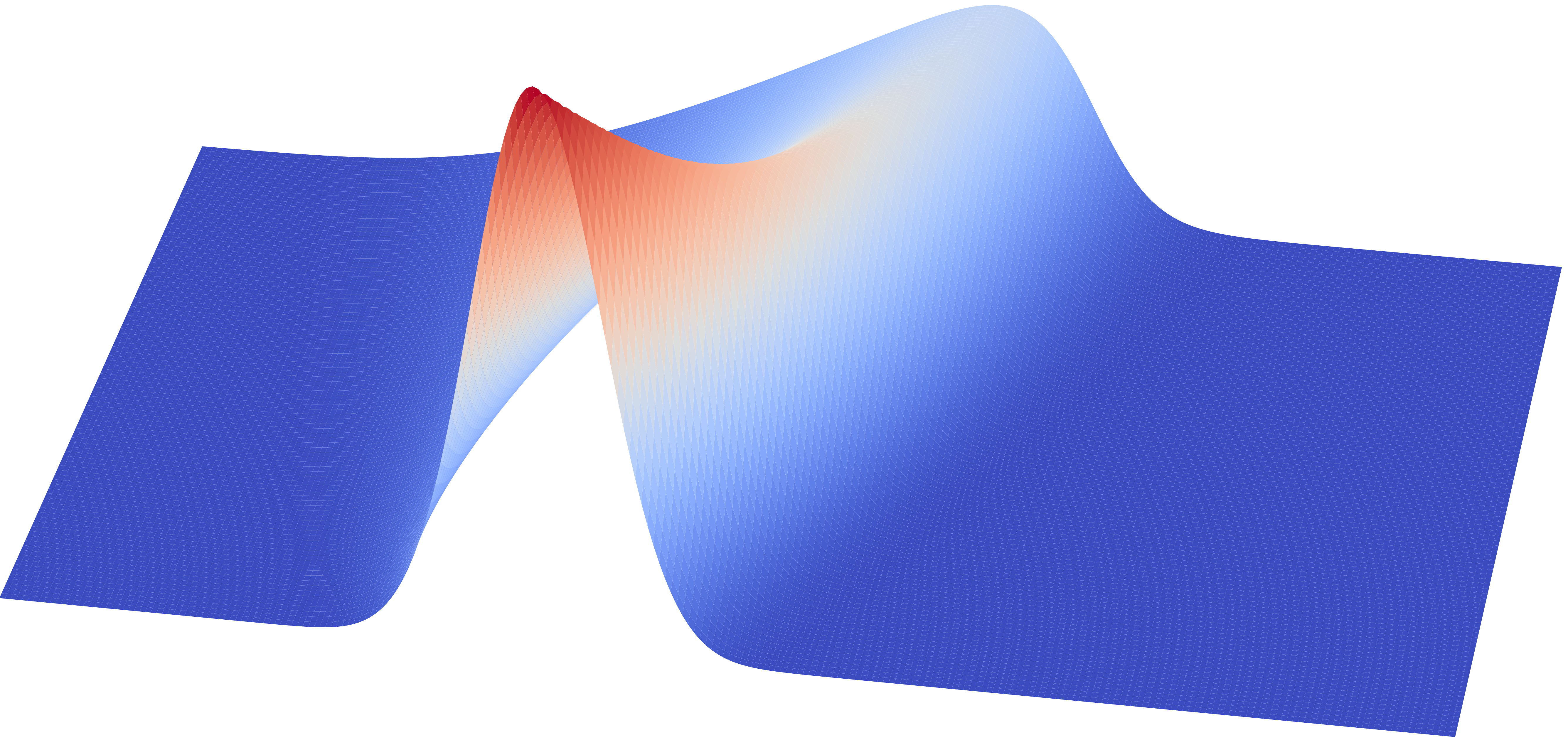}
    \end{minipage}
     & $\partial_t u = -\frac{1}{2} \partial_x (u^2)$ \\
    \hline
    \textbf{\begin{tabular}[c]{@{}c@{}}Kuramoto-\\ Sivashinsky \\ (KS)\end{tabular}} & 
    \begin{minipage}{.3\textwidth}
        \centering
      \includegraphics[width=0.7\linewidth]{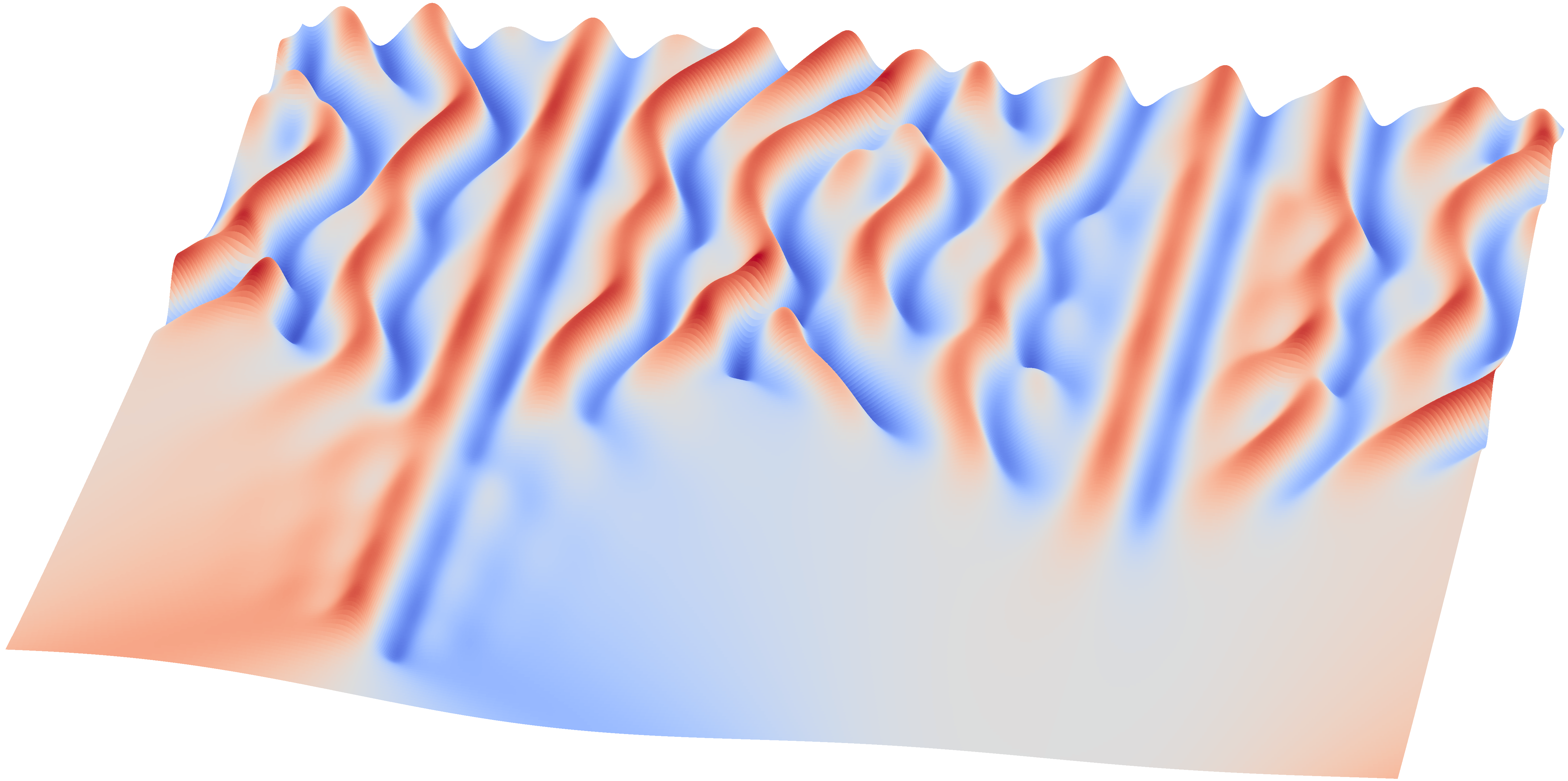}
    \end{minipage}
    & $\partial_t u = -\frac{1}{2} \partial_x (u^2) - \partial_{xx} u - \partial_{xxxx} u$ \\
    \hline
    \textbf{\begin{tabular}[c]{@{}c@{}}Nonlinear \\ Schrödinger\\ (NLS)\end{tabular}} & 
    \begin{minipage}{.3\textwidth}
        \centering
      \includegraphics[width=\linewidth]{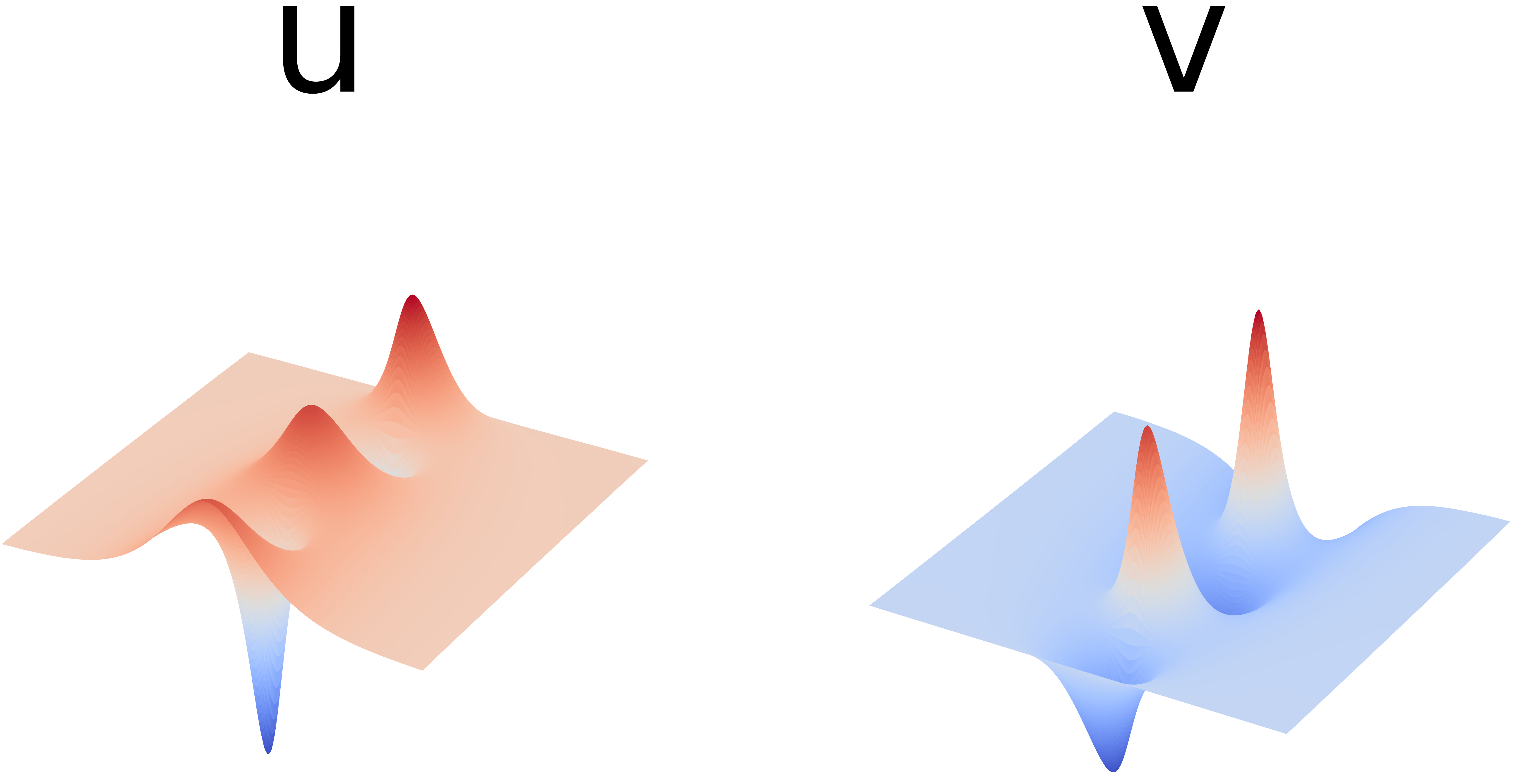}
    \end{minipage}
    &
    $
    \begin{cases}
    \partial_t u = \frac{1}{2} \partial_{xx} v + u^2 v + v^3 \\
    \partial_t v = -\frac{1}{2} \partial_{xx} u - u v^2 - u^3
    \end{cases}
    $ \\
    \hline
    \textbf{\begin{tabular}[c]{@{}c@{}}Reaction-\\ Diffusion\\ (RD)\end{tabular}} & 
    \begin{minipage}{.3\textwidth}
        \centering
      \includegraphics[width=\linewidth]{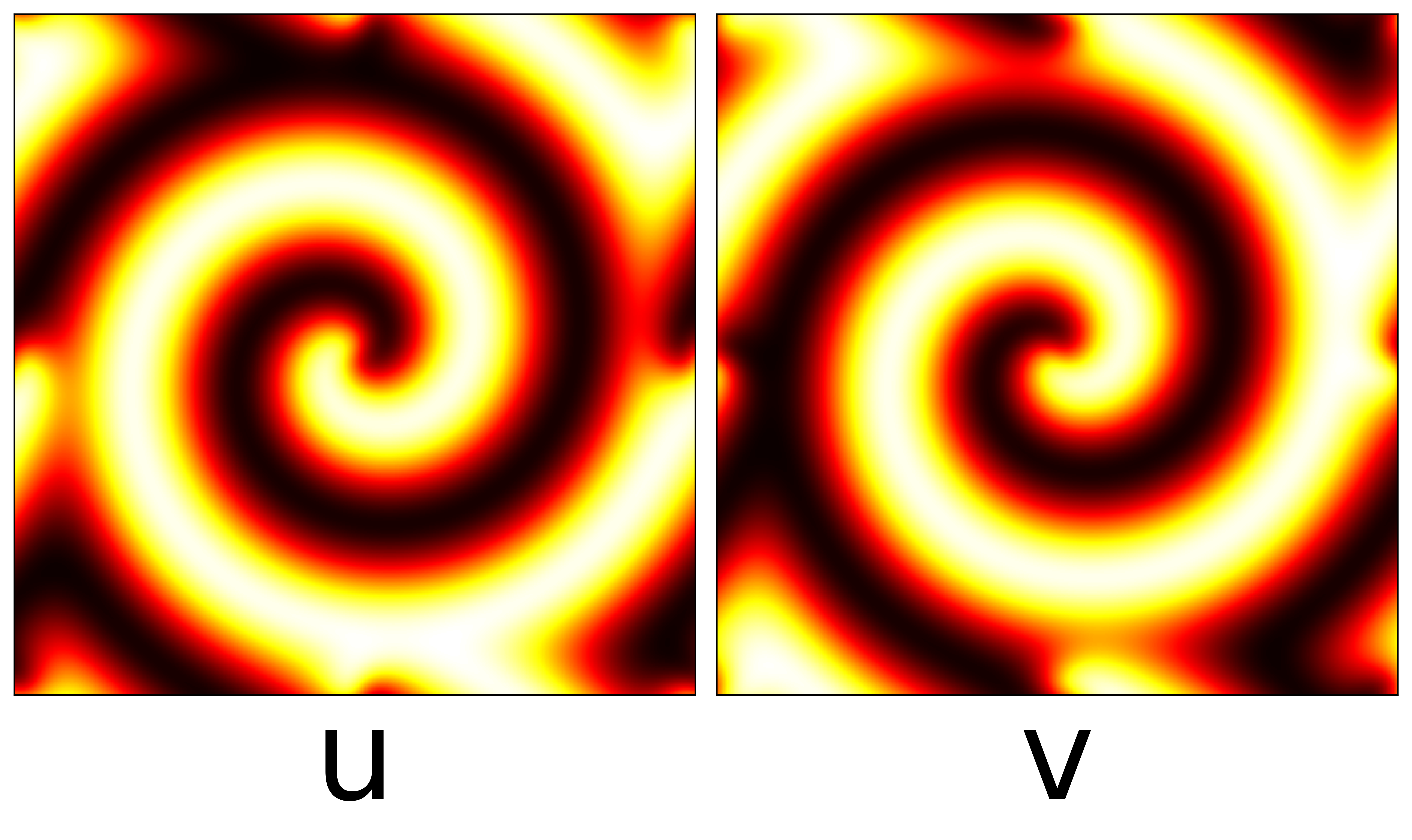}
    \end{minipage}
    &
    $
    \begin{cases}
    \partial_t u = \frac{1}{10} \partial_{xx} u + \frac{1}{10} \partial_{yy} u - u v^2 - u^3 + v^3 + u^2 v + u \\
    \partial_t v = \frac{1}{10} \partial_{xx} v + \frac{1}{10} \partial_{yy} v + v - u v^2 - u^3 - v^3 - u^2 v
    \end{cases}
    $ \\
    \hline
    \textbf{\begin{tabular}[c]{@{}c@{}}Viscoelastic\\Flow\\(16-Roll)\end{tabular}} & 
    \begin{minipage}{.3\textwidth}
        \centering
      \includegraphics[width=\linewidth]{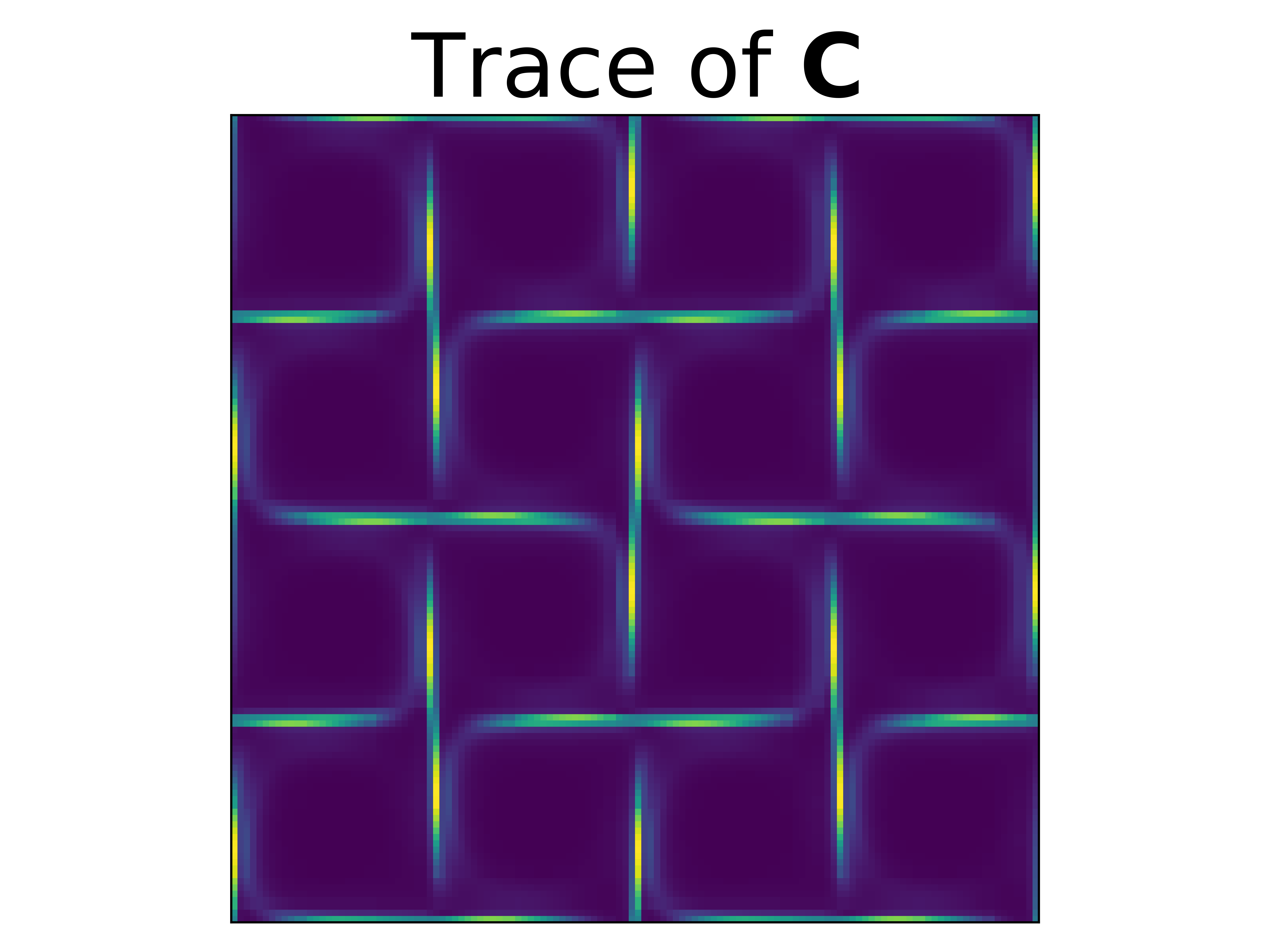}
    \end{minipage}
    &
    $
    \begin{cases}
        \nabla \cdot \mathbf{u} = 0\\
        \partial_t\mathbf{u} + \mathbf{u} \cdot \nabla \mathbf{u} = -\nabla p + \frac{\beta}{Re}\nabla^2\mathbf{u} + \frac{1}{Re}\nabla \cdot \boldsymbol{\tau} + \mathbf{f} \\
    \partial_t \mathbf{C}+ (\mathbf{u}\cdot\nabla)\mathbf{C}
= (\nabla\mathbf{u})\,\mathbf{C} + \mathbf{C}\,(\nabla\mathbf{u})^{T}
- \frac{1}{W_i}\big(\mathbf{C}-\mathbf{I}\big)
    \end{cases}
    $ \\
    \hline
    \end{tabular}}
\end{table}

The simulation shows that the exhaustive search for the correct number of library terms in PDEs is successful . However, a major limitation is that PDE libraries are significantly larger than their ODE counterparts, making exhaustive search considerably more computationally expensive. For example, in the Reaction-Diffusion (RD) case, each field contains 7 correct terms, and the full library consists of 42 terms, rendering a complete combinatorial search infeasible.

\begin{figure}[t!]
    \centering
    \includegraphics[width=\linewidth]{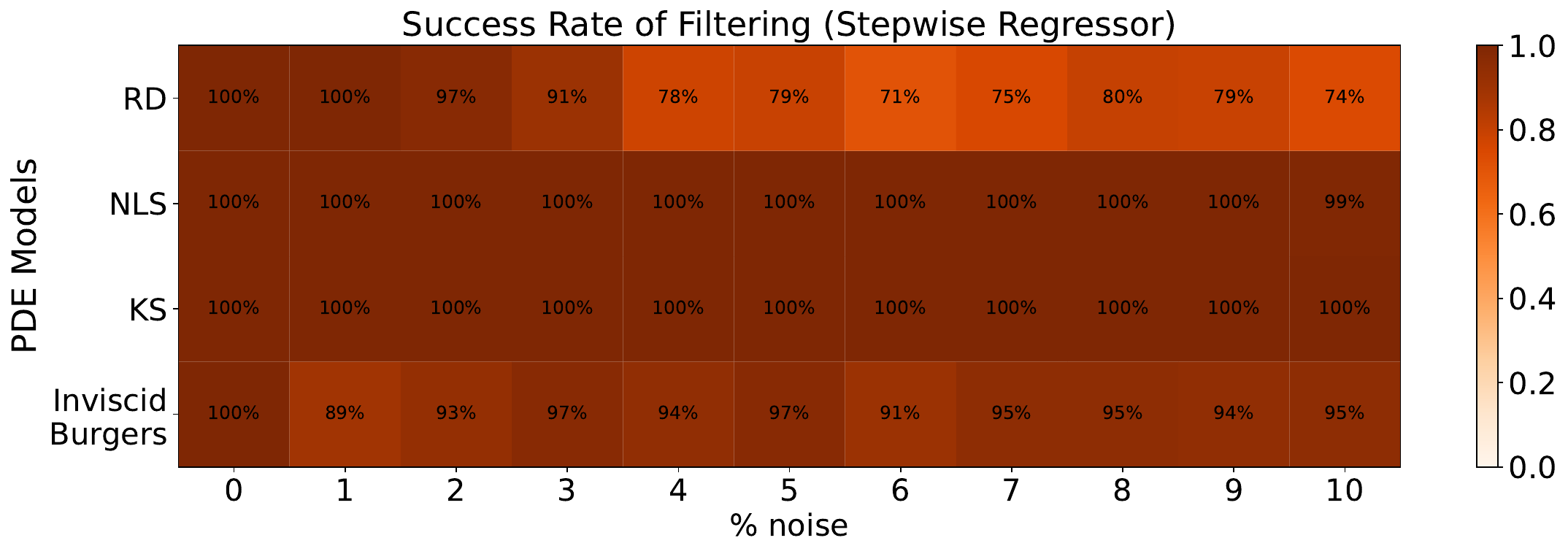}
    \caption{Each row presents the success rates of recovering the correct terms associated with a particular coordinate of the weak-formulated test system using score-based stepwise search. Each column indicates the level of additive noise.}
    \label{fig:pde_stepwise}
\end{figure}

In contrast, the PDE filtering using a stepwise regressor proved to be much more effective, achieving a success rate above 70\% across all cases, even under high noise levels as shown in Figure \ref{fig:pde_stepwise}. This is particularly advantageous for PDEs, where larger libraries are common and stepwise search is less computationally intensive.

\subsection{Proper Orthogonal Decomposition on viscoelastic flow}\label{sec:numerical_viscoelastic}
We apply our method to the dataset from \cite{Oishi24}, which modeled the Proper Orthogonal Decomposition (POD) modes of a numerical simulation of viscoelastic flow in the four-roll mill setup \cite{Thomases2009} using SINDy. 

\begin{figure}[t!]
    \centering
    \includegraphics[width=0.55\textwidth]{Figures/visc_example.png}
    \includegraphics[width=0.4\textwidth]{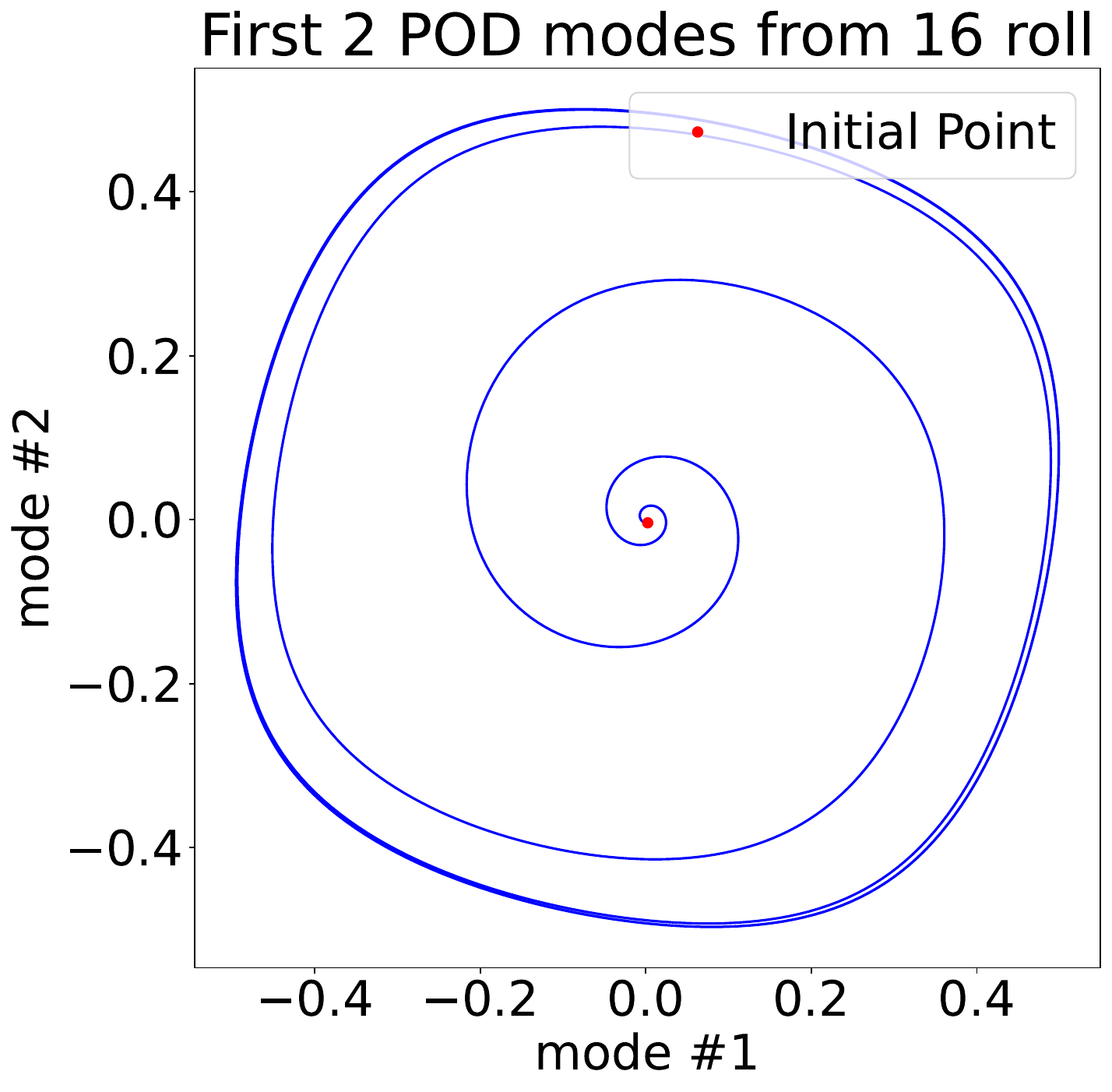}
    \caption{16-Roll data example. On the left we have the full order dada representation. On the right the planar curve is a trajectory from two time-temporal POD modes. The red dot denotes the starting point of the trajectory. The graph seemingly tends to a limit cycle.}
    \label{fig:16_roll}
\end{figure}

In Ref. \cite{Oishi24}, the authors applied POD to a dataset generated by the viscoelastic flow simulation and took two time-temporal modes of them. The two modes produce a planar curve which is reconstructed via SINDy method. On the other hand, Weak SINDy application over POD modes in fluid dynamics also was studied \cite{Russo25}. We use the same dataset of Ref. \cite{Oishi24} in the present paper: for a fixed non-dimensional parameter value of $Wi=3.5$, Figure~\ref{fig:16_roll} shows the temporal evolution of the first two modes. Starting from the origin, the dynamics evolve toward a limit cycle.

In their work, the authors used a function library containing linear and cubic terms, excluding quadratic terms as these negatively impacted the results. Here, we show how our scoring procedure provides a principled justification for this choice by automatically filtering out quadratic terms.

The system discovered by SINDy is given by:

\begin{equation}
    \begin{pmatrix}
        \dot{a}_1\\
        \dot{a}_2
    \end{pmatrix} = 
    \begin{pmatrix}
        \epsilon_1 & \epsilon_2\\
        -\epsilon_2 & \epsilon_1\\
    \end{pmatrix}
    \begin{pmatrix}
        a_1\\
        a_2
    \end{pmatrix}+
    \begin{pmatrix}
        \delta_1a_1^2 + \delta_2a_2^2 + \delta_3a_1a_2 & \delta_4a_2^2\\
        -\delta_4a_1^2 & \delta_2a_1^2 + \delta_1a_2^2-\delta_3a_1a_2
    \end{pmatrix}
    \begin{pmatrix}
        a_1\\
        a_2
    \end{pmatrix}
\end{equation}

\begin{figure}[t!]
    \centering
    \includegraphics[width=\linewidth]{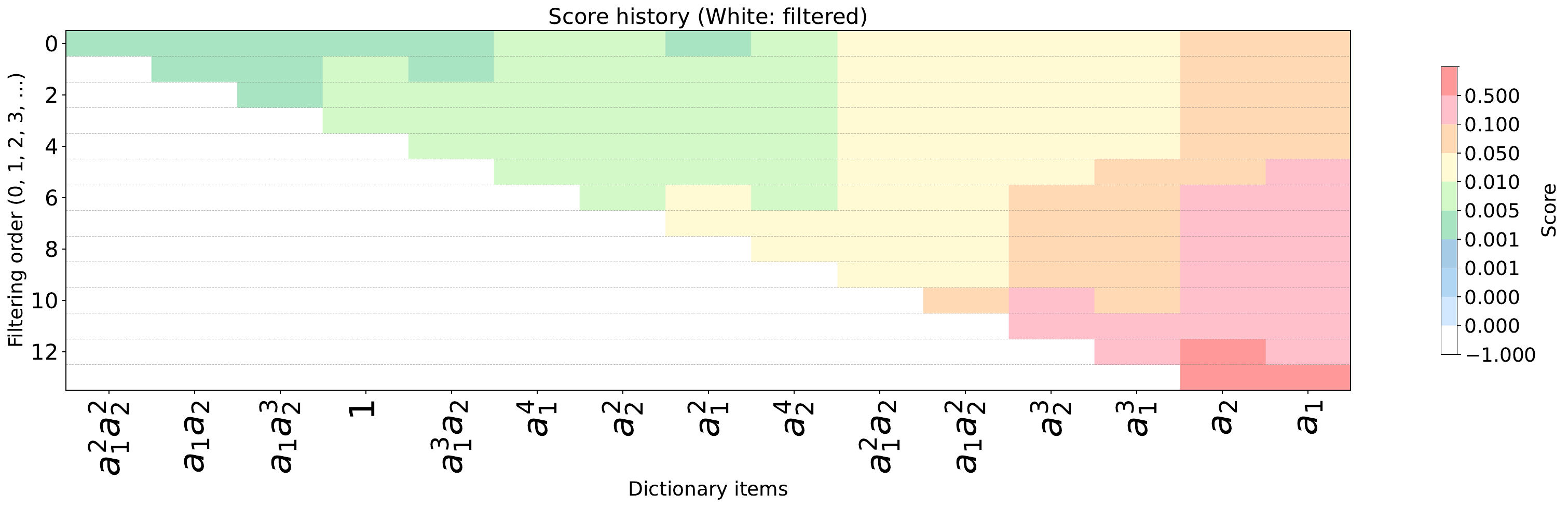}
    \caption{A comparison between items, color-coded in the iteration step of \ref{SSR_score} for POD modes of viscoelastic flow simulation.}
    \label{fig:fitlering_pod}
\end{figure}

\begin{figure}[t!]
    \centering
    \includegraphics[width=\textwidth]{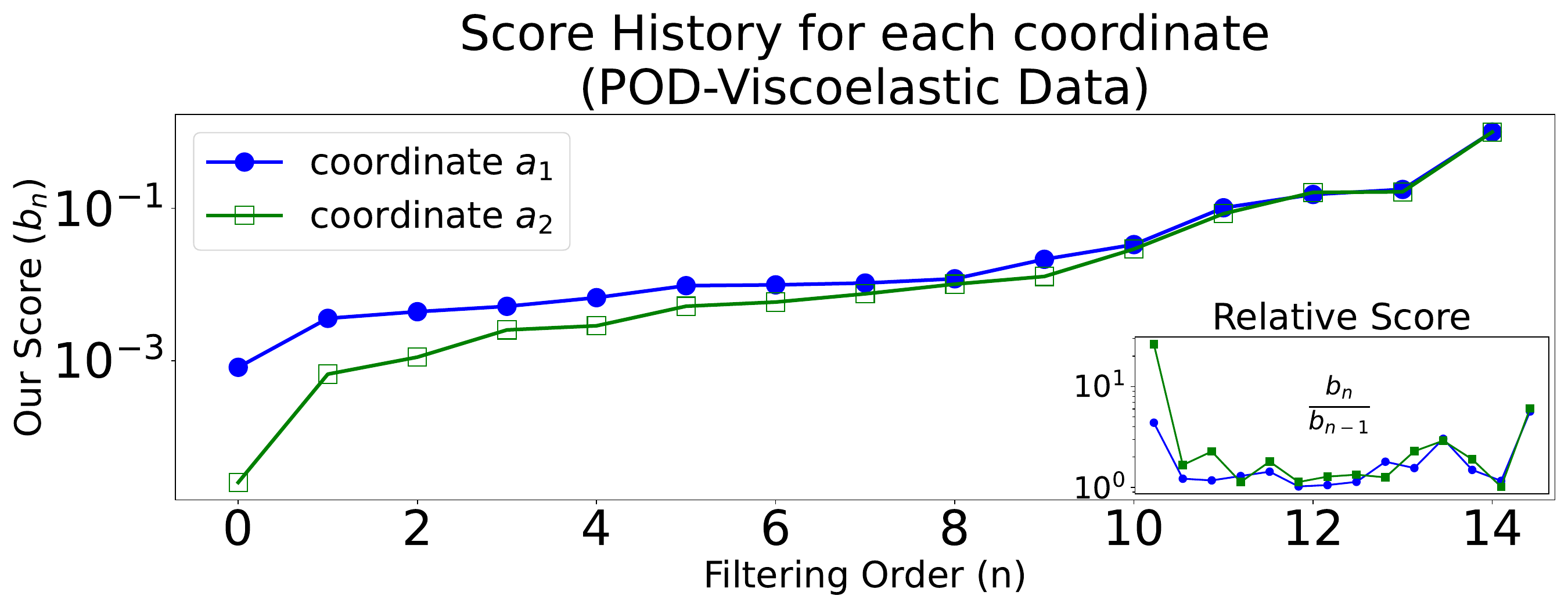}
    \caption{Stepwise regression scores computed using our proposed score for the POD modes of a viscoelastic simulation. The filtering order suggests that quadratic terms are less relevant to the system. However, no clear "jump" in the score is observed, which complicates the identification of an optimal sparsity level.}
    \label{fig:pod_line_score}
\end{figure}

We apply our filtering considering a polynomial library with degree upmost 4. The filtering order in Figure \ref{fig:fitlering_pod} shows how the quadratic and fourth order term were removed first, leaving the linear and cubic term at the end, although it is hard to see the data pattern we expected in Figure \ref{fig:pod_line_score}. It might be due to the non-existence of a ground dynamics consisting with dictionary functions.

\section{Conclusion}\label{sec:conclustion}
In this paper, we proposed a score-based variable selection method that enables the identification of important terms even when their coefficients are small. 
Over the rationale of Pareto curve, we studied the connection between coefficient and score, thereby clarifying how STLS relates to our score framework.
We investigated the power and limitation of our approach through various numerical examples, and also conducted noise tests for the practical utility of our approach for SINDy users. 
In particular, we implemented scoring to weak SINDy formulation.
SINDy user may get help to select threshold or may use our scoring as a pre-processing step for SINDy (See Appendix \ref{sec:screening-process}).
While the concept of score-based selection is not new, our contribution lies in exploring it within the context of dynamical system identification and uncovering connections to existing methods.

The issue of system identification with small coefficients may be mitigated. Our method replaces threshold-based selection a alternative based on score-guided filtering, that only requires a discrete integer hyperparameter: the number of terms, avoiding the need to fine-tune a real-valued threshold. As a result, the burden of parameter tuning is reduced, and systems with small but important terms become more recoverable in practice.

Our results suggest that the pattern of error or score variation across sparsity levels contains important structural information. In particular, the presence of distinct "jumps" in the score can indicate the right model complexity, which guides the discrete hyperparameter mentioned before.

Finally, our framework offers compatibility with various SINDy variants and extensions. The method may integrate into pipelines for system discovery, e.g. implicit differential equation identification, SINDy-PI \cite{Kaheman20}, or unsupervised dataset where right dictionary functions are not clear, as exemplified using the work of \cite{Oishi24}.

Several open problems and improvements remain for future work. These include developing an automatic selection algorithm, deriving optimal conditions in Proposition \ref{P1.2}, conducting statistical analysis of score distributions under noise, explore parametric systems with multiple simulation at once, and aligning model selection strategies with asymptotic system behaviors.

\section*{Acknowledgments}
The work by H. Cho was supported by the National Research Foundation of Korea (NRF) grant funded by the Korea government (MSIT) (RS-2023-00253171), F. Amaral and C. Oishi would like to thank the financial support given by the São Paulo Research Foundation (FAPESP) process numbers  2013/07375-0, 2021/13833-7, 2023/06035-2, 2021/07034-4, and the National Council for Scientific and Technological Development (CNPq), grants 305383/2019-1 and 307228/2023-1. The authors acknowledge support from the National Science Foundation AI Institute in Dynamic Systems (grant number 2112085).. We thank Joel Tropp for pointing us to interesting literature. H. Cho thanks Zackary Nicoulau, Nick Zolman and Doris Voina for helpful discussions.

\newpage
\appendix
\section{Assumption on learning dynamics}\label{sec:App1}
In this section, we introduce several assumptions underlying the learning of dynamical systems -- SINDy and discuss their limitations:
The existence of a well-defined (continuous) ground-truth dynamical system; The assumption that the projected dynamics onto the function space spanned by the dictionary does not diverge. 

Before proceeding, we adopt an assumption for learning dynamics similar to that used in Ref. \cite{Berry25}.\newline
\noindent {\bf Assumption 1.}
For a given time-series dataset, there exists an underlying dynamical system governed by differential equations with a continuously differentiable, time-independent vector field. 

\noindent This is assumed throughout the present paper. Moreover, we do not consider cases involving partial observations (on supervised datasets).

The accuracy of the best approximation depends on the finite difference scheme used to approximate the derivative. Specifically, we have $\|\dot{\mathbf{x}}^* -\dot{\mathbf{x}} \|_{\ell^\infty} = O(\Delta)$ where $\Delta$ is the (uniform) time step. 
Thus, even if one obtains a well-constructed dictionary $D(\mathbf{x})$ and a coefficient vector $\xi$ such that $\|\dot{\mathbf{x}} - D(\mathbf{x})\cdot \xi \|_{\ell^\infty}\leq\|\dot{\mathbf{x}} - D(\mathbf{x})\cdot \xi \|_{\ell^2}= m\cdot O(\Delta)$, the resulting error still satisfies $\|\dot{\mathbf{x}}^*-  D(\mathbf{x})\cdot\xi\|_{\ell^\infty}= O(\Delta)$.
Defining $\|\dot{\mathbf{x}}^*-  D(\mathbf{x})\cdot\xi\|_{\ell^\infty}\eqqcolon \Delta^\infty$, this implies that for $t\in [t_1,t_m]$, $\Big|\mathbf{x}^*(t)-\mathbf{x}^*(t_1) - \int_{t_1}^t D(\mathbf{x}(s))\cdot \xi ds \Big| = (t-t_1)\cdot \Delta^\infty$.
Thus, our objective is to determine the optimal $\xi$ such that the error remains at the order of $O(\Delta)$. However, we will not further discuss this limitation in this paper.
On the other hand, Tran and Ward \cite{Tran17} proved the exact recovery of chaotic behavior from an $\ell^1$ minimization problem under this reformulation.

To overcome such limitations, several studies have proposed methods that avoid the direct computation of numerical derivatives. One notable approach is the weak formulation, which led to the development of Weak SINDy \cite{Messenger21, Russo24}. By introducing test functions, this framework provides greater flexibility when dealing with noisy or sparse datasets. Another method, RK4-SINDy \cite{Goyal22}, incorporates the structure of the Runge--Kutta integration scheme to bypass derivative estimation altogether, embedding finite difference constraints directly into the model identification process.

These derivative-free or derivative-relaxed approaches reveal a more fundamental learning limitation. Given a prescribed dictionary, learning the coefficients amounts to approximating the projection $\mathcal{P}_D\dot{\mathbf{x}}$ of the true derivatives onto the span of the dictionary, rather than recovering $\dot{\mathbf{x}}$ itself. To proceed with our analysis under this viewpoint, we make the following assumption:

\vspace{.1cm}
\noindent {\bf Assumption 2.} Along the notation in the second paragraph in Introduction, the system $z=z(t)$ governed by $\dot{z}(t) = \sum_{i=1}^n \tilde{\xi}_i d_i(z(t))$ does not blow-up in finite-time.
\vspace{.1cm}

\section{Recovery of dynamics}\label{app:R}
In this section, we study how, in the low-score regime, one can bound the difference between the trajectories generated by the time derivative projected onto the full dictionary and those generated by its projection onto the complement of a low-score sub-dictionary.

In this section, we use perturbation theory to derive a theoretical bound on phase differences in relation to the score.
Suppose the state space is compact in an ambient Euclidean space.  According to the classical ODE theory, the perturbed trajectory $x^\varepsilon(t)$, governed by $\dot{x}=f(x)+\varepsilon g(x)$ and $x^\varepsilon(t_0)=x_0$, from $x^0(t)$, governed by $\dot{x}=f(x)$ and $x^0(t_0)=x_0$, could be described in an explicit form.
Kasz\'as and Haller \cite{kaszas20} found that for any $\delta>0$, there exists $\varepsilon_0>0$, such that for $\varepsilon<\varepsilon_0$,
\begin{align}\label{P-1}
|x^\varepsilon(t) - x^0(t)| \leq \varepsilon\left(\int_0^t \Lambda_s^t(x^0(s))ds +\delta \right) \|g(x^0(\cdot))\|_{{\rm L}^2([t_0,t])]},\quad t\geq t_0,
\end{align}
where $\Lambda_s^t(x^0(s))$ is the maximal eigenvalue of the Gram matrix of the fundamental matrix $\Phi(t;s,x^0(s))$ to the dynamics $\dot{x}=f(x)$.\footnote{The original statement uses the maximum norm of $g(x^0(\cdot))$.}

Let $U\subset\mathbb{R}^d$ be a region, and let $h:U\to U$ be a continuously differentiable and bounded vector field.
Consider a system given by $\dot{x}^*(t)=h(x^*(t))$ on $0\leq t\leq T$ with the initial condition $x^*(0)=p\in U$, and suppose $x^*(t) \in U$ for all $t\in[0,T]$.
Let $\mathcal{D}=\{d_1,\ldots,d_n \}\subset \mathcal{C}(U)$ be a dictionary of scalar-valued continuous functions on $U$.
For each $d\in\mathcal{D}$, we assume that the function is normalized along the trajectory $\{x^*(t):0\leq t\leq T\}$, meaning that $\|d(x^*(\cdot))\|_{{\rm L}^2[0,T]}=1$ for any $d\in\mathcal{D}$.
We denote by $\mathcal{D}(x)=\{d_1\circ x,\ldots,d_n\circ x\}$, the set of scalar functions obtained by evaluating the dictionary elements along a trajectory $x:[0,T]\to U$. So, $\mathcal{D}(x)\subset {\rm L}^2[0,T]$.
Suppose the dictionary $\mathcal{D}(x^*)$ approximates $h$ well (along the trajectory $x^*$, in the sense that $\|h\circ x^* -\mathcal{P}_{\mathcal{D}(x^*)}(h\circ x^*)\|_{{\rm L}^2[0,T]}\ll 1$.

Denote $\mathcal{D}_{sub}\coloneqq\mathcal{D}\setminus\{d_{i_1},\ldots, d_{i_\ell}\} \subset \mathcal{D}$, where $d_n\in\mathcal{D}$ is a specific dictionary element.
Let 
\begin{align*}
\mathcal{P}_{\mathcal{D}(x^*)}(h\circ x^*)=\sum_{k=1}^n c_k^{(1)} d_k\circ x^* ,\quad
\mathcal{P}_{\mathcal{D}_{sub}(x^*)}(h\circ x^*)= \sum_{k=1}^{\ell} c_{i_k}^{(2)} d_{i_k}\circ x^*.
\end{align*}
Consider the following two systems:
\begin{align*}
&\dot{x}_1(t) = \sum_{k=1}^n c_k^{(1)} d_k(x_1(t))\eqqcolon F_1(x_1(t)) ,\\
&\dot{x}_2(t) =\sum_{k=1}^{\ell} c_{i_k}^{(2)} d_{i_k}(x_2(t))\eqqcolon F_2(x_2(t)),\quad 0\leq t\leq T,\\
& x_1(0)=x_2(0)=p\in U.
\end{align*}

Assume that the system governing $x_2$ is an $\varepsilon$-perturbed version of the system governing $x_1$. That is, let $f = F_1$ and $f+\varepsilon g = F_2$, where $\|g\circ x^*\|_{{\rm L}^2[0,T]}=1$, following the notation from the previous paragraph.\footnote{In the referenced work \cite{kaszas20}, the restriction $\|g\circ x^*\|_{{\rm L}^2[0,T]}=1$ is not explicitly imposed. To the best of the authors' knowledge, the essential condition is that the perturbation term $g$ is independent of the parameter $\varepsilon>0$. Therefore, our normalization assumption is harmless and does not affect the generality of the analysis.}
This situation is typical in the SINDy framework: coefficients corresponding to important terms (i.e., the survivors) remain relatively stable across iterations, while non-essential terms--those that will be eliminated--tend to have small coefficients and are more sensitive to perturbations.
Then, \eqref{P-1} implies that
\begin{align*}
|x_1(t) - x_2(t)| \leq \left(\int_0^t\Lambda_s^t(x_1(s))ds +\delta\right)  \|g\circ x_1\|_{{\rm L}^2[0,T]},\quad 0\leq t\leq T,
\end{align*}
provided that $\varepsilon$ is sufficiently small. 
The right factor in the right-hand-side approximates $\|g\circ x^*\|_{{\rm L}^2[0,T]}=1$ if $x_1$ approximates $x^*$.\footnote{To simplify the analysis, one may assume that $h\in\spann(\mathcal{D})$.} According to the result in Ref. \cite{kaszas20}, one may numerically choose $\delta=0$, and the resulting integral does exhibit tighter than exponential growth. Therefore, we may argue that the trajectory $x_2$, governed by the sparser system, remains close to the trajectory $x_1$, with a theoretical bound on the difference.
On the other hand, the assumption
\[
\| (\mathcal{P}_{\mathcal{D}(x^*)}-\mathcal{P}_{\mathcal{D}_{sub}(x^*)})(h\circ x^*)\|_{{\rm L}^2[0,T]}=\|F_1\circ x^* - F_2\circ x^*\|_{{\rm L}^2[0,T]}=\varepsilon\ll1
\]
may be achieved when the dictionary element $d_n$ has a small score in the discretized setting. Therefore, to obtain a theoretical bound on the data-fitting error with a sparser sub-dictionary, one may remove the dictionary element with the lowest score.

\section{Collection of Proofs}\label{app:proofs}
In this appendix, we list the proofs for all propositions in the present work. We will use the following lemma without mentioning throughout this section.
\begin{lemma}[\cite{Szedmak24} Theorem 8]
Let $\mathcal{P}_{\mathcal{D}_1}$ and $\mathcal{P}_{\mathcal{D}_2}$ be projections on ${\rm L}^2(\mathcal{M})$. $\mathcal{P}=\mathcal{P}_{\mathcal{D}_1} -\mathcal{P}_{\mathcal{D}_2}$ is projection if and only if $\spann(\mathcal{D}_1)\subset \spann(\mathcal{D}_2)$. Then $\mathcal{P}:{\rm L}^2(\mathcal{M})\to L$, where $\spann(\mathcal{D}_2)=\spann(\mathcal{D}_1)\oplus L$, namely $L$ is the orthogonal complement of $\spann(\mathcal{D}_1)$ in $\spann(\mathcal{D}_2)$.
\end{lemma}

\subsection{Proof for Lemma \ref{L1.1}}
\begin{proof}
This can be deduced from the inverse formula of a partitioned matrix (e.g. Appendix C.3.5 \cite{Montgomery21}). Here, we provide another proof.

We fix an index $i_0\in[n]$.
We swap the $i_0$th column to the last position in $D$, denoting the resulting matrix as $\hat{D}=[\hat{\mathbf{d}}_1|\cdots|\hat{\mathbf{d}}_n]$. Then,
\begin{align}\label{A-1}
    \mathcal{P}_{\hat{D}} y =  \hat{D} \hat{D}^\dagger y = \sum_{i=1}^n [\hat{D}^\dagger y]_i \hat{\mathbf{d}}_i.
\end{align}
The orthonormalization of  $(\hat{\mathbf{d}}_1,\ldots,\hat{\mathbf{d}}_n)$ generates the orthogonal matrix 
\[
\hat{U}=[\mathbf{u}_1|\cdots|\mathbf{u}_n]
\]
where $\mathbf{u}_1=\hat{\mathbf{d}}_1/\|\hat{\mathbf{d}}_1\|_2$, $\mathbf{u}_i= (\hat{\mathbf{d}}_i-f_i(\hat{\mathbf{d}}_1,\ldots,\hat{\mathbf{d}}_{i-1}))/\|\hat{\mathbf{d}}_i-f_i(\hat{\mathbf{d}}_1,\ldots,\hat{\mathbf{d}}_{i-1})\|_2$, and $f_i(\hat{\mathbf{d}}_1,\ldots,\hat{\mathbf{d}}_{i-1}) = \mathcal{P}_{\hat{U}_{\{1,\ldots,i-1\}}}\mathbf{d}_i$ for $i=2,\ldots, n$.
We observe that
\begin{align}\label{A-2}
\mathcal{P}_{\hat{U}} y = \hat{U} \hat{U}^\dagger y = \sum_{i=1}^{n-1} [\hat{U}^\dagger y]_i \mathbf{u}_i+[\hat{U}^\dagger y]_n \mathbf{u}_n
= \sum_{i=1}^{n-1}c_i \hat{\mathbf{d}}_i + \frac{[\hat{U}^\dagger y]_n}{\|\hat{\mathbf{d}}_n-f_n(\hat{\mathbf{d}}_1,\ldots,\hat{\mathbf{d}}_{n-1})\|} \hat{\mathbf{d}}_n,
\end{align}
for some sequence $(c_i)_{i=1}^{n-1}$. We recall the geometric property of the orthonormalization, $\mathcal{P}_{\hat{D}}\equiv\mathcal{P}_{\hat{U}}$. Thus, from \eqref{A-1} and \eqref{A-2}, we have $[\hat{D}^\dagger y]_n = [\hat{U}^\dagger y]_n/\|\hat{\mathbf{d}}_n-f_n(\hat{\mathbf{d}}_1,\ldots,\hat{\mathbf{d}}_{n-1})\|$ since $\hat{\mathbf{d}}_i$ are linearly independent.
\footnote{According to the formula for pseudo-inverse matrices of block matrices, it is the same as the following: $[\hat{D}^\dagger y]_n = [\mathcal{P}^\perp_{\hat{D}_{sub}} \mathbf{d}_n]^\dagger y= [(I - \mathcal{P}_{\hat{D}_{sub}}) \mathbf{d}_n]^\dagger y= \frac{[\mathbf{d}_n - (\mathcal{P}_{\hat{D}_{sub}}\mathbf{d}_n)]^T y}{\|\mathbf{d}_n - (\mathcal{P}_{\hat{D}_{sub}}\mathbf{d}_n)\|^2}$.}
Recalling that $\hat{D}$ was generated by swapping columns, we have $\|\hat{\mathbf{d}}_n-f_n(\hat{\mathbf{d}}_1,\ldots,\hat{\mathbf{d}}_{n-1})\| = \|\mathbf{d}_{i_0}-\mathcal{P}_{D_{[n]\setminus\{i_0\}}}\mathbf{d}_{i_0}\|$ and also the following
\begin{align*}
\begin{aligned}
|[\hat{U}^\dagger y]_n| = \|[\hat{U}^\dagger y]_n \mathbf{u}_n\| 
= \|\mathcal{P}_{\mathbf{u}_n}y\| 
&= \|(\mathcal{P}_{\hat{U}} - \mathcal{P}_{\hat{U}_{[n]\setminus\{n\}}})y\| \\
&=\|(\mathcal{P}_{\hat{D}} -\mathcal{P}_{\hat{D}_{[n]\setminus\{n\}}})y\|
=\|(\mathcal{P}_{D} -\mathcal{P}_{D_{[n]\setminus\{i_0\}}})y\|.
\end{aligned}
\end{align*}
Combining altogether, we have the conclusion.
\end{proof}

\subsection{Proof for Lemma \ref{L1.3}}
\begin{proof}
From the trivial relationship $\spann(D\setminus D_{sub})\subset\spann(D)$, we have 
\[
\mathcal{P}_{D}  \mathcal{P}_{D\setminus D_{sub}}=\mathcal{P}_{D\setminus D_{sub}}\mathcal{P}_{D} = \mathcal{P}_{D\setminus D_{sub}}
\]
and so the map $\mathcal{P}_{D} - \mathcal{P}_{D\setminus D_{sub}}$ is a projection. Since the operator norm of a projection map is less than $1$, we have
\begin{align*}
\mbox{Score}(D_{sub};D)
&\leq \left\|\mathcal{P}_{D} - \mathcal{P}_{D\setminus D_{sub}} \right\|_{op} \leq 1.
\end{align*}
\end{proof}

\subsection{Proof for Proposition \ref{P1.1}}
\begin{proof}
Suppose the sub-dictionary score is sufficiently small, $\score(D_{sub};D)< \lambda\omega/\|y\|_2$. Then, 
\begin{align}\label{S-7}
\score(\mathbf{d}_i; D) \overset{\eqref{S-5}}{\leq} \score(D_{sub};D) \leq \lambda\omega/\|y\|_2.
\end{align}
The condition \eqref{S-6} and the above \eqref{S-7} imply \eqref{A-3}:
\[
\|y\|_2 \score(\mathbf{d}_i;D) <
\lambda \|\mathbf{d}_i - \mathcal{P}_{D_{[n]\setminus\{i\}}}\mathbf{d}_i \|_2.
\]
\end{proof}

\subsection{Proof for Proposition \ref{P1.2}}
\begin{proof}
Without loss of generality, let $S^0 = \{1,2,\ldots, k\}$.
By definition of $S^0$, $|c_i|\geq \lambda_1$ for $i\leq k$ and $|c_i|<\lambda_2$ for $i\geq k+1$.
Our claim is to show $\mathcal{A}_{n-k}=\{d_{k+1},\ldots, d_n\}$. In other words,
\begin{align}\label{S-10}
\begin{aligned}
&\score(D_{[n]\setminus S^0};D) \leq \score (D_{[n]\setminus T};D)\\\
&\quad \mbox{for}~T\subset[n],\quad |T|=k,~T\cap S^0 \neq \emptyset,~T\cap([n]\setminus S^0)\neq\emptyset.
\end{aligned}
\end{align}
Fix such $T$ with $|T\cap S^0|=\ell$. We compute the denominators of the scores in \eqref{S-10} through
\begin{align*}
\mathcal{P}_{D_{S^0}}y &= \sum_{i=1}^k c_i\mathbf{d}_i + \sum_{i=k+1}^n c_i\mathcal{P}_{D_{[n]\setminus S^0}}\mathbf{d}_i + \mathcal{P}_{D_{[n]\setminus S^0}}\vec{e},\\
\mathcal{P}_D y - \mathcal{P}_{D_{S^0}}y 
&=  \sum_{i=k+1}^n c_i(\mathcal{I}-\mathcal{P}_{D_{[n]\setminus S^0}})\mathbf{d}_i - \mathcal{P}_{D_{[n]\setminus S^0}}\vec{e}.
\end{align*}
We obtain
\begin{align}\label{S-8}
\|\mathcal{P}_D y - \mathcal{P}_{D_{S^0}}y \|_2 &\leq (n-k)\lambda_2+\varepsilon.
\end{align}
Second, we compute
\begin{align*}
\mathcal{P}_{D_{[n]\setminus T}} y &= \sum_{i\in [n]\setminus T}c_i\mathbf{d}_i + \sum_{i\in T}c_i\mathcal{P}_{[n]\setminus T} \mathbf{d}_i + \mathcal{P}_{D_{[n]\setminus T}} \vec{e}\\
\mathcal{P}_{D} y - \mathcal{P}_{D_{[n]\setminus T}} y
&= \sum_{i\in T}c_i\mathbf{d}_i - \sum_{i\in T}c_i\mathcal{P}_{[n]\setminus T} \mathbf{d}_i - \mathcal{P}_{D_{[n]\setminus T}} \vec{e}\\
&= \sum_{i\in T\cap S^0} c_i (\mathcal{I} -\mathcal{P}_{[n]\setminus T} ) \mathbf{d}_i+\sum_{i\in T\setminus S^0} c_i \mathcal{P}_{\spann(D_{[n]\setminus T})^\perp}\mathbf{d}_i - \mathcal{P}_{D_{[n]\setminus T}} \vec{e}.
\end{align*}
Thus, we have
\begin{align}\label{S-9}
\begin{aligned}
\|\mathcal{P}_{D} y - \mathcal{P}_{D_{[n]\setminus T}} y\|_2 
&\geq \| (I - P_{D_{[n]\setminus T}})D_{T\cap S^0}\vec{c}_{T\cap S^0}\|_2  -(n-k-\ell) \lambda_2 -\varepsilon\\
&\geq \sqrt{{\rm R}( G, \vec{c}_{T\cap S^0})} \|\vec{c}_{T\cap S^0}\|_2 -(n-k-\ell) \lambda_2 -\varepsilon\\
&\geq \ell \sqrt{{\rm R}( G, \vec{c}_{T\cap S^0})} \lambda_1 -(n-k-\ell) \lambda_2 -\varepsilon
\end{aligned}
\end{align}
where $G$ is the Gram matrix $D_{T\cap S^0}^T(I-P_{D_{[n]\setminus T}})D_{T\cap S^0}$, and $R(G,\vec{c}_{T\cap S^0})$ is the Rayleigh quotient for the matrix $G$ and the vector $\vec{c}_{T\cap S^0}$. Combining the condition \eqref{S-11}, estimate \eqref{S-10} and \eqref{S-9}, we prove \eqref{S-10}.
\end{proof}

\subsection{Proof for Theorem \ref{P1.3}}
\begin{proof}
We use the mathematical induction in $n=0,\ldots, k-1$.

($n=1$) We show $\mathbf{d}_{j^0}\in [n]\setminus S^0$. Suppose not. Then, $|c_{j^0}|\geq \lambda_1$. Recalling the condition \eqref{S-12}(; $\lambda_1\omega\geq \lambda_2$) and the fact $\omega_i\leq 1$, we have
\[
|c_{j^0}|\omega_{j^0}\geq \lambda_1\omega\geq \lambda_2\geq \lambda_2\omega_i,\quad \forall~i=1,\ldots,n.
\]
On the other hand, score is determined by $|c_i|\omega_i$ thanks to Lemma \ref{L1.1}. Thus, we have the contradiction to the minimal condition of $\mathbf{d}_{j^0}$ in \eqref{SSR_score}.

($n=m\leq k-1$) Assume that $\{\mathbf{d}_{j^0},\ldots,\mathbf{d}_{j^{m-1}}\}\subset [n]\setminus S^0$. Our claim is to show $\mathbf{d}_{j^m}\in [n]\setminus S^0$. Suppose not. By the construction \eqref{SSR_score}, we have 
\begin{align}\label{S-15}
\score(D_{\{j^0,\ldots,j^m\}};D) \leq \score(D_{\{j^0,\ldots,j^{m-1},i_0\}};D),\quad i_0\in[n]\setminus  \{j^0,\ldots,j^{m-1},j^m\}.
\end{align}
On the other hand, for $i^m\in [n] \setminus (S^0\cup \{j^0,\ldots,j^{m-1}\})$ and $I^m := \{j^0,\ldots,j^{m-1},i^m\}$,
\begin{align*}
\mathcal{P}_{D_{[n]\setminus I^m}}y &= \sum_{i\in [n]\setminus  I^m}c_i\mathbf{d}_i + \sum_{i\in I^m} c_i\mathcal{P}_{D_{I^m}}\mathbf{d}_i + \mathcal{P}_{D_{[n]\setminus I^m}}\vec{e},\\
\mathcal{P}_D y - \mathcal{P}_{D_{[n]\setminus I^m}}y 
&=  \sum_{i\in  I^m} c_i(\mathcal{I}-\mathcal{P}_{D_{[n]\setminus I^m}})\mathbf{d}_i - \mathcal{P}_{D_{[n]\setminus I^m}}\vec{e}.
\end{align*}
These directly yield the following:
\begin{align}\label{S-13}
\|\mathcal{P}_D y - \mathcal{P}_{D_{I^m}}y \|_2 \leq (m+1)\lambda_2 + \varepsilon
\end{align}
On the other hand, projections to the sub-dictionary generated by $J^m$ give us
\begin{align*}
\mathcal{P}_{D_{J^m}} y &= \sum_{i\in J^m} c_i\mathbf{d}_i + \sum_{i\in[n]\setminus J^m}c_i\mathcal{P}_{D_{J^m}} \mathbf{d}_i + \mathcal{P}_{D_{J^m}} \vec{e},\\
\mathcal{P}_Dy - \mathcal{P}_{D_{J^m}} y &= \sum_{i\in [n]\setminus J^m} c_i\mathbf{d}_i - \sum_{i\in[n]\setminus J^m}c_i\mathcal{P}_{D_{J^m}} \mathbf{d}_i - \mathcal{P}_{D_{J^m}} \vec{e},\\
&= c_{j^m}(\mathcal{I} -\mathcal{P}_{D_{J^m}})\mathbf{d}_{j^m} + \sum_{i\in [n]\setminus J^{m-1}}c_i\mathcal{P}_{\spann(D_{J^m})^\perp}\mathbf{d}_i- \mathcal{P}_{D_{J^m}} \vec{e}.
\end{align*}
So, we obtain
\begin{align}\label{S-14}
\lambda_1 \|(\mathcal{I} -\mathcal{P}_{D_{J^m}})\mathbf{d}_{j^m}\| - m\lambda_2-\varepsilon \leq \|\mathcal{P}_Dy - \mathcal{P}_{D_{J^m}} y\|_2.
\end{align}
Note that the condition $\eqref{S-12}_1$ implies $\|(\mathcal{I} -\mathcal{P}_{D_{J^m}})\mathbf{d}_{j^m}\|>\omega$, since $\spann(D_{J^m})\subset \spann(D_{[n]\setminus\{j^m\}})$. Together with the estimates \eqref{S-13} and \eqref{S-14}, and the condition $\eqref{S-12}_2$, we obtain
\[
\|\mathcal{P}_D y - \mathcal{P}_{D_{I^m}}y \|_2 < \|\mathcal{P}_Dy - \mathcal{P}_{D_{J^m}} y\|_2.
\]
This contradicts the minimality assumption \eqref{S-15}.

We showed that $\{j^0,\ldots, j^{k-1}\}=[n]\setminus S^0$. The conclusion comes from Proposition \ref{P1.2}.
\end{proof}

\section{Literature}\label{app2}

\subsection{Z-SINDy}\label{app2-Z}
Z-SINDy defines the free energy for each sub-dictionary and finds the minimal energy regime, using Bayesian inference within a statistical mechanical approach to sparse equation discovery \cite{KBKM24}. In this subsection, we see a similarity between two minimization problems.

Z-SINDy computes the free energy $F_\gamma$ of a sub-dictionary $\mathcal{D}_{sub}=\gamma$
\[
F_\gamma = -\log\mathcal{Z}_0 -\frac{|\gamma|}{2}\log(2\pi \rho^2) + \frac{1}{2}\log \det C_{\gamma} - \frac{1}{2\rho^2} \vec{V}_\gamma^T C_\gamma^{-1} \vec{V}_\gamma + \lambda \frac{T}{\Delta t}|\gamma|
\]
where $\mathcal{Z}_0$ is the evidence of an empty set, $\rho$ is the noise resolution, matrix of dictionary $V_\gamma = D^T_{sub}y$ and empirical correlation matrix $C_\gamma =D^{T}_{sub}D_{sub}$. Here, we solve the following minimization problem:
\begin{align}
\begin{aligned}
\mbox{arg}\min_\gamma F_\gamma 
&= \mbox{arg}\max_{\mathcal{D}_{sub}\subset \mathcal{D}}  \mathbf{y}^T \mathcal{P}_{D_{sub}}\mathbf{y}-c_1|D_{sub}|-c_2\log(\det C_\gamma)\\
&=\mbox{arg}\min_{\mathcal{D}_{sub}\subset \mathcal{D}}  
\| \mathbf{y} - \mathcal{P}_{D\setminus D_{sub}}\mathbf{y}\|^2
+c_1|D_{sub}|+c_2\log(\det C_\gamma)
\end{aligned}
\end{align}

Recall the tradeoff to sparsity in Subsection \ref{sec:tradeoff}. Suppose $y\in\mbox{span}(\mathcal{D}(\mathbf{x}))$.
Then, the minimization problem is
\begin{align*}
    \min_{\mathcal{D}_{sub}\subset \mathcal{D}} &\left\|(\mathcal{P}_{\mathcal{D}(\mathbf{x})} - \mathcal{P}_{\mathcal{D}-\mathcal{D}_{sub}(\mathbf{x})}) y\right\| +\lambda |\mathcal{D}\setminus\mathcal{D}_{sub}|\\
    &= \min_{\mathcal{D}_{sub}\subset \mathcal{D}} \left\|y - \mathcal{P}_{\mathcal{D}-\mathcal{D}_{sub}(\mathbf{x})} y\right\|^2 +\lambda |\mathcal{D}\setminus\mathcal{D}_{sub}| 
    = \max_{S=\mathcal{D}-\mathcal{D}_{sub}} y^T \mathcal{P}_{S} y+\lambda |S|
\end{align*}
where we used $\mathcal{P}_{\mathcal{D}-\mathcal{D}_{sub}(\mathbf{x})}=\mathcal{P}_{\mathcal{D}-\mathcal{D}_{sub}(\mathbf{x})}^2=\mathcal{P}_{\mathcal{D}-\mathcal{D}_{sub}(\mathbf{x})}^T$ and ignored constant terms. Since $\mathcal{P}_S = SS^\dagger$, we have
\begin{align*}
    \max_{S=\mathcal{D}-\mathcal{D}_{sub}} y^T S S^\dagger y+\lambda |S|
    = \max_{S=\mathcal{D}-\mathcal{D}_{sub}} (S^Ty)^T (S^TS)^{-1} S^Ty+\lambda |S|.
\end{align*}

\subsection{D-SINDy}\label{app2-D}
In the paper \cite{Wentz23}, the authors combined two algorithms; Projection based State Denoising (PSDN) and Iteratively reweighted second order cone program (IRW-SOCP). While our method deletes dictionary items with minimal scores each iteration, D-SINDy imposes a weight for each item along its importance. The authors use an architecture similar to iteratively reweighted Lasso (IRW-Lasso), which solves the Lasso $\ell_1$-regularization problem:
\[
\mbox{minimize}_\mathbf{c} \|A\mathbf{c} - y\|^2 + \lambda \|W\mathbf{c}\|_1.
\]
Here, $W$ is a diagonal matrix such that $W_{ii} = (|\mathbf{c}_i|+\varepsilon)^{-1}$ and $\mathbf{c}$ is the coefficients vector obtained in the previous iteration. In D-SINDy, their weight vector is $\mathbf{c}=\mathcal{P}_\mathcal{D}y$. In other words, it emphasizes smallness of coefficients of projected vectors.


\subsection{Orthogonal Matching Pursuit}\label{app2-OMP}
The Matching Pursuit (MP) is a sparse approximation algorithm \cite{Mallat93}. The Orthogonal Matching Pursuit (OMP) is a well-known extension of MP \cite{Pati93}. For a dictionary matrix $D=[\mathbf{d}_1|\cdots|\mathbf{d}_n]$ which is normalized columnwise and a target signal $y$, we want to find sparse coefficient vector $x$ which is the best matching projections of $y$ onto the span of dictionary. For a threshold $\delta>0$, it detects subindex sets $S^k$ and a sequence $(x^k)$
\begin{align*}
&R^0 = y,\quad S^0 = \emptyset,\\
&\gamma_k = \argmax_{i\notin S^{k-1}} |\langle \mathbf{d}_i,R^{k-1}\rangle|,\quad 
S^k = S^{k-1}\cup \{\gamma_k\},\quad x^k = \argmin_{\supp(x)\subset S^k}\|Dx - y\|_2,\\
&R^k = y - Dx^k,\quad k=1,\ldots, \min_i \{ i: |R^i|<\delta\}.
\end{align*}
At each iteration step, it finds the maximum of the following for a greedy searching
\begin{align}\label{app:C-1}
 |\langle \mathbf{d}_i,R^{k}\rangle|
=  |\langle \mathbf{d}_i, y - \mathcal{P}_{D_{S^k}}y\rangle|
=  |\langle \mathbf{d}_i, (I - \mathcal{P}_{D_{S^k}})y\rangle|
=  |\langle (I - \mathcal{P}_{D_{S^k}}) \mathbf{d}_i, y\rangle|
\end{align}
where we used the projection property $(I - \mathcal{P}_{D_{S^k}})^T =I - \mathcal{P}_{D_{S^k}}$. Also,
\begin{align}\label{app:C-2}
\begin{aligned}
|\langle \mathbf{d}_i,R^{k}\rangle| &= \|(\mathcal{P}_{D_{S^k\cup\{i\}}} - \mathcal{P}_{D_{S^k}})y\|_2 \|(I- \mathcal{P}_{D_{S^k}})\mathbf{d}_i\|_2 \\
&= \score(\mathbf{d}_i;D_{S^k\cup\{i\}}) \|y\|_2\|(I- \mathcal{P}_{D_{S^k}})\mathbf{d}_i\|_2
\end{aligned}
\end{align}
where we used Lemma \ref{L1.1} and $D_{S^k\cup\{i\}}$ is a super-matrix $D_{S^k}$ and its last column is $\mathbf{d}_i$. Precisely, from the block matrix pseudoinverse formula, we have 
\begin{align}\label{app:C-3}
[P_{D_{S^k\cup\{i\}}}y]_{k+1}= (P^\perp_{D_{S^k}}\mathbf{d}_i)^\dagger y = \frac{|\langle (I - \mathcal{P}_{D_{S^k}}) \mathbf{d}_i, y\rangle|}{\|\mathbf{d}_i-P_{D_{S^k}}\mathbf{d}_i\|_2^2}.
\end{align}
We combine \eqref{app:C-1} and \eqref{app:C-3} to obtain \eqref{app:C-4}, which implies \eqref{app:C-2}.
\begin{align}\label{app:C-4}
\frac{\score(\mathbf{d}_i;D_{S^k\cup\{i\}}) \|y\|_2}{\|\mathbf{d}_i - \mathcal{P}_{D_{S^k}}\mathbf{d}_i \|_2}
 = \Big|[P_{D_{S^k\cup\{i\}}}y]_{k+1}\Big| = \frac{(P_{D_{S^k}}^\perp\mathbf{d}_i)^T y}{\|\mathbf{d}_i-P_{D_{S^k}}\mathbf{d}_i\|_2^2}.
\end{align}

As a comparison, if the score of an item is small at each iteration for OMP, then OMP may not take the item (provided that mutual incoherences between dictionary items are similar).

\subsection{Screening process}\label{sec:screening-process}
This procedure addresses the case with many variables whose numbers exceed the number of sampling points, safely discarding variables to facilitate speed for the LASSO problem. Several methods have been developed such as
Sure Independence Screening which selects the top features most correlated with the target variable \cite{Fan08, Fan10} and
SAfe Feature Elimination (SAFE) which safely discards features guaranteed to have zero coefficients in the LASSO solution, based on a conservative bound involving the regularization parameter \cite{Ghaoui10}.

In the present paper, we do not use their method, but use the structure of the schemes. As an application of scoring, we propose a preprocessing step of STLS which (unsafely) discard dictionary items in advance but safely under certain conditions.

We tested whether the filtering approach can aid in constructing a base dictionary for STLS. In this context, we treat the filtered dictionary terms as the base dictionary within the weak formulation of STLS—a process analogous to a screening step. A natural question arises: “How many terms should be filtered out to improve upon standard STLS?” We provide a partial numerical answer to this question and also discuss the limitations of our screening-like strategy.

\begin{figure}
    \centering
    \includegraphics[width=\linewidth]{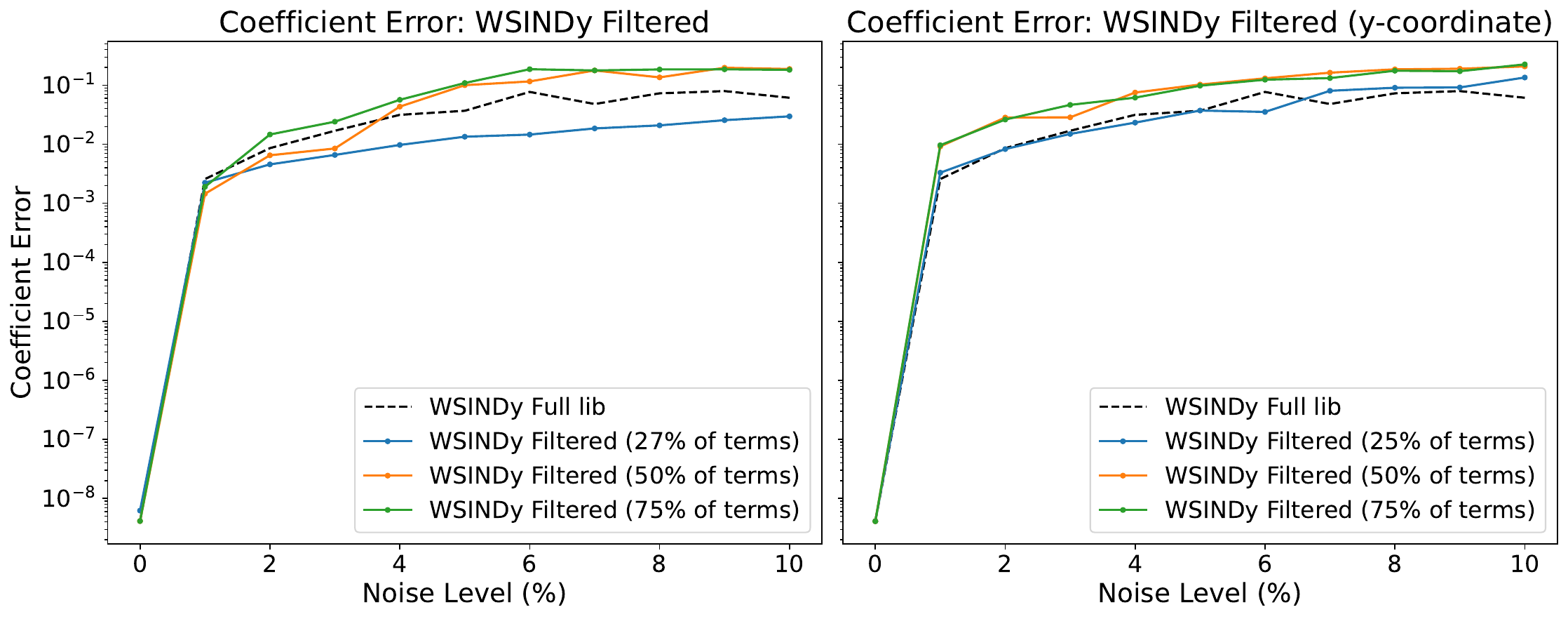}
    \caption{Coefficient errors of STLS results over filtered dictionaries in the weak formulation at each noise level. The left figure presents results based on filtered sub-dictionaries that use all coordinate scores; that is, the removal order is determined by the sum of scores across coordinates. The right figure shows the result based on the $y$ coordinate only. }
    \label{fig:noise_screening}
\end{figure}

In Figure \ref{fig:noise_screening}, we compare WSINDy results with dictionaries filtered $n$ percents by scoring from the based dictionary. Power of scoring in noise test appears when we filter more than half amount of items over the base dictionary. On the other hand, the other cases do not this is because the significant variables appearing uncorrelated until conditioned on other variables.

\section{Greedy Forward Stepwise Regressor}
\label{app:GFSR}
In this section, we numerically show that GBSR is better than GFSR, which is given as follows:
\begin{align}\label{eq:GFSR}\tag{GFSR}
\begin{aligned}
&{j^0_f}=\argmin_{i\in[m]} \mbox{Score}(D_{[n]\setminus\{i\}};D,y),\quad J^0_f=\{j^0_f\},\\
&{j^i_f}=\argmin_{\ell\not\in J^{i-1}_f} \mbox{Score}(D_{[n]\setminus (J_f^{i-1}\cup\{\ell\})};D,y),\quad J^i_f = J^{i-1}_f\cup\{j^i_f\},\quad i=1,\ldots, n. 
\end{aligned}
\end{align}
A natural question arises: "Why remove items from a library instead of building it up incrementally?" Given the assumption that the underlying system is sparse, a removal-based approach (i.e., filtering) could, in principle, require more iterations.

\begin{figure}[t!]
    \centering
    \includegraphics[width=\linewidth]{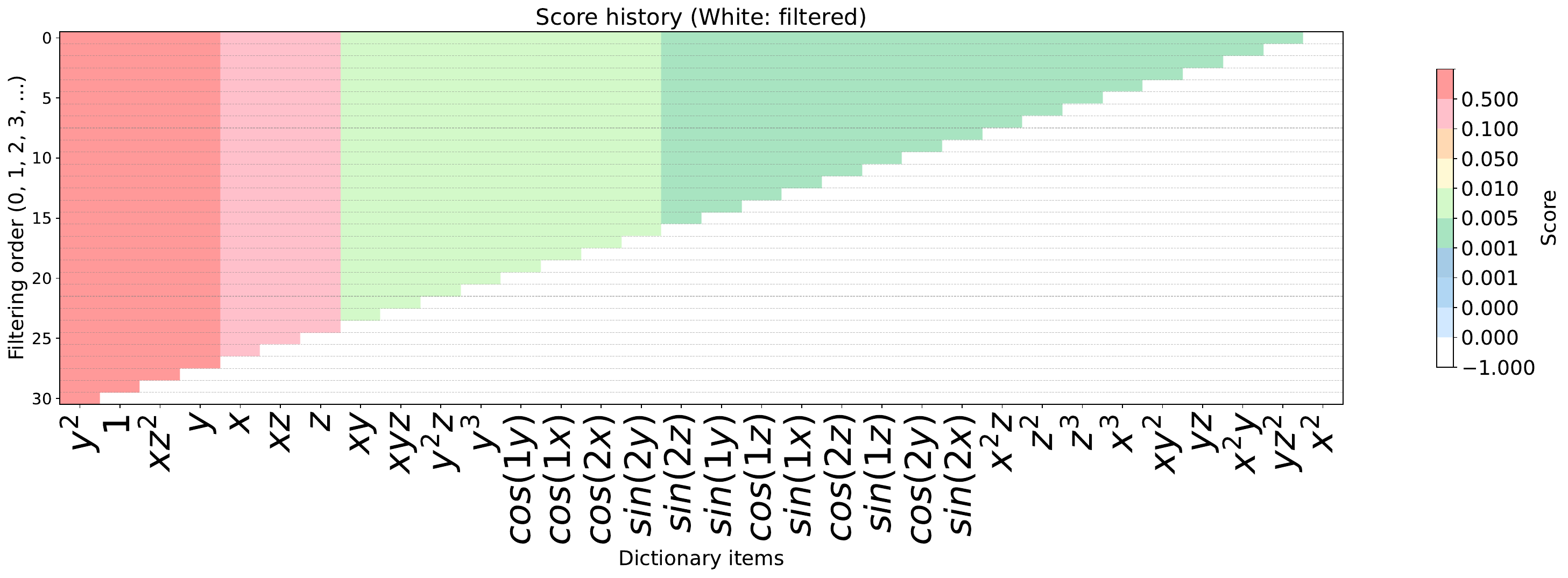}
    \caption{Ranking the library terms for Lorenz system \eqref{eq:lorenz} using the score \eqref{app:GFSR}. In the very first selection step, an incorrect term is chosen, which compromises the remainder of the selection process.}
    \label{fig:lorenz_forward}
\end{figure}

In our numerical experiments, we tested a forward selection strategy using our scoring method. However, as shown in Figure~\ref{fig:lorenz_forward} for the Lorenz system, the initial selection is often random and highly prone to error, leading to poor identification of the correct terms.

One potential solution to this problem is to begin the process with an exhaustive search over combinations of multiple terms. However, this strategy introduces complications for systems in which some equations contain only a single relevant term---such cases would be missed by multi--term initialization.

%
%
%
\bibliographystyle{elsarticle-num}
\bibliography{bib}

@article{Berry25,
  title = {Limits of Learning Dynamical Systems},
  volume = {67},
  ISSN = {1095-7200},
  DOI = {10.1137/24m1696974},
  number = {1},
  journal = {SIAM Review},
  publisher = {Society for Industrial & Applied Mathematics (SIAM)},
  author = {Berry,  Tyrus and Das,  Suddhasattwa},
  year = {2025},
  month = feb,
  pages = {107–137}
}

@book{Bjorck96,
  title = {Numerical Methods for Least Squares Problems},
  ISBN = {9781611971484},
  DOI = {10.1137/1.9781611971484},
  publisher = {Society for Industrial and Applied Mathematics},
  author = {Bj{\"o}rck, {\AA}ke},
  year = {1996},
  month = jan 
}

@article{Bongard07,
  title = {Automated reverse engineering of nonlinear dynamical systems},
  volume = {104},
  ISSN = {1091-6490},
  DOI = {10.1073/pnas.0609476104},
  number = {24},
  journal = {Proceedings of the National Academy of Sciences},
  publisher = {Proceedings of the National Academy of Sciences},
  author = {Bongard,  Josh and Lipson,  Hod},
  year = {2007},
  month = jun,
  pages = {9943–9948}
}

@article{Boninsegna18,
  title = {Sparse learning of stochastic dynamical equations},
  volume = {148},
  ISSN = {1089-7690},
  DOI = {10.1063/1.5018409},
  number = {24},
  journal = {The Journal of Chemical Physics},
  publisher = {AIP Publishing},
  author = {Boninsegna,  Lorenzo and N\"{u}ske,  Feliks and Clementi,  Cecilia},
  year = {2018},
  month = mar 
}

@article{Bramburger20,
  title = {Poincaré maps for multiscale physics discovery and nonlinear Floquet theory},
  volume = {408},
  ISSN = {0167-2789},
  DOI = {10.1016/j.physd.2020.132479},
  journal = {Physica D: Nonlinear Phenomena},
  publisher = {Elsevier BV},
  author = {Bramburger,  Jason J. and Kutz,  J. Nathan},
  year = {2020},
  month = jul,
  pages = {132479}
}

@book{Bramburger24,
  title = {Data-Driven Methods for Dynamic Systems},
  ISBN = {9781611978162},
  DOI = {10.1137/1.9781611978162},
  publisher = {Society for Industrial and Applied Mathematics},
  author = {Bramburger,  Jason J.},
  year = {2024},
  month = jan 
}

@article{Brunton16,
  title = {Discovering governing equations from data by sparse identification of nonlinear dynamical systems},
  volume = {113},
  ISSN = {1091-6490},
  DOI = {10.1073/pnas.1517384113},
  number = {15},
  journal = {Proceedings of the National Academy of Sciences},
  publisher = {Proceedings of the National Academy of Sciences},
  author = {Brunton,  Steven L. and Proctor,  Joshua L. and Kutz,  J. Nathan},
  year = {2016},
  month = mar,
  pages = {3932–3937}
}

@book{Brunton22,
  title = {Data-Driven Science and Engineering: Machine Learning,  Dynamical Systems,  and Control},
  ISBN = {9781108422093},
  DOI = {10.1017/9781108380690},
  publisher = {Cambridge University Press},
  author = {Brunton,  Steven L. and Kutz,  J. Nathan},
  year = {2019},
  month = jan 
}

@article{Candes06,
author = {Candès, Emmanuel J. and Romberg, Justin K. and Tao, Terence},
title = {Stable signal recovery from incomplete and inaccurate measurements},
journal = {Communications on Pure and Applied Mathematics},
volume = {59},
number = {8},
pages = {1207-1223},
doi = {https://doi.org/10.1002/cpa.20124},
year = {2006}
}

@article{Carderera21,
  title={CINDy: Conditional gradient-based Identification of Non-linear Dynamics--Noise-robust recovery},
  author={Carderera, Alejandro and Pokutta, Sebastian and Sch{\"u}tte, Christof and Weiser, Martin},
  journal={Journal of Computational and Applied Mathematics},
  year={2021}
}

@article{Champion19,
  title = {Data-driven discovery of coordinates and governing equations},
  volume = {116},
  ISSN = {1091-6490},
  DOI = {10.1073/pnas.1906995116},
  number = {45},
  journal = {Proceedings of the National Academy of Sciences},
  publisher = {Proceedings of the National Academy of Sciences},
  author = {Champion,  Kathleen and Lusch,  Bethany and Kutz,  J. Nathan and Brunton,  Steven L.},
  year = {2019},
  month = oct,
  pages = {22445–22451}
}

@ARTICLE{Champion20,
  author={Champion, Kathleen and Zheng, Peng and Aravkin, Aleksandr Y. and Brunton, Steven L. and Kutz, J. Nathan},
  journal={IEEE Access}, 
  title={A Unified Sparse Optimization Framework to Learn Parsimonious Physics-Informed Models From Data}, 
  year={2020},
  volume={8},
  number={},
  pages={169259-169271},
  doi={10.1109/ACCESS.2020.3023625}}

@ARTICLE{Donoho06,
  author={Donoho, D.L.},
  journal={IEEE Transactions on Information Theory}, 
  title={Compressed sensing}, 
  year={2006},
  volume={52},
  number={4},
  pages={1289-1306},
  doi={10.1109/TIT.2006.871582}}

@article{Fan08,
  title={Sure independence screening for ultrahigh dimensional feature space},
  author={Fan, Jianqing and Lv, Jinchi},
  journal={Journal of the Royal Statistical Society Series B: Statistical Methodology},
  volume={70},
  number={5},
  pages={849--911},
  year={2008},
  publisher={Oxford University Press}
}

@article{Fan10,
  title={A selective overview of variable selection in high dimensional feature space},
  author={Fan, Jianqing and Lv, Jinchi},
  journal={Statistica Sinica},
  volume={20},
  number={1},
  pages={101},
  year={2010}
}

@ARTICLE{FKBB22,
  author={Fasel, Urban and Kutz, J. Nathan and Brunton, Bingni W. and Brunton, Steven L.},
  journal={Proc. R. Soc. A}, 
  title={Ensemble-SINDy: Robust sparse model discovery in the low-data, high-noise limit, with active learning and control}, 
  year={2022},
  volume={478},
  number={20210904},
  doi={10.1098/rspa.2021.0904}}

@INPROCEEDINGS{Fasel21,
  author={Fasel, Urban and Kaiser, Eurika and Kutz, J. Nathan and Brunton, Bingni W. and Brunton, Steven L.},
  booktitle={2021 60th IEEE Conference on Decision and Control (CDC)}, 
  title={SINDy with Control: A Tutorial}, 
  year={2021},
  volume={},
  number={},
  pages={16-21},
  doi={10.1109/CDC45484.2021.9683120}}

@book{Foucart13,
title={A Mathematical Introduction to Compressive Sensing}, ISBN={9780817649487}, ISSN={2296-5017}, 
DOI={10.1007/978-0-8176-4948-7},
journal={Applied and Numerical Harmonic Analysis},
publisher={Springer New York},
author={Foucart, Simon and Rauhut, Holger},
year={2013} 
}

@inproceedings{fung25,
  title={Rapid Bayesian identification of sparse nonlinear dynamics from scarce and noisy data},
  author={Fung, Lloyd and Fasel, Urban and Juniper, Matthew},
  booktitle={Proceedings A},
  volume={481},
  pages={20240200},
  year={2025},
  organization={The Royal Society}
}

@article{GFBK23,
  title={Convergence of uncertainty estimates in ensemble and Bayesian sparse model discovery},
  author={Gao, L and Fasel, Urban and Brunton, Steven L and Kutz, J Nathan},
  journal={arXiv preprint arXiv:2301.12649},
  year={2023}
}

@article{Ghadami22,
  title={Data-driven prediction in dynamical systems: recent developments},
  author={Ghadami, Amin and Epureanu, Bogdan I},
  journal={Philosophical Transactions of the Royal Society A},
  volume={380},
  number={2229},
  pages={20210213},
  year={2022},
  publisher={The Royal Society}
}

@techreport{Ghaoui10,
	Author = {Laurent {El Ghaoui} and Vivian Viallon and Tarek Rabbani},
	Institution = {EECS Dept., University of California at Berkeley},
	Month = {September},
	Number = {UC/EECS-2010-126},
	Title = {Safe Feature Elimination in Sparse Supervised Learning},
	Year = {2010}}

@article{Goyal22,
  title={Discovery of nonlinear dynamical systems using a Runge--Kutta inspired dictionary-based sparse regression approach},
  author={Goyal, Pawan and Benner, Peter},
  journal={Proceedings of the Royal Society A},
  volume={478},
  number={2262},
  pages={20210883},
  year={2022},
  publisher={The Royal Society}
}

@article{Hokanson23,
author = {Hokanson, Jeffrey M. and Iaccarino, Gianluca and Doostan, Alireza},
title = {Simultaneous Identification and Denoising of Dynamical Systems},
journal = {SIAM Journal on Scientific Computing},
volume = {45},
number = {4},
pages = {A1413-A1437},
year = {2023},
doi = {10.1137/22M1486303}
}

@misc{
Jacobs24,
title={HyperSINDy: Deep Generative Modeling of Nonlinear Stochastic Governing Equations},
author={Mozes Jacobs and Bingni W Brunton and Steven Brunton and J. Nathan Kutz and Ryan V. Raut},
year={2024},
url={https://openreview.net/forum?id=B4XM9nQ8Ns}
}

@article{Kaheman20,
  title={SINDy-PI: a robust algorithm for parallel implicit sparse identification of nonlinear dynamics},
  author={Kaheman, Kadierdan and Kutz, J Nathan and Brunton, Steven L},
  journal={Proceedings of the Royal Society A},
  volume={476},
  number={2242},
  pages={20200279},
  year={2020},
  publisher={The Royal Society Publishing}
}

@article{kaszas20,
	title = {Universal upper estimate for prediction errors under moderate model uncertainty},
	volume = {30},
	issn = {1054-1500},
	doi = {10.1063/5.0021665},
	number = {11},
	journal = {Chaos: An Interdisciplinary Journal of Nonlinear Science},
	author = {Kaszás, Bálint and Haller, George},
	month = nov,
	year = {2020},
    pages = {113144},
}

@article{KBKM24,
  title = {Statistical mechanics of dynamical system identification},
  author = {Klishin, Andrei A. and Bakarji, Joseph and Kutz, J. Nathan and Manohar, Krithika},
  journal = {Phys. Rev. Res.},
  volume = {7},
  issue = {3},
  pages = {033181},
  numpages = {16},
  year = {2025},
  month = {Aug},
  publisher = {American Physical Society},
  doi = {10.1103/4d98-tdlp},
}

@article{Mangan17,
  title={Model selection for dynamical systems via sparse regression and information criteria},
  author={Mangan, Niall M and Kutz, J Nathan and Brunton, Steven L and Proctor, Joshua L},
  journal={Proceedings of the Royal Society A: Mathematical, Physical and Engineering Sciences},
  volume={473},
  number={2204},
  pages={20170009},
  year={2017},
  publisher={The Royal Society Publishing}
}

@ARTICLE{Mallat93,
  author={Mallat, S.G. and Zhifeng Zhang},
  journal={IEEE Transactions on Signal Processing}, 
  title={Matching pursuits with time-frequency dictionaries}, 
  year={1993},
  volume={41},
  number={12},
  pages={3397-3415},
  doi={10.1109/78.258082}}

@book{Mallat99,
  title={A wavelet tour of signal processing},
  author={Mallat, St{\'e}phane},
  year={1999},
  publisher={Elsevier}
}

@article{Messenger24,
  title={The Weak Form is Stronger Than You Think},
  author={Messenger, Daniel and Tran, April and Dukic, Vanja and Bortz, David},
  journal={SIAM News},
  volume={57},
  number={8},
  year={2024},
  publisher={SIAM}
}

@article{Messenger21,
author = {Messenger, Daniel A. and Bortz, David M.},
title = {Weak SINDy: Galerkin-Based Data-Driven Model Selection},
journal = {Multiscale Modeling \& Simulation},
volume = {19},
number = {3},
pages = {1474-1497},
year = {2021},
doi = {10.1137/20M1343166}
}

@article{Messenger21-pde,
title = {Weak SINDy for partial differential equations},
journal = {Journal of Computational Physics},
volume = {443},
pages = {110525},
year = {2021},
issn = {0021-9991},
doi = {https://doi.org/10.1016/j.jcp.2021.110525},
author = {Daniel A. Messenger and David M. Bortz}
}

@book{Montgomery21,
  title={Introduction to linear regression analysis},
  author={Montgomery, Douglas C and Peck, Elizabeth A and Vining, G Geoffrey},
  year={2021},
  publisher={John Wiley \& Sons}
}

@article{Naozuka22,
  title={SINDy-SA framework: enhancing nonlinear system identification with sensitivity analysis},
  author={Naozuka, Gustavo T and Rocha, Heber L and Silva, Renato S and Almeida, Regina C},
  journal={Nonlinear Dynamics},
  volume={110},
  number={3},
  pages={2589--2609},
  year={2022},
  publisher={Springer}
}

@article{Nicolaou23,
  title = {Data-driven discovery and extrapolation of parameterized pattern-forming dynamics},
  author = {Nicolaou, Zachary G. and Huo, Guanyu and Chen, Yihui and Brunton, Steven L. and Kutz, J. Nathan},
  journal = {Phys. Rev. Res.},
  volume = {5},
  issue = {4},
  pages = {L042017},
  numpages = {7},
  year = {2023},
  month = {Nov},
  publisher = {American Physical Society},
  doi = {10.1103/PhysRevResearch.5.L042017},
}

@INPROCEEDINGS{Naik12,
  author={Naik, Manjish and Cochran, Douglas},
  booktitle={2012 Conference Record of the Forty Sixth Asilomar Conference on Signals, Systems and Computers (ASILOMAR)}, 
  title={Nonlinear system identification using compressed sensing}, 
  year={2012},
  volume={},
  number={},
  pages={426-430},
  doi={10.1109/ACSSC.2012.6489039}}

@article{Oishi24,
  title={Nonlinear parametric models of viscoelastic fluid flows},
  author={Oishi, Cassio M and Kaptanoglu, Alan A and Kutz, J Nathan and Brunton, Steven L},
  journal={Royal Society Open Science},
  volume={11},
  number={10},
  pages={240995},
  year={2024},
  publisher={The Royal Society}
}

@INPROCEEDINGS{Pati93,
  author={Pati, Y.C. and Rezaiifar, R. and Krishnaprasad, P.S.},
  booktitle={Proceedings of 27th Asilomar Conference on Signals, Systems and Computers}, 
  title={Orthogonal matching pursuit: recursive function approximation with applications to wavelet decomposition}, 
  year={1993},
  volume={},
  number={},
  pages={40-44 vol.1},
  doi={10.1109/ACSSC.1993.342465}}

@article{
Rudy17,
author = {Samuel H. Rudy  and Steven L. Brunton  and Joshua L. Proctor  and J. Nathan Kutz },
title = {Data-driven discovery of partial differential equations},
journal = {Science Advances},
volume = {3},
number = {4},
pages = {e1602614},
year = {2017},
doi = {10.1126/sciadv.1602614}}

@article{Russo25,
author = {Russo, Benjamin P. and Laiu, M. Paul and Archibald, Richard},
title = {Streaming Compression of Scientific Data via Weak-SINDy},
journal = {SIAM Journal on Scientific Computing},
volume = {47},
number = {1},
pages = {C207-C234},
year = {2025},
doi = {10.1137/23M1599331}
}

@article{Russo24,
author = {Russo, Benjamin P. and Laiu, M. Paul},
title = {Convergence of Weak-SINDy Surrogate Models},
journal = {SIAM Journal on Applied Dynamical Systems},
volume = {23},
number = {2},
pages = {1017-1051},
year = {2024},
doi = {10.1137/22M1526782}
}

@article{Shea21,
  title={SINDy-BVP: Sparse identification of nonlinear dynamics for boundary value problems},
  author={Shea, Daniel E and Brunton, Steven L and Kutz, J Nathan},
  journal={Physical Review Research},
  volume={3},
  number={2},
  pages={023255},
  year={2021},
  publisher={APS}
}

@article{schmidt09,
  title={Distilling free-form natural laws from experimental data},
  author={Schmidt, Michael and Lipson, Hod},
  journal={science},
  volume={324},
  number={5923},
  pages={81--85},
  year={2009},
  publisher={American Association for the Advancement of Science}
}

@article{Su17,
  title={False discoveries occur early on the lasso path},
  author={Su, Weijie and Bogdan, Ma{\l}gorzata and Candes, Emmanuel},
  journal={The Annals of statistics},
  pages={2133--2150},
  year={2017},
  publisher={JSTOR}
}

@article{Szedmak24,
  title={Scalable variable selection for two-view learning tasks with projection operators},
  author={Szedmak, Sandor and Huusari, Riikka and Duong Le, Tat Hong and Rousu, Juho},
  journal={Machine Learning},
  volume={113},
  number={6},
  pages={3525--3544},
  year={2024},
  publisher={Springer}
}

@article{Tibshirani96,
  title={Regression shrinkage and selection via the lasso},
  author={Tibshirani, Robert},
  journal={Journal of the Royal Statistical Society Series B: Statistical Methodology},
  volume={58},
  number={1},
  pages={267--288},
  year={1996},
  publisher={Oxford University Press}
}

@article{Tran17,
author = {Tran, Giang and Ward, Rachel},
title = {Exact Recovery of Chaotic Systems from Highly Corrupted Data},
journal = {Multiscale Modeling \& Simulation},
volume = {15},
number = {3},
pages = {1108-1129},
year = {2017},
doi = {10.1137/16M1086637}
}

@article{tropp2006,
  title={Just relax: Convex programming methods for identifying sparse signals in noise},
  author={Tropp, Joel A},
  journal={IEEE transactions on information theory},
  volume={52},
  number={3},
  pages={1030--1051},
  year={2006},
  publisher={IEEE}
}

@article{Van09,
  title={Probing the Pareto frontier for basis pursuit solutions},
  author={Van Den Berg, Ewout and Friedlander, Michael P},
  journal={Siam journal on scientific computing},
  volume={31},
  number={2},
  pages={890--912},
  year={2009},
  publisher={SIAM}
}

@article{Viknesh24,
  title={Adam-sindy: An efficient optimization framework for parameterized nonlinear dynamical system identification},
  author={Viknesh, Siva and Tatari, Younes and Arzani, Amirhossein},
  journal={arXiv preprint arXiv:2410.16528},
  year={2024}
}

@book{wainwright2019,
  title={High-dimensional statistics: A non-asymptotic viewpoint},
  author={Wainwright, Martin J},
  volume={48},
  year={2019},
  publisher={Cambridge university press}
}

@article{Wang11,
  title = {Predicting Catastrophes in Nonlinear Dynamical Systems by Compressive Sensing},
  author = {Wang, Wen-Xu and Yang, Rui and Lai, Ying-Cheng and Kovanis, Vassilios and Grebogi, Celso},
  journal = {Phys. Rev. Lett.},
  volume = {106},
  issue = {15},
  pages = {154101},
  numpages = {4},
  year = {2011},
  month = {Apr},
  publisher = {American Physical Society},
  doi = {10.1103/PhysRevLett.106.154101},
}

@article{Wanner24,
author = {Wanner, Mathias and Mezi\'{c}, Igor},
title = {On Higher Order Drift and Diffusion Estimates for Stochastic SINDy},
journal = {SIAM Journal on Applied Dynamical Systems},
volume = {23},
number = {2},
pages = {1504-1539},
year = {2024},
doi = {10.1137/23M1567011}
}

@article{Wentz23,
title = {Derivative-based SINDy (DSINDy): Addressing the challenge of discovering governing equations from noisy data},
journal = {Computer Methods in Applied Mechanics and Engineering},
volume = {413},
pages = {116096},
year = {2023},
issn = {0045-7825},
doi = {https://doi.org/10.1016/j.cma.2023.116096},
author = {Jacqueline Wentz and Alireza Doostan}
}

@article{Zhang19,
author = {Zhang, Linan and Schaeffer, Hayden},
title = {On the Convergence of the SINDy Algorithm},
journal = {Multiscale Modeling \& Simulation},
volume = {17},
number = {3},
pages = {948-972},
year = {2019},
doi = {10.1137/18M1189828}
}

@article{Zheng18,
  title={A unified framework for sparse relaxed regularized regression: SR3},
  author={Zheng, Peng and Askham, Travis and Brunton, Steven L and Kutz, J Nathan and Aravkin, Aleksandr Y},
  journal={IEEE Access},
  volume={7},
  pages={1404--1423},
  year={2018},
  publisher={IEEE}
}

@article{Zolman24,
	title = {{SINDy}-{RL} for interpretable and efficient model-based reinforcement learning},
	volume = {16},
	issn = {2041-1723},
	doi = {10.1038/s41467-025-65738-4},
	number = {1},
	journal = {Nature Communications},
	author = {Zolman, Nicholas and Lagemann, Christian and Fasel, Urban and Kutz, J. Nathan and Brunton, Steven L.},
	month = nov,
	year = {2025},
	pages = {10714},
}

@article{Thomases2009,
  title={Transition to mixing and oscillations in a Stokesian viscoelastic flow.},
  author={Becca Thomases and Michael J. Shelley},
  journal={Physical review letters},
  year={2009},
  volume={103 9},
  pages={
          094501
        },
  url={https://api.semanticscholar.org/CorpusID:586707}
}
\end{document}